\documentclass[10pt,twocolumn,letterpaper]{article}

\usepackage{cvpr}              % To produce the CAMERA-READY version
\usepackage{multirow}
\usepackage{array}
\usepackage{colortbl}
\usepackage{graphicx}
\usepackage{tabularx}
\usepackage{booktabs}
\usepackage{subcaption}
\usepackage{floatrow}
\definecolor{lightgray}{gray}{0.9}

\newfloatcommand{capbtabbox}{table}[][\FBwidth]
\usepackage[utf8]{inputenc} % allow utf-8 input
\usepackage[T1]{fontenc}    % use 8-bit T1 fonts
\usepackage{xcolor}
\definecolor{lightblue}{RGB}{100,150,255} % tweak RGB as needed
\usepackage[colorlinks=true,citecolor=cyan,linkcolor=red,urlcolor=blue]{hyperref}
\usepackage{hyperref}       % hyperlinks
\usepackage{url}            % simple URL typesetting
\usepackage{booktabs}       % professional-quality tables
\usepackage{amsfonts}       % blackboard math symbols
\usepackage{nicefrac}       % compact symbols for 1/2, etc.
\usepackage{microtype}      % microtypography
\usepackage{xcolor}         % colors
\usepackage{sidecap}
\usepackage{float}
\usepackage{placeins}
\usepackage{wrapfig}
\usepackage{comment}
\usepackage{nicefrac}       % compact symbols for 1/2, etc.
\usepackage{microtype}      % microtypography
\usepackage{xcolor}         % colors
\usepackage{subcaption}
\usepackage{multirow}
\usepackage{verbatim}
\usepackage{diagbox}
\usepackage{bm}
\usepackage{makecell}
\usepackage{graphicx}
\usepackage{multirow}
\usepackage{amsmath}
\usepackage[english]{babel}
\usepackage{amsthm}
\usepackage{mathtools}
\usepackage[ruled,vlined]{algorithm2e}
\usepackage{xcolor}
\usepackage[utf8]{inputenc}
\usepackage{csquotes}
\usepackage{enumitem}
\usepackage{soul}
\soulregister\cite7
\soulregister\citet7
\soulregister\citep7
\soulregister\ref7
\soulregister\eqref7
\soulregister\textit7
\usepackage{lipsum} % for dummy text

\usepackage{booktabs} % For better looking tables
\usepackage{caption} % For customizing captions
\usepackage{rotating} % For rotating the text
\usepackage{multirow} % For multirow headers
\usepackage{lipsum} % for generating dummy text
\usepackage{colortbl}
\usepackage{arydshln}
\usepackage[utf8]{inputenc} % allow utf-8 input
\usepackage[T1]{fontenc}    % use 8-bit T1 fonts     % hyperlinks
\usepackage{url}            % simple URL typesetting
\usepackage{booktabs}       % professional-quality tables
\usepackage{amsfonts}       % blackboard math symbols
\usepackage{nicefrac}       % compact symbols for 1/2, etc.
\usepackage{microtype}      % microtypography
\usepackage{xcolor}         % colors

\usepackage{rotating}
\usepackage{tcolorbox}
\usepackage{enumitem}

\definecolor{cvprblue}{rgb}{0.21,0.49,0.74}

\title{Towards Calibrating Prompt Tuning of Vision- Language Models}
\author{
    Ashshak Sharifdeen$^{1,2}$, \hspace{0.25em}Fahad Shamshad$^{1}$, \hspace{0.25em} Muhammad Akhtar Munir$^{1}$, \\ \hspace{0.25em} Abhishek Basu$^{1}$, \hspace{0.25em} Mohamed Insaf Ismithdeen$^{1}$, \hspace{0.25em} 
Jeyapriyan Jeyamohan$^{2}$, \\ \hspace{0.25em} 
Chathurika Sewwandi Silva$^{2}$,\hspace{0.25em} 
Karthik Nandakumar$^{1,3}$, \hspace{0.25em} 
Muhammad Haris Khan$^{1}$ \\
    $^{1}$Mohamed bin Zayed University of AI, $^{2}$University of Colombo, $^{3}$Michigan State University \\
    {\tt\small \{ashshak.sharifdeen, muhammad.haris\}@mbzuai.ac.ae}
}

\begin{document}
 \maketitle
%\maketitlesupplementary
 \begin{abstract}
Prompt tuning of large-scale vision-language models such as CLIP enables efficient task adaptation without updating model weights. 
However, it often leads to poor confidence calibration and unreliable predictive uncertainty.
%However, it often leads to poor confidence calibration and degrades the semantic structure of the text embedding space. 
We address this problem by proposing a calibration framework that enhances predictive reliability while preserving the geometry of the pretrained CLIP embedding space, which is required for robust generalization. Our approach extends the standard cross-entropy loss with two complementary regularizers:
(1) a mean-variance margin penalty that stabilizes inter-class logit margins by maximizing their average while minimizing dispersion, mitigating underconfidence %on base classes 
and overconfidence spikes; and (2) a text moment-matching loss that aligns the first and second moments of tuned text embeddings with their frozen CLIP counterparts, preserving semantic dispersion crucial for generalization.
Through extensive experiments across 7 prompt-tuning methods and 11 diverse datasets, we demonstrate that our approach significantly reduces the Expected Calibration Error (ECE) compared to competitive calibration techniques on both base and novel classes. Our code is available at \url{https://github.com/ashshaksharifdeen/TCPT}.
\end{abstract}    
 \section{Introduction}
\label{sec:intro}

Vision language models (VLM), such as CLIP~\citep{radford2021learning}, have significantly advanced open-vocabulary image recognition by effectively using large-scale natural language supervision. 
To efficiently adapt these pre-trained models to downstream tasks, parameter-efficient techniques, particularly prompt tuning, have become popular~\citep{liu2023pre}.
Prompt tuning modifies only a small subset of parameters, substantially enhancing performance on seen (base) classes while preserving the model's inherent generalization ability to unseen (novel) classes~\citep{khattak2023maple, zhou2022conditional}.
This balance between specialization and generalization has driven the widespread adoption of prompt-tuned VLMs in healthcare, autonomous systems, and industrial applications where recognizing expected and unexpected visual concepts is essential for safe operation~\citep{zhao2025clip,elhenawy2025vision}.

\begin{figure}[t]
    \centering
    \includegraphics[width=\linewidth, trim=15 25 15 15, clip]{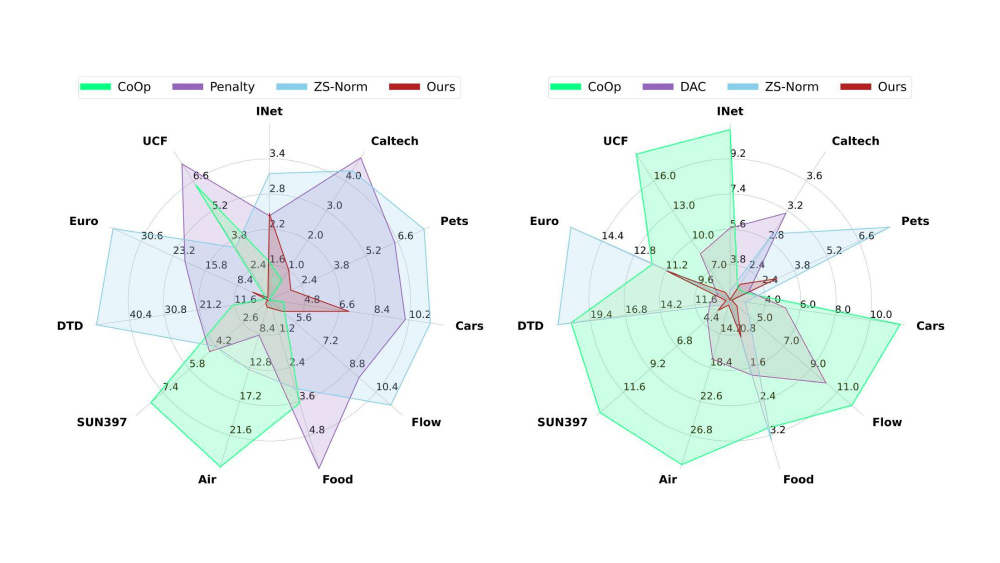}
    \caption{\small \textbf{Expected Calibration Error (ECE) on 11 datasets with CoOp}~\citep{coop} shown as radar plots. \textit{Left}: Base classes, our method (red) consistently yields lower ECE than competing approaches, with notable gains on DTD, EuroSat, and Food. \textit{Right}: Novel classes, our method reduces miscalibration relative to vanilla CoOp (yellow) and outperforms DAC~\citep{dac} and ZS-Norm~\citep{eccv}, especially on Aircraft and Cars. The uniformly smaller footprint of our curve indicates superior calibration, %across the benchmark, 
    supporting the effectiveness of the proposed dual-regularization approach in addressing both underconfidence on base classes and overconfidence on novel classes.}
    \vspace{-1em}
    %\caption{\small \textbf{Radar plots comparing ECE across 11 datasets for different calibration methods with CoOp}~\citep{coop}. \textit{Left}: Performance on base classes shows our method (red) consistently maintains lower ECE values than competing approaches, with particular improvements on DTD, EuroSat, and Food datasets. \textit{Right}: For novel classes, our approach dramatically reduces miscalibration compared to vanilla CoOp (yellow), outperforming state-of-the-art methods including DAC~\citep{dac} and ZS-Norm~\citep{eccv} on challenging datasets like Aircraft and Cars. The consistently smaller footprint of our method's plot demonstrates superior calibration across the entire benchmark suite, confirming the effectiveness of our dual regularization strategy in addressing both underconfidence and overconfidence issues simultaneously {\color{red}Ashs - change yellow color}.
%}
    \label{fig:radar_plot}
\end{figure}
Despite these advances, existing prompt tuning techniques predominantly prioritize accuracy, often neglecting the critical issue of confidence calibration. Miscalibration occurs when a model's predicted confidence poorly aligns with its actual likelihood of correctness, resulting in unreliable uncertainty estimates~\citep{wang2023calibration,guo2017calibration}.
%This reliability gap severely hampers the deployment of prompt-tuned VLMs in safety-critical applications where erroneous high-confidence predictions could lead to harmful outcomes.
This reliability gap poses substantial challenges for the deployment of prompt-tuned VLMs in applications where incorrect high-confidence predictions can have serious consequences, such as autonomous systems that fail to identify obstacles or medical imaging tools that overlook critical abnormalities~\citep{lambert2024trustworthy,shao2024uncertainty}. 
We note that maintaining well-calibrated confidence estimates across both base and novel categories remains largely unexplored, despite being crucial for real-world VLM deployment~\citep{gawlikowski2023survey}.

Only a few recent efforts have explicitly addressed calibration in the context of prompt-tuned CLIP.  DAC~\citep{wang2024open} implements post-hoc temperature scaling for novel classes based on semantic distances between class embeddings. However, this method cannot constrain how prompt tuning alters the original embedding space, allowing problematic transformations like embedding collapse or clustering that introduce spurious semantic relationships. Consequently, the model makes overconfident predictions for novel inputs that fall near distorted decision boundaries. Similarly, \citep{eccv} attempt to normalize output logits to match zero-shot CLIP's distribution characteristics. While this approach adjusts the global statistical properties of model outputs, it often fails to preserve the %local decision boundary structure or maintain proper 
inter-class relationships in the embedding space. This limitation prevents the method from effectively addressing both underconfidence on base classes and overconfidence on novel classes simultaneously.

We address the dual calibration problem in prompt-tuned CLIP with a train-time regularization that preserves pretrained semantics while stabilizing predictive margins. Our approach has two complementary components that jointly target underconfidence on base classes and overconfidence on novel classes. Specifically, our contributions are:

\begin{itemize}
    \item We propose a \textbf{mean-variance margin regularization} that shapes logit distributions by encouraging sufficiently large margins between correct and incorrect predictions while constraining margin variability to prevent spurious confidence spikes. 
    \item We introduce a \textbf{text moment-matching loss} that preserves the geometric structure of CLIP's pretrained embedding space by aligning the statistical moments of tuned and frozen text embeddings. This preserves critical semantic relationships by maintaining relative class structure, without restricting task-specific adaptations. 
    \item We evaluate across 11 diverse datasets and 7 prompt-tuning frameworks, demonstrating consistent improvements in calibration without compromising accuracy, outperforming post-hoc and training-time baselines (see Figure~\ref{fig:radar_plot}).
    
    %Our approach outperforms existing post-hoc and training-time calibration methods (see Fig.~\ref{fig:equivariance-ece}) while remaining computationally efficient, making it readily applicable to a wide range of vision-language applications requiring reliable uncertainty estimates.
\end{itemize}
\textit{Importantly, our method is agnostic to the underlying prompt tuning technique, does not require additional inference time computation and functions as a plug-and-play module for existing frameworks}.

\section{Related Work}

\noindent \textbf{Prompt Tuning for Vision-Language Models.}  
Prompt tuning adapts VLMs like CLIP by learning a small set of text tokens while freezing the image/text encoders, enabling parameter-efficient specialization with minimal supervision.
%Prompt tuning has emerged as a lightweight and effective technique for adapting large vision-language models (VLMs) to downstream tasks without modifying the pretrained image or text encoders. Unlike traditional fine-tuning, which updates all model parameters, prompt tuning introduces a small set of learnable tokens prepended to the input text, allowing models like CLIP to specialize with minimal supervision. 
Early work explored static prompts shared across all inputs~\citep{zhou2022learning}, while subsequent extensions proposed instance-conditioned prompts that adapt dynamically to each image~\citep{zhou2022conditional}. Further advances have incorporated multi-modal prompt learning~\citep{khattak2023maple} and visual context modulation. These methods are typically trained with few-shot supervision on a set of \emph{base} classes and evaluated for zero-shot generalization on \emph{novel} classes, aiming to improve open-vocabulary recognition without fine-tuning the backbone.
\textit{However, these existing methods have been focusing on improving classification accuracy, with little attention paid to the calibration of predicted probabilities}. 

\noindent \textbf{Calibration of Deep Neural Networks.}
% Confidence calibration measures the alignment between a model’s predicted confidence scores and the actual accuracy of its predictions. This alignment is critical in safety-sensitive applications, such as medical diagnosis and autonomous driving, where overconfident errors can have severe consequences~\citep{guo2017calibration,lambert2024trustworthy}. 
Traditional post-hoc calibration methods, such as Temperature Scaling~\citep{guo2017calibration}, calibrate confidence scores by applying a scalar temperature to the model’s logits, typically learned on a held-out validation set. While effective under in-distribution settings, these methods assume access to labeled data from the target domain and often fail to generalize to out-of-distribution scenarios, where such supervision is unavailable~\citep{niculescu2005predicting,wang2024open}. On the other hand, train-time calibration approaches integrate auxiliary loss terms during model training that penalize miscalibrated predictions~\citep{kumar2018trainable,ovadia2019can}, yielding models with more reliable uncertainty estimates. \textit{However, these methods typically require fully labeled datasets and involve fine-tuning the entire model, rendering them unsuitable for settings like prompt tuning, which operate in a few-shot regime and update only a small number of parameters}.

\noindent \textbf{Calibrating Prompt Tuning.} 
Recent work has begun to address calibration in prompt-tuned VLMs. 
DAC~\citep{wang2024open} applies post-hoc temperature scaling based on semantic distances between class embeddings but often degrade the sharp decision boundaries of their pretrained counterparts.
%\citep{wang2024open} proposed Distance-Aware Calibration (DAC), a post-hoc method that scales softmax temperatures based on semantic distances between base and novel class embeddings, but found that fine-tuned VLMs often degrade the sharp decision boundaries of their pretrained counterparts. 
Concurrently, \citep{eccv} tackled calibration within the prompt tuning framework by identifying expanded logit distributions as a key issue and introducing zero-shot normalization and sample-adaptive logit scaling to restore alignment with zero-shot CLIP. 
Methods aiming to improve calibration for test-time prompt tuning \citep{yoon2024c,sharifdeen2025tpt} have aimed to enhance the dispersion of text features, but require additional computation inference.
These efforts reveal key insights: preserving CLIP's embedding geometry is essential for novel class calibration, sufficient class separation helps prevent overconfidence, and post-hoc corrections cannot fully restore pretrained calibration properties. \textit{Our method unifies these insights by directly constraining embedding transformations during training while simultaneously handling the distinct miscalibration patterns of base and novel classes, without compromising semantic relationships.}

%Other approaches aim to improve calibration of test-time prompt optimization: \citep{yoon2024c} related text feature dispersion to calibration performance, while \citep{sharifdeen2025tpt} added orthogonality constraints to improve both reliability and accuracy. Together, these works reveal three insights: (\textbf{1}) preserving CLIP’s embedding geometry is crucial for novel class calibration, (\textbf{2}) sufficient class separation helps prevent overconfidence, and (\textbf{3}) post-hoc methods alone cannot fully recover pretrained calibration behavior. Our method unifies these insights to address miscalibration across both base and novel categories, without adding inference overhead or compromising fine-grained semantic relationships essential for open-vocabulary generalization. 

\section{Method} \label{sec:Method}

\subsection{Preliminaries}

\noindent \textbf{Zero-Shot Inference for CLIP.} 
CLIP enables zero-shot classification by learning a joint embedding space for images and natural language descriptions through large-scale contrastive pretraining. The model comprises an image encoder $\mathbf{E}_{\text{img}}: \mathcal{I} \rightarrow \mathbb{R}^d$ and a text encoder $\mathbf{E}_{\text{txt}}: \mathcal{T} \rightarrow \mathbb{R}^d$, where $\mathcal{I}$ and $\mathcal{T}$ denote the image and text spaces, respectively.
During inference, given an input image $\mathbf{I} \in \mathcal{I}$, the image encoder produces a feature embedding $\mathbf{v} = \mathbf{E}_{\text{img}}(\mathbf{I})$. To perform classification, CLIP compares this visual embedding against textual representations of candidate class labels. Specifically, each class $y_i \in \{y_1, \dots, y_K\}$ is converted into text prompts using a fixed prompt template (e.g., $\mathbf{t}(y_i) = \texttt{"A photo of a \{class\}"}$), and encoded by the text encoder as $\mathbf{u}_i = \mathbf{E}_{\text{txt}}(\mathbf{t}(y_i))$. The similarity between image and text embeddings is computed by cosine similarity as $s_i = \text{cos}(\mathbf{v}, \mathbf{u}_i)$, and the predicted class probabilities are obtained using a temperature-scaled softmax: $\mathbb{P}(y_i \mid \mathbf{I}) = \exp(\tau s_i) / \sum_{j=1}^{K} \exp(\tau s_j)$, where $\tau$ is softmax temperature parameter. The predicted label $\hat{y}$ and its associated confidence $\hat{p}$ are given by $\hat{y} = \arg\max_i \mathbb{P}(y_i \mid \mathbf{I})$ and $\hat{p} = \max_i \mathbb{P}(y_i\mid\mathbf{I})$, respectively.

\noindent \textbf{Prompt Learning for CLIP.} While zero-shot classification with handcrafted templates is effective, it may not provide optimal task-specific context. Prompt learning addresses this by optimizing the prompt tokens directly for downstream performance. 
Instead of using fixed text templates, a set of learnable tokens $\mathcal{T} = \{\mathbf{p}_1, \dots, \mathbf{p}_M\}$ is introduced as a prefix to the class name in each prompt. For any class $y \in \mathcal{Y}$, the composed prompt is defined as $\mathbf{t}(y) = [\mathbf{p}_1, \dots, \mathbf{p}_M, \mathbf{e}_y]$, where $\mathbf{e}_y$ is a static embedding of the class name. The text encoder maps this into a class representation $\mathbf{c}_y = \mathbf{E}_{\text{txt}}(\mathbf{t}(y))$. Given an image $\mathbf{x}$, the logit for class $y$ is computed as $s_y = \tau \cdot \cos(\mathbf{E}_{\text{img}}(\mathbf{x}), \mathbf{c}_y)$, and class probabilities are obtained via softmax as in the zero-shot case. The prompt tokens $\mathcal{T}$ are optimized using a small labeled training set $\mathcal{D} = \{(\mathbf{x}_i, y_i)\}_{i=1}^N$, typically by minimizing the cross-entropy loss over predicted class probabilities. We denote the set of classes seen during prompt tuning as $\mathcal{Y}_\text{base} \subset \mathcal{Y}$, and the remaining classes encountered at test time as $\mathcal{Y}_\text{novel} = \mathcal{Y} \setminus \mathcal{Y}_\text{base}$.

%\noindent \textbf{Prompt Learning for CLIP.} While zero-shot classification with handcrafted templates is effective and broadly applicable, it may not provide optimal task-specific context. Prompt learning addresses this by optimizing the prompt tokens directly for downstream performance. Instead of using fixed text templates, a set of learnable tokens $\mathcal{T} = \{\mathbf{p}_1, \dots, \mathbf{p}_M\}$ is introduced as a prefix to the class name in each prompt. For any class $y \in \mathcal{Y}$, the composed prompt is defined as $\mathbf{t}(y) = [\mathbf{p}_1, \dots, \mathbf{p}_M, \mathbf{e}_y]$, where $\mathbf{e}_y$ is a static embedding of the class name. The text encoder maps this into a class representation $\mathbf{u}_y = \mathbf{E}_{\text{txt}}(\mathbf{t}(y))$. Given an image $\mathbf{x}$, the logit for class $y$ is computed as $s_y = \tau \cdot \cos(\mathbf{E}_{\text{img}}(\mathbf{x}), \mathbf{u}_y)$, and class probabilities are obtained via softmax as in the zero-shot case. The prompt tokens $\mathcal{T}$ are optimized using a small labeled training set $\mathcal{D} = \{(\mathbf{x}_i, y_i)\}_{i=1}^N$, typically by minimizing the cross-entropy loss over predicted class probabilities. We denote the set of classes seen during prompt tuning as $\mathcal{Y}_\text{base} \subset \mathcal{Y}$, and the remaining classes encountered only at test time as $\mathcal{Y}_\text{novel} = \mathcal{Y} \setminus \mathcal{Y}_\text{base}$. 

\subsection{Proposed Method}

\begin{figure*}[t]
    \centering
    \includegraphics[width=0.95\linewidth, trim=10 55 10 35, clip]{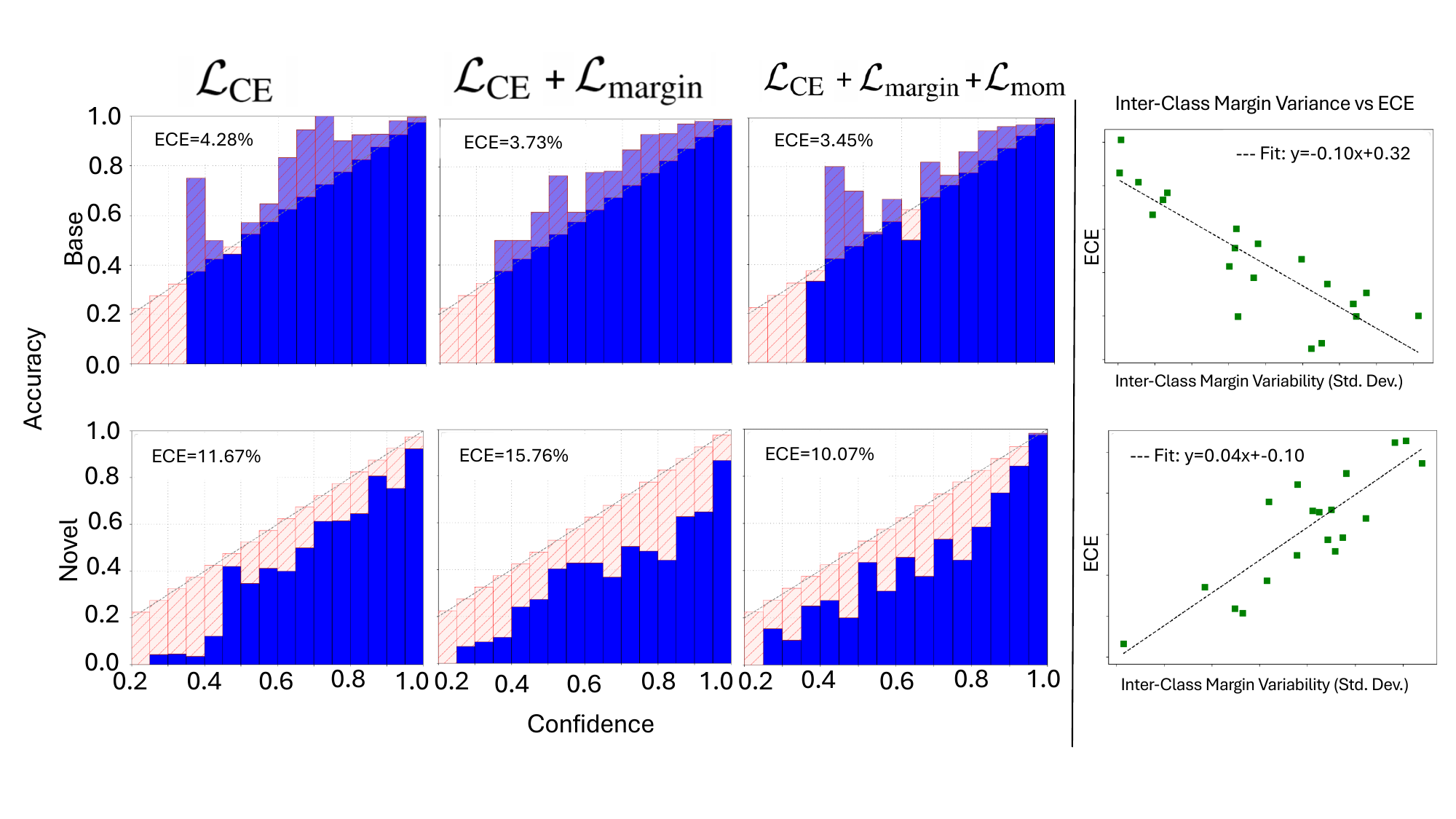}
    \caption{\small \textbf{Dual miscalibration in prompt-tuned CLIP.} \textit{Left (top row)}: Base classes are underconfident (accuracy exceeds confidence) that improves with our regularization terms. \textit{Left (bottom row)}: Novel classes exhibit overconfidence (confidence exceeds accuracy) that our method effectively mitigates. (\textit{Right}) Inter-class margin variability vs.\ ECE shows a \emph{negative} correlation for base classes and a \emph{positive} correlation for novel classes, indicating that prompt tuning tightens margins on base classes and inflates them on novel classes, degrading reliability. These trends motivate our margin-stabilizing and moment-preserving regularizers.
}
\vspace{-0.5em}
    \label{fig:equivariance}
\end{figure*}

Our goal is to improve the reliability of prompt-tuned CLIP by ensuring well-calibrated confidence estimates across \textit{both base and novel classes}. 
This dual calibration challenge arises because prompt tuning introduces underconfidence for base classes (reduced logit margins) and  overconfidence for novel classes(inflated margins)
%This dual calibration challenge arises because prompt tuning introduces asymmetric boundary distortions: reduced logit margins for base classes (causing underconfidence) and inflated margins for novel classes (causing overconfidence).
%This dual calibration problem is particularly challenging, as prompt tuning often introduces asymmetric shifts in decision boundaries, reducing logit margins for base classes, which causes underconfidence, while inflating margins for novel classes, leading to overconfidence. 
Unlike existing methods~\citep{eccv,wang2024open} that address either base class underconfidence or novel class overconfidence in isolation, our approach simultaneously tackles both through: (1) margin-based regularization that encourages more discriminative decision boundaries, and (2) moment-matching loss that preserves CLIP's well-calibrated embedding structure.
%Unlike existing methods~\citep{eccv,wang2024open} that address only one aspect of this dual calibration problem, our approach simultaneously tackles both through: (1) a margin-based regularization term that encourages stable, discriminative decision boundaries, and (2) a moment-matching loss that preserves the well-calibrated semantic structure of the pretrained CLIP embedding space. 
%Next, we begin by analyzing the dual miscalibration phenomena in prompt-tuned CLIP, then introduce our training-time regularization framework that addresses both issues through complementary margin and moment-matching constraints.
Next, we first analyze miscalibration in prompt-tuned CLIP and then present a training-time solution based on margin regularization and moment matching.

\textbf{Dual Miscalibration in Fine-tuned CLIP.} To investigate calibration issues in prompt-tuned CLIP, we systematically analyzed calibration behavior across diverse prompt configurations and datasets. Figure~\ref{fig:equivariance} demonstrates that  dual calibration problem through reliability diagrams and margin analysis.
For \textbf{base classes}, reliability diagrams show underconfidence: predicted probabilities trail behind actual accuracy, reflecting reduced margins between the top-1 and runner-up classes. Larger margins correspond to lower calibration error, confirming the link between boundary tightness and underconfidence. For \textbf{novel classes}, the opposite pattern emerges: predictions are overconfident, with inflated margins driving calibration error upward. The scatter plots in Figure \ref{fig:equivariance} (right) quantify these complementary trends, revealing a negative correlation between margin variability and calibration error for base classes, and a positive correlation for novel classes.

%As shown in Figure~\ref{fig:equivariance}, standard prompt tuning methods like MaPLe exhibit significant calibration errors across both base and novel classes.  For base classes, the reliability diagram shows underconfidence, with the accuracy consistently exceeding confidence levels. Conversely, for novel classes, we observe severe overconfidence, where predicted probabilities are higher than actual accuracies. This calibration disparity directly relates to distorted inter-class margin distributions introduced during prompt tuning. The scatter plots in Figure~\ref{fig:equivariance} (\textit{right}) quantify this relationship, showing a negative correlation between inter-class margin variability and ECE for the base classes, while margin variability positively correlates with ECE for the novel classes.
%\sethlcolor{yellow!40}\hl{ As illustrated in the Reliability diagram of the base classes, it has an underconfident issue. In an underconfident issue, the model assigns a tight logit distance between the top-1 class and the runner-up class.  As you can see in the base class Inter-Class Margin figure, where the margin between the top-1 class and the runner-up class is low, it tends to show higher calibration error. When the margin is increased, the calibration error gets reduced. Similarly, On novel classes, it has an overconfidence issue due to assigning high confidence to incorrect predictions. In the Inter-Class Margin figure for the novel classes, a larger margin between the top-1 class and the runner-up is associated with higher calibration error.} 

These findings indicate that prompt tuning disrupts the naturally well-calibrated decision boundaries of zero-shot CLIP. An ideal calibrated open-vocabulary model should maintain class-appropriate confidence margins across both base and novel categories while preserving the semantic geometry of the pretrained text embedding space. This motivates our framework that explicitly stabilizes decision margins and preserves pretrained semantics.

\subsubsection{Mean-Variance Margin Regularization}  
To address the dual miscalibration in prompt-tuned CLIP, we introduce a mean-variance margin regularization that shapes logit distributions during training.
Our mean-variance loss maintains sufficiently large margins between correct and incorrect predictions to prevent base class underconfidence while enforcing margin consistency across samples to avoid novel class overconfidence.
%To mitigate the dual miscalibration observed in prompt-tuned CLIP models, we introduce a mean-variance margin regularization objective designed to shape the logit distribution during training. Our key insight is that a well-calibrated model should produce sufficiently large logit margins between the top-1 and runner-up predictions to support confident correct predictions, particularly on base classes, while maintaining consistency across examples to avoid spurious overconfidence, especially on novel classes.
Formally, for a batch of samples \(\{(\mathbf{x}_i, y_i)\}_{i=1}^B\), let \(\mathbf{z}_i\) denote the predicted logits for sample \(i\). We define the per-sample logit margin as the difference between the ground-truth class logit and the highest logit among incorrect classes as $m_i = z_{i,y_i} - \max_{j \neq y_i} z_{i,j}$. The \emph{mean-variance margin regularization} loss is then defined as:
\[
\mathcal{L}_{\text{Margin}} = -\alpha \cdot \frac{1}{B} \sum_{i=1}^B m_i + \beta \cdot \operatorname{Var}(m_1, \dots, m_B),
\]
where \(\alpha, \beta > 0\) are hyperparameters that control the trade-off between maximizing the average margin and minimizing its variance across the batch.

The dual design creates complementary objectives that act synergistically: the mean term (weighted by $\alpha$) promotes sufficient separation for confident base class predictions, while the variance term (weighted by $\beta$) prevents margin inconsistency that creates overconfident novel class predictions. Without variance regularization, models may develop erratic decision boundaries with spurious confidence spikes on novel classes. Without mean regularization, uniformly small margins cause systematic underconfidence on base classes, as shown in Figure~\ref{fig:combine_moti}. Unlike prior margin-based approaches such as MBLS~\citep{mbls} that impose hard per-sample constraints, our batch-level statistical approach avoids over-regularization while maintaining adaptability across diverse class distributions, yielding more reliable confidence estimates in open-vocabulary settings.

%This regularizer balances two complementary objectives: the first term (controlled by $\alpha$) promotes high average margins across the batch, encouraging confident correct predictions for base classes. The second term (controlled by $\beta$) penalizes excessive margin variance to prevent overconfident outliers. \sethlcolor{yellow!40}\hl{Without the variance term, the model may collapse to sharp but erratic decision boundaries, undermining calibration on novel classes. Conversely, omitting the mean-margin term can lead to uniformly low margins, resulting in overly cautious predictions.} The two components act in synergy: the mean promotes global separation between classes, while the variance stabilizes margin consistency across examples. Unlike prior margin-based approaches such as MBLS~\citep{mbls}, which impose hard per-sample constraints (e.g., upper bounds via ReLU clipping), our loss operates at the batch level using soft statistical objectives. This design avoids over-regularization and enhances adaptability across class distributions, leading to more reliable confidence estimates and improved generalization in open-vocabulary settings.

\subsubsection{Text Moment-Matching Loss} 

\begin{figure}[t]

    \centering
    \includegraphics[width=\linewidth, trim=10 10 10 32, clip]{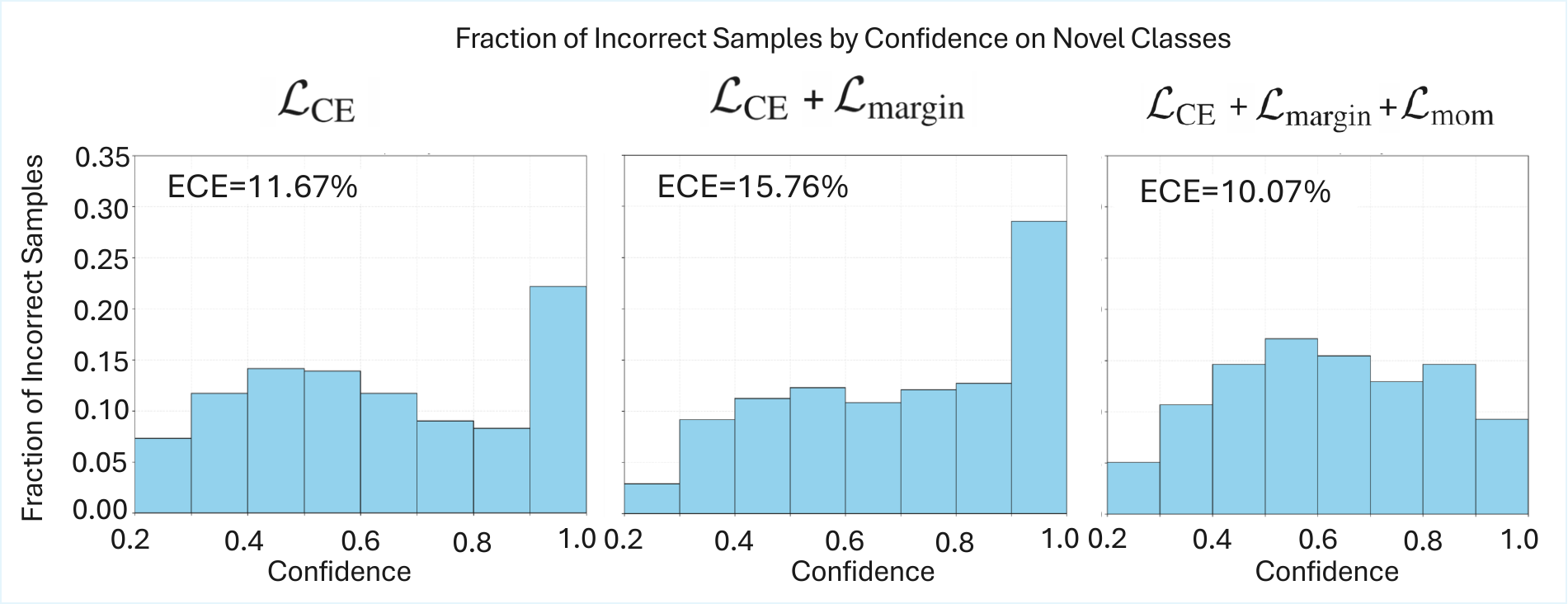}
    \caption{\small \textbf{Errors by confidence on novel classes.} Higher error mass in high-confidence bins indicates overconfidence. Both \textbf{Cross-Entropy} and \textbf{Cross-Entropy + Margin} place more misclassified samples in high-confidence regions, whereas adding \textbf{Text Moment-Matching} to the \textbf{Margin} term shifts errors away from these bins, reducing overconfidence.
    %The Mean-Variance Margin loss, which works in logit space, maintains a consistent margin between the predicted class and the runner-up class. It will help to reduce underconfident while keeping huge margin spikes consistent through variance term. Meanwhile, the loss, which works in embedding space, helps to generalize to unseen classes by maintaining a discriminative relationship between different classes. Cross-Entropy only with Mean-Variance Margin Regularization will enhance the inter-class separation but it may have chances to increase the inter-class distances of incorrectly predicted samples. It will further promote the overconfidence issue when assigning higher confidence to incorrectly predicted samples. as you can see in the Novel Class, It has a high calibration error with a higher fraction of incorrectly predicted samples with high confidence. The combination of Mean-Variance Margin and Text Moment-Matching will help to maintain proper class decision boundaries and reduce the chances for incorrect predictions with high confidence while increasing the margin between the correct class and the runner-up class. This joint objective will help to address the overconfidence issue if the model assigns high confidence for incorrect predictions; meanwhile, having a consistent margin separation will reduce the underconfident issue.
}
    \label{fig:text_mom_motivation}
    %\vspace{-1.5em}
\end{figure}

While the margin regularizer shapes confidence near decision boundaries, it operates on logits and does not directly constrain the geometry of the text embedding space. Empirically (Figure.~\ref{fig:equivariance}), using the margin term alone can \emph{increase} ECE on novel classes: when the top-1 prediction is incorrect, maximizing the margin widens the gap to the runner-up for the \emph{wrong} class, amplifying overconfidence.
To maintain calibrated generalization on novel classes, we therefore propose to preserve the global semantic relationships between class text embeddings. To this end, we introduce a \emph{text moment-matching} objective that aligns the statistical properties of the prompt-tuned text embeddings with those of frozen CLIP.
Let $\{\tilde{\mathbf{c}}_y\}_{y \in \mathcal{B}}$ denote the prompt-tuned text embeddings for a batch of classes $\mathcal{B} \subset \mathcal{Y}_{\text{base}}$, and $\{\mathbf{c}_y^0\}_{y \in \mathcal{B}}$ the corresponding frozen (zero-shot) embeddings. We match both the first- and second-order moments of these sets:
\begin{subequations}
\label{eq:moments}
\begin{align}
\mu_{\tilde c} &= \tfrac{1}{|\mathcal B|} \sum_{y\in\mathcal B} \tilde{\mathbf c}_y, &\\
\mu_{c^0} &= \tfrac{1}{|\mathcal B|} \sum_{y\in\mathcal B} \mathbf c^{0}_y, \\
\Sigma_{\tilde c}
&= \tfrac{1}{|\mathcal B|} \sum_{y\in\mathcal B}
\bigl(\tilde{\mathbf c}_y - \mu_{\tilde c}\bigr)
\bigl(\tilde{\mathbf c}_y - \mu_{\tilde c}\bigr)^{\!\top}, \\
\Sigma_{c^0}
&= \tfrac{1}{|\mathcal B|} \sum_{y\in\mathcal B}
\bigl(\mathbf c^{0}_y - \mu_{c^0}\bigr)
\bigl(\mathbf c^{0}_y - \mu_{c^0}\bigr)^{\!\top}.
\end{align}
\end{subequations}
%Here, $\mu_{\tilde{c}}$ and $\mu_{c^0}$ are the empirical means (first moments) of the tuned and frozen class embeddings, respectively, while $\Sigma_{\tilde{c}}$ and $\Sigma_{c^0}$ are the corresponding empirical covariance matrices (second moments). 
The moment-matching loss constrains both distributional center and spread:
\begin{align}
\mathcal{L}_{\mathrm{mom}} = \| \mu_{\tilde{c}} - \mu_{c^0} \|_2^2 + \| \Sigma_{\tilde{c}} - \Sigma_{c^0} \|_F^2.
\end{align}

\noindent Minimizing $\mathcal{L}_{\mathrm{mom}}$ preserves the semantic center and dispersion of the frozen CLIP space, which supports generalization and curbs high-confidence errors introduced by prompt-induced drift. Unlike direct $\ell_1/\ell_2$ alignment, which forces rigid instance-level correspondence and can hinder adaptation, moment matching constrains only global distributional statistics, keeping local task-specific prompt adjustments expressive.
%Minimizing $\mathcal{L}_{\mathrm{mom}}$ encourages the tuned embeddings to preserve both the semantic center and dispersion of the frozen CLIP space, thereby maintaining generalization and mitigating high-confidence mispredictions caused by prompt-induced drift. This formulation offers key advantages over simpler alternatives such as direct $\ell_1$/$\ell_2$ alignment between tuned and frozen embeddings. Such methods enforce strict instance-level correspondence, potentially suppressing the flexibility needed for task adaptation. Our moment-matching loss constrains only the global distributional statistics, allowing local prompt adaptations to remain expressive. The mean alignment term maintains stability in the overall class embedding distribution, while covariance matching preserves relative dispersion, orientation, and inter-class relationships originally learned by CLIP’s pertaining. 

\textbf{Complementarity with the margin loss.} 
The margin regularizer (logit space) increases inter-class separation but, when the top-1 class is incorrect, can enlarge the wrong gap and worsen novel-class overconfidence. Moment loss (embedding space) counterbalances this by preserving CLIP’s semantic geometry, maintaining relative class structure and angular spread, thus curbing such failure modes. Together, the two terms act synergistically: the margin term enforces discriminability; the moment term stabilizes geometry, yielding calibrated decision boundaries for both base and novel classes without sacrificing downstream performance (see Figure~\ref{fig:text_mom_motivation}).
Finally, we combine the two regularizers with cross-entropy to form the full objective:
\begin{align}
\mathcal{L}_{\text{total}}
= \mathcal{L}_{\text{CE}}
+ \lambda_{\text{Margin}}\,\mathcal{L}_{\text{Margin}}
+ \lambda_{\text{mom}}\,\mathcal{L}_{\text{mom}},
\end{align}
where $\lambda_{\text{Margin}}, \lambda_{\text{mom}}\!\ge\!0$ controls the strength of each term.
This joint optimization addresses both aspects of the dual calibration problem: reducing underconfidence in base classes and overconfidence in novel classes while preserving task-specific adaptation. The terms are complementary: the margin loss enforces discriminability and the moment loss preserves semantic geometry, allowing prompt-tuned models to inherit the well-calibrated behavior of zero-shot CLIP.

\section{Experiments} \label{sec:experiments}

\begin{table*}[t]
%\small
\centering
\caption {\small  \textbf{ Accuracy and calibration performance on base classes across 11 fine-grained classification benchmarks}.
Top-1 accuracy (Acc) and Expected Calibration Error (ECE) for multiple prompt-tuning strategies and diverse calibration baselines. Higher Acc. indicates better classification performance, while lower ECE reflects better calibration.}
\label{table:main_result}
\resizebox{0.9\textwidth}{!}{ 
\setlength{\tabcolsep}{4pt} % Reduce column spacing
\begin{tabular}{l c ||c c c c c c c c c c c c}
\toprule
\thead{\textbf{Method}} & &  \multicolumn{1}{c}{\thead{\rotatebox{90}{INet}}} & \multicolumn{1}{c}{\thead{\rotatebox{90}{Calt}}} & \multicolumn{1}{c}{\thead{\rotatebox{90}{Pets}}} & \multicolumn{1}{c}{\thead{\rotatebox{90}{Cars}}} & \multicolumn{1}{c}{\thead{\rotatebox{90}{Flow}}} & \multicolumn{1}{c}{\thead{\rotatebox{90}{Food}}} & \multicolumn{1}{c}{\thead{\rotatebox{90}{Air}}} & \multicolumn{1}{c}{\thead{\rotatebox{90}{SUN}}} & \multicolumn{1}{c}{\thead{\rotatebox{90}{DTD}}} & \multicolumn{1}{c}{\thead{\rotatebox{90}{Euro}}} & \multicolumn{1}{c}{\thead{\rotatebox{90}{UCF}}} & \multicolumn{1}{c}{\thead{\rotatebox{90}{\textbf{Avg}}}}\\
\midrule
\multirow{2}{*}{$\text{Zero Shot}$}
 & Acc. & 72.40 & 97.20 & 91.30 & 63.60 & 71.80 & 90.10 & 27.70 & 69.40 & 53.00 & 57.00 & 71.00 & 69.50 \\
 & ECE & 1.51 & 6.49 & 2.25 & 3.74 & 3.11 & 1.57 & 3.03 & 1.59 & 4.53 & 8.35 & 3.24 & 3.58 \\
\hline
\rowcolor{orange!20} \multicolumn{14}{c}{\textbf{CoOp}~\citep{coop} } \\ 
\hline
\multirow{2}{*}{$\text{CoOp}$ \citep{coop}} 
 & Acc. & 75.60 & 97.98 & 94.77 & 76.22 & 90.00 & 90.20 & 35.23 & 81.14 & 76.27 & 90.24 & 83.32 & 81.00 \\
 & ECE & 1.65 & 0.66 & 1.00 & 3.73 & 4.93 & 3.66 & 25.70 & 8.11 & 12.17 & 1.75 & 6.44 & 6.35 \\
\hdashline[2pt/2pt]
\multirow{2}{*}{$\text{MBLS}$~\citep{mbls}} 
 & Acc. & 75.12 & 97.89 & 91.11 & 76.21 & 89.34 & 89.78 & 34.32 & 81.32 & 76.34 & 90.12 & 82.78 & 80.39 \\
  & ECE & 2.98 & 9.6 & 7.70 & 12.20 & 5.69 & 12.34 & 10.48 & 16.80 & 4.25 & 8.02 & 9.39 & 9.04 \\
\hdashline[2pt/2pt]
\multirow{2}{*}{$\text{Temp. Scal.}$ \citep{guo2017calibration}}
 & Acc. & 75.60 & 98.19 & 94.15 & 78.65 & 97.72 & 90.10 & 42.00 & 81.32 & 80.67 & 90.70 & 84.56 & 83.06 \\
  & ECE & 1.50 & 1.20 & 2.54 & 6.63 & 4.60 & 0.50 & 3.43 & 2.01 & 3.86 & 4.76 & 1.57 & 2.96 \\
\hdashline[2pt/2pt]
\multirow{2}{*}{$\text{DAC}$ \citep{dac}}
 & Acc. & - & - & - & - & - & - & - & - & - & - & - & - \\
  & ECE & - & - & - & - & - & - & - & - & - & - & - & - \\
\hdashline[2pt/2pt]
\multirow{2}{*}{$\text{$\text{ZS-Norm}$}$ \citep{eccv}}
 & Acc. & 76.10 & 97.85 & 94.38 & 77.78 & 95.76 & 89.52 & 39.74 & 81.37 & 81.02 & 90.45 & 84.01 & 82.54 \\
 & ECE & 3.15 & 4.35 & 7.75 & 11.30 & 11.29 & 3.14 & 13.05 & 4.22 & 49.53 & 37.04 & 3.47 & 13.48 \\
 \hdashline[2pt/2pt]
\multirow{2}{*}{$\text{$\text{Penalty}$}$ \citep{eccv}}
 & Acc. & 76.44 & 97.72 & 95.11 & 77.05 & 96.30 & 87.92 & 38.07 & 81.04 & 77.32 & 47.09 & 80.47 & 77.68 \\
 & ECE & 2.43 & 4.79 & 6.47 & 10.01 & 9.38 & 5.98 & 8.59 & 4.59 & 21.84 & 20.47 & 7.42 & 9.27 \\
  \hdashline[2pt/2pt]
 \multirow{2}{*}{$\textbf{Ours}$}
 & Acc. & \cellcolor{green!20}76.53 & \cellcolor{green!20}98.06 & \cellcolor{green!20}94.95 & \cellcolor{green!20}77.32 & \cellcolor{green!20}97.21 & \cellcolor{green!20}90.38 & \cellcolor{green!20}38.62 & \cellcolor{green!20}81.68 & \cellcolor{green!20}80.44 & \cellcolor{green!20}88.56 & \cellcolor{green!20}84.68 & \cellcolor{green!20}82.58 \\
 & ECE & \cellcolor{green!20}2.47 & \cellcolor{green!20}1.01 & \cellcolor{green!20}1.94 & \cellcolor{green!20}7.10 & \cellcolor{green!20}4.80 & \cellcolor{green!20}0.30 & \cellcolor{green!20}4.96 & \cellcolor{green!20}1.22 & \cellcolor{green!20}2.42 & \cellcolor{green!20}4.90 & \cellcolor{green!20}1.11 & \cellcolor{green!20}2.93 \\

\hline
\rowcolor{orange!20} \multicolumn{14}{c}{\textbf{MaPLe} \citep{maple}} \\ 
\hline
\multirow{2}{*}{$\text{MaPLe}$ \citep{maple}}
 & Acc. & 76.71 & 97.97 & 95.53 & 72.93 & 96.00 & 90.80 & 36.33 & 80.55 & 79.63 & 91.13 & 83.20 & 82.41\\
 & ECE & 2.27 & 1.54 & 2.68 & 7.25 & 4.28 & 0.78 & 3.86 & 1.27 & 4.18 & 3.42 & 2.68 & 3.19\\
\hdashline[2pt/2pt]
\multirow{2}{*}{$\text{MBLS}$ \citep{mbls}}
 & Acc. & 75.59 & 98.23 & 95.23 & 72.77 & 95.93 & 90.80 & 36.20 & 80.73 & 80.03 & 90.93 & 84.13 & 82.50\\
  & ECE & 29.06 & 5.03 & 6.64 & 19.06 & 12.74 & 6.55 & 5.60 & 11.01 & 4.79 & 3.73 & 8.46 & 8.36\\
\hdashline[2pt/2pt]
\multirow{2}{*}{$\text{Temp. Scal.} $ \citep{guo2017calibration}}
 & Acc. & 76.66 & 97.97 & 94.93 & 72.70 & 95.93 & 90.63 & 36.37 & 80.73 & 78.60 & 93.60 & 84.00 & 82.55\\
  & ECE & 2.37 & 1.26 & 2.28 & 4.96 & 3.44 & 0.71 & 3.04 & 2.84 & 5.98 & 1.31 & 3.07 & 2.89\\
\hdashline[2pt/2pt]
\multirow{2}{*}{$\text{DAC}$ \citep{dac}}
 & Acc. & - & - & - & - & - & - & - & - & - & - & - & -\\
  & ECE & - & - & - & - & - & - & - & - & - & - & - & -\\
\hdashline[2pt/2pt]
\multirow{2}{*}{$\text{$\text{ZS-Norm}$}$ \citep{eccv}}
 & Acc. & 76.63 & 97.57 & 95.70 & 73.07 & 95.63 & 90.57 & 36.00 & 80.97 & 80.43 & 91.30 & 83.87 & 82.51 \\
 & ECE & 1.64 & 23.30 & 5.91 & 8.66 & 11.49 & 1.13 & 7.87 & 2.33 & 7.02 & 19.38 & 3.86 & 9.10 \\
 \hdashline[2pt/2pt]
\multirow{2}{*}{$\text{$\text{Penalty}$}$ \citep{eccv}}
 & Acc. & 76.72 & 98.07 & 95.30 & 72.43 & 95.77 & 90.73 & 34.33 & 80.93 & 64.60 & 36.77 & 83.03 & 75.20 \\
 & ECE & 3.87 & 5.41 & 6.37 & 13.53 & 12.67 & 3.87 & 8.42 & 7.28 & 19.97 & 13.43 & 8.50 & 9.95 \\
  \hdashline[2pt/2pt]
 \multirow{2}{*}{$\textbf{Ours}$}
 & Acc. & \cellcolor{green!20}76.72 & \cellcolor{green!20}97.97 & \cellcolor{green!20}94.93 & \cellcolor{green!20}72.80 & \cellcolor{green!20}96.20 & \cellcolor{green!20}90.43 & \cellcolor{green!20}36.80 & \cellcolor{green!20}81.10 & \cellcolor{green!20}80.73 & \cellcolor{green!20}92.00 & \cellcolor{green!20}84.50 & \cellcolor{green!20}82.75 \\
 & ECE & \cellcolor{green!20}2.39 & \cellcolor{green!20}1.19 & \cellcolor{green!20}1.54 & \cellcolor{green!20}7.92 & \cellcolor{green!20}3.45 & \cellcolor{green!20}0.65 & \cellcolor{green!20}4.50 & \cellcolor{green!20}1.55 & \cellcolor{green!20}3.56 & \cellcolor{green!20}1.33 & \cellcolor{green!20}2.12 & \cellcolor{green!20}2.78 \\

\hline
\rowcolor{orange!20} \multicolumn{14}{c}{\textbf{KGCoOp} \citep{kgcoop}} \\ 
\hline
\multirow{2}{*}{$\text{KGCoOp}$ \citep{kgcoop}}
 & Acc. & 75.75 & 97.70 & 94.68 & 72.70 & 95.16 & 90.57 & 36.77 & 80.59 & 79.40 & 86.14 & 83.51 & 81.18\\
 & ECE & 2.52 & 2.92 & 3.27 & 10.16 & 12.12 & 1.68 & 3.27 & 4.92 & 8.39 & 11.90 & 5.03 & 6.02\\
\hdashline[2pt/2pt]
\multirow{2}{*}{$\text{MBLS}$ \citep{mbls}}
 & Acc. & 76.23 & 97.81 & 95.00 & 75.34 & 96.24 & 90.49 & 38.28 & 80.86  & 79.94 & 87.96 & 83.45 & 81.96 \\
  & ECE & 6.19 &4.30 & 5.26 & 13.43 & 12.48 & 4.08 & 8.01 & 8.16 & 9.03 & 11.97 & 5.86 & 8.07 \\
\hdashline[2pt/2pt]
\multirow{2}{*}{$\text{Temp. Scal.}$ \citep{guo2017calibration}}
 & Acc. & 75.77 & 97.66 & 94.67 & 70.08 & 94.65 & 90.50 & 35.81 & 80.51 & 78.74 & 86.44 & 83.32 & 80.74\\
  & ECE & 6.47 & 4.16 & 5.13 & 11.70 & 15.35 & 3.64 & 7.41 & 8.50 & 11.12 & 15.79 & 7.39 & 8.79\\
\hdashline[2pt/2pt]
\multirow{2}{*}{$\text{DAC}$ \citep{dac}}
 & Acc. & - & - & - & - & - & - & - & - & - & - & - & -\\
  & ECE & - & - & - & - & - & - & - & - & - & - & - & -\\
\hdashline[2pt/2pt]
\multirow{2}{*}{$\text{$\text{ZS-Norm}$}$ \citep{eccv}}
 & Acc. & 75.78 & 94.14 & 97.65 & 74.55 & 73.90 & 91.71 & 30.79 & 76.50 & 51.49 & 65.39 & 76.44 & 73.49 \\
 & ECE & 2.70 & 1.65 & 3.51 & 3.85 & 4.72 & 2.20 & 8.42 & 3.23 & 6.37 & 6.16 & 3.83 & 4.24 \\
 \hdashline[2pt/2pt]
\multirow{2}{*}{$\text{$\text{Penalty}$}$ \citep{eccv}}
 & Acc. & 75.65 & 97.70 & 94.68 & 72.45 & 93.86 & 90.59 & 37.76 & 80.63 & 78.40 & 83.09 & 82.97  & 80.71 \\
 & ECE & 2.73 & 3.27 & 3.22 & 10.58 & 13.01 & 1.73 & 9.59 & 6.51 & 20.40 & 6.51 & 6.07 & 7.57 \\
  \hdashline[2pt/2pt]
\multirow{2}{*}{$\textbf{Ours}$}
 & Acc. & \cellcolor{green!20}75.84 & \cellcolor{green!20}97.68 & \cellcolor{green!20}94.84 & \cellcolor{green!20}71.65 & \cellcolor{green!20}95.22 & \cellcolor{green!20}90.52 & \cellcolor{green!20}36.03 & \cellcolor{green!20}80.70 & \cellcolor{green!20}78.47 & \cellcolor{green!20}85.10 & \cellcolor{green!20}83.16 & \cellcolor{green!20}80.34 \\
 & ECE & \cellcolor{green!20}2.14 & \cellcolor{green!20}1.88 & \cellcolor{green!20}2.96 & \cellcolor{green!20}8.10 & \cellcolor{green!20}11.21 & \cellcolor{green!20}1.12 & \cellcolor{green!20}4.81 & \cellcolor{green!20}4.12 & \cellcolor{green!20}7.01 & \cellcolor{green!20}12.64 & \cellcolor{green!20}4.14 & \cellcolor{green!20}5.47 \\

\bottomrule
\end{tabular}
 }
\end{table*}

\begin{table*}[t]
\centering
\caption {\small  \textbf{ Accuracy and calibration performance on novel classes across 11 fine-grained classification benchmarks}.
We report top-1 accuracy (Acc) and Expected Calibration Error (ECE) for multiple prompt-tuning strategies and diverse calibration baselines.} %Higher Acc. indicates better classification performance, while lower ECE reflects better calibration.}
\label{table:novel_classes}
\resizebox{0.9\textwidth}{!}{ 
\setlength{\tabcolsep}{4pt} % Reduce column spacing
\begin{tabular}{l c ||c c c c c c c c c c c c}
\toprule
\thead{\textbf{Method}} & &  \multicolumn{1}{c}{\thead{\rotatebox{90}{INet}}} & \multicolumn{1}{c}{\thead{\rotatebox{90}{Calt}}} & \multicolumn{1}{c}{\thead{\rotatebox{90}{Pets}}} & \multicolumn{1}{c}{\thead{\rotatebox{90}{Cars}}} & \multicolumn{1}{c}{\thead{\rotatebox{90}{Flow}}} & \multicolumn{1}{c}{\thead{\rotatebox{90}{Food}}} & \multicolumn{1}{c}{\thead{\rotatebox{90}{Air}}} & \multicolumn{1}{c}{\thead{\rotatebox{90}{SUN}}} & \multicolumn{1}{c}{\thead{\rotatebox{90}{DTD}}} & \multicolumn{1}{c}{\thead{\rotatebox{90}{Euro}}} & \multicolumn{1}{c}{\thead{\rotatebox{90}{UCF}}} & \multicolumn{1}{c}{\thead{\rotatebox{90}{\textbf{Avg}}}}\\
\midrule
\multirow{2}{*}{$\text{Zero Shot}$}
 & Acc. & 72.40 & 94.10 & 97.10 & 75.00 & 77.50 & 91.10 & 35.90 & 75.50 & 60.60 & 63.80  & 78.60 & 74.30 \\
 & ECE & 2.09 & 1.55 & 3.42 & 3.31 & 4.91 & 1.83 & 6.55 & 3.48 & 6.86 & 9.12 & 5.52 & 4.43 \\
\hline
\rowcolor{orange!20} \multicolumn{14}{c}{\textbf{CoOp}~\citep{coop}} \\ 
\hline
\multirow{2}{*}{$\text{CoOp}$ \citep{coop}}
 & Acc. & 59.07 & 94.18 & 96.49 & 65.29 & 69.90 & 90.57 & 24.79 & 70.77 & 52.98 & 64.68 & 62.83 & 68.32 \\
 & ECE & 10.69 & 2.16 & 1.67 & 11.73 & 12.13 & 3.03 & 30.44 & 13.70 & 20.82 & 11.88 & 18.74 & 12.45 \\
\hdashline[2pt/2pt]
\multirow{2}{*}{$\text{MBLS}$ \citep{mbls}}
 & Acc. & 59.11 & 95.1 & 96.23 & 65.28 & 69.89 & 90.23 & 24.80 & 70.12 & 53.12 & 64.65 & 62.97 & 68.31 \\
  & ECE & 4.09 & 2.21 & 3.45 & 9.7 & 18.9 & 13.8 & 10.2 & 9.7 & 8.9 & 12.1 & 13.21 & 9.66 \\
\hdashline[2pt/2pt]
\multirow{2}{*}{$\text{Temp. Scaling}$ \citep{guo2017calibration}}
 & Acc. & 59.07 & 93.45 & 96.03 & 66.70 & 65.86 & 96.60 & 27.37 & 70.67 & 48.19 & 54.70 & 57.51 & 66.92 \\
  & ECE & 7.33 & 3.17 & 3.65 & 5.01 & 8.06 & 1.06 & 18.80 & 6.93 & 20.21 & 15.13 & 14.55 & 9.45 \\
\hdashline[2pt/2pt]
\multirow{2}{*}{$\text{DAC}$ \citep{dac}}
 & Acc. & - & - & - & - & - & - & - & - & - & - & - & - \\
  & ECE & 5.67 & 3.17 & 1.82 & 5.16 & 10.19 & 1.78 & 17.38 & 4.05 & 10.48 & 8.62 & 8.67 & 7.00 \\
\hdashline[2pt/2pt]
\multirow{2}{*}{$\text{ZS-Norm}$ \citep{eccv}}
 & Acc. & 66.26 & 93.30 & 93.98 & 66.62 & 67.21 & 88.91 & 25.76 & 70.51 & 44.08 & 50.42 & 62.52 & 66.32 \\
 & ECE & 2.46 & 2.89 & 7.94 & 2.87 & 4.41 & 3.32 & 10.18 & 2.47 & 21.80 & 15.93 & 4.28 & 7.14 \\
 \hdashline[2pt/2pt]
\multirow{2}{*}{$\text{Penalty}$ \citep{eccv}}
 & Acc. & 66.71 & 92.87 & 96.14 & 68.11 & 68.65 & 78.34 & 29.29 & 71.65 & 40.78 & 41.44 & 67.53 & 65.59 \\
 & ECE & 2.36 & 2.52 & 7.42 & 2.73 & 4.93 & 4.70 & 7.81 & 2.79 & 4.20 & 13.11 & 4.66 & 5.20 \\
  \hdashline[2pt/2pt]
 \multirow{2}{*}{$\text{Ours}$}
 & Acc. & \cellcolor{green!20}67.03 & \cellcolor{green!20}93.56 & \cellcolor{green!20}97.36 & \cellcolor{green!20}69.49 & \cellcolor{green!20}71.63 & \cellcolor{green!20}90.84 & \cellcolor{green!20}30.83 & \cellcolor{green!20}70.03 & \cellcolor{green!20}48.07 & \cellcolor{green!20}56.70 & \cellcolor{green!20}66.49 & \cellcolor{green!20}69.28 \\
 & ECE & \cellcolor{green!20}2.02 & \cellcolor{green!20}2.21 & \cellcolor{green!20}3.03 & \cellcolor{green!20}2.10 & \cellcolor{green!20}3.51 & \cellcolor{green!20}0.87 & \cellcolor{green!20}10.64 & \cellcolor{green!20}3.08 & \cellcolor{green!20}9.31 & \cellcolor{green!20}11.15 & \cellcolor{green!20}4.75 & \cellcolor{green!20}4.79 \\

 \hline
\rowcolor{orange!20} \multicolumn{14}{c}{\textbf{MaPLe} \citep{maple}} \\ 
\hline
\multirow{2}{*}{$\text{MaPLe}$ \citep{maple}}
 & Acc. & 70.50 & 95.10 & 97.85 & 73.57 & 72.80 & 92.10 & 34.53 & 78.20 & 58.47 & 75.90 & 77.85 & 75.17\\
 & ECE & 1.93 & 1.62 & 2.63 & 3.09 & 11.67 & 1.19 & 11.24 & 2.21 & 12.16 & 11.68 & 3.98 & 5.76\\
\hdashline[2pt/2pt]
\multirow{2}{*}{$\text{MBLS}$ \citep{mbls}}
 & Acc. & 68.47 & 94.17 & 96.97 & 71.93 & 68.93 & 91.40 & 33.77 & 78.10 & 54.70 & 75.97 & 78.23 & 73.88\\
  & ECE & 22.82 & 4.06 & 7.41 & 11.41 & 4.84 & 7.06 & 6.06 & 10.41 & 10.31 & 11.25 & 6.63 & 9.30\\
\hdashline[2pt/2pt]
\multirow{2}{*}{$\text{Temp. Scaling}$ \citep{guo2017calibration}}
 & Acc. & 70.46 & 94.83 & 97.30 & 73.47 & 72.77 & 91.77 & 34.07 & 78.13 & 57.97 & 73.77 & 75.33 & 74.53\\
  & ECE & 1.95 & 2.56 & 2.13 & 4.08 & 12.76 & 0.72 & 19.11 & 5.09 & 16.47 & 8.05 & 7.13 & 7.28\\
\hdashline[2pt/2pt]
\multirow{2}{*}{$\text{DAC}$ \citep{dac}}
 & Acc. & - & - & - & - & - & - & - & - & - & - & - & -\\
  & ECE & 2.11 & 1.26 & 2.51 & 2.75 & 11.28 & 1.50 & 9.06 & 1.22 & 8.16 & 8.55 & 2.30 & 4.61\\
\hdashline[2pt/2pt]
\multirow{2}{*}{$\text{ZS-Norm}$ \citep{eccv}}
 & Acc. & 70.63 & 90.30 & 97.23 & 73.30 & 70.03 & 91.83 & 34.07 & 78.47 & 60.70 & 68.13 & 77.80 & 73.86 \\
 & ECE & 3.67 & 23.02 & 5.00 & 3.26 & 6.05 & 1.62 & 7.82 & 2.65 & 5.23 & 14.53 & 3.33 & 6.93 \\
 \hdashline[2pt/2pt]
\multirow{2}{*}{$\text{Penalty}$ \citep{eccv}}
 & Acc. & 70.66 & 93.60 & 97.33 & 73.90 & 70.87 & 91.90 & 34.70 & 78.67 & 45.47 & 36.77 & 76.83 & 70.06 \\
 & ECE & 1.49 & 3.25 & 6.23 & 5.94 & 5.76 & 4.27 & 4.92 & 5.70 & 8.47 & 13.35 & 6.07 & 5.95 \\
  \hdashline[2pt/2pt]
 \multirow{2}{*}{$\text{Ours}$}
 & Acc. & \cellcolor{green!20}70.28 & \cellcolor{green!20}94.87 & \cellcolor{green!20}97.57 & \cellcolor{green!20}75.27 & \cellcolor{green!20}73.87 & \cellcolor{green!20}91.77 & \cellcolor{green!20}36.03 & \cellcolor{green!20}78.13 & \cellcolor{green!20}61.27 & \cellcolor{green!20}67.60 & \cellcolor{green!20}79.87 & \cellcolor{green!20}75.14 \\
 & ECE & \cellcolor{green!20}1.74 & \cellcolor{green!20}1.42 & \cellcolor{green!20}2.29 & \cellcolor{green!20}2.60 & \cellcolor{green!20}10.07 & \cellcolor{green!20}0.86 & \cellcolor{green!20}8.33 & \cellcolor{green!20}1.11 & \cellcolor{green!20}7.37 & \cellcolor{green!20}7.45 & \cellcolor{green!20}3.32 & \cellcolor{green!20}4.23 \\

 \hline
\rowcolor{orange!20} \multicolumn{14}{c}{\textbf{KGCoOp} \citep{kgcoop}} \\ 
\hline
\multirow{2}{*}{$\text{KGCoOp}$ \citep{kgcoop}} 
 & Acc. & 69.70  & 94.43 & 97.67 & 74.25 & 75.10 & 91.65 & 36.77 & 76.33 & 54.23 & 64.68 & 75.59 & 73.67\\
 & ECE & 1.84 & 1.71 & 3.42 & 3.36 & 5.03 & 2.04 & 6.06 & 1.66 & 4.38 & 8.67 & 2.65 & 3.71\\
\hdashline[2pt/2pt]
\multirow{2}{*}{$\text{MBLS}$ \citep{mbls}}
 & Acc. & 69.14 & 94.32 & 94.24 & 73.01 & 73.90 & 90.49 & 28.87 & 75.75 & 56.28 & 64.27 & 73.84 & 72.19  \\
  & ECE & 4.60 & 1.62 & 3.16 & 3.95 & 4.00 & 4.00 & 11.39 & 5.56 & 3.23 & 5.30 & 4.10 & 4.63 \\
\hdashline[2pt/2pt]
\multirow{2}{*}{$\text{Temp. Scaling}$ \citep{guo2017calibration}}
 & Acc. & 69.79 & 94.54 & 97.56 & 74.94 & 75.37 & 91.66 & 32.35 & 76.79 & 53.83 & 62.17 & 76.91 & 73.27\\
  & ECE & 5.81 & 1.89 & 4.91 & 6.35 & 4.63 & 4.02 & 5.40 & 6.18 & 3.83 & 7.60 & 6.43 & 5.18\\
\hdashline[2pt/2pt]
\multirow{2}{*}{$\text{DAC}$ \citep{dac}}
 & Acc. & - & - & - & - & - & - & - & - & - & - & - & -\\
  & ECE & 4.32 & 1.84 & 3.11 & 3.12 & 5.90 & 1.94 & 11.78 & 1.67 & 7.09 & 6.59 & 2.69 & 4.47\\
\hdashline[2pt/2pt]
\multirow{2}{*}{$\text{ZS-Norm}$ \citep{eccv}}
 & Acc. & 69.68 & 94.14 & 97.65 & 74.55 & 73.90 & 91.71 & 30.79 & 76.50 & 51.49 & 65.39 & 76.44 & 72.19 \\
 & ECE & 1.80 & 1.65 & 3.51 & 3.85 & 4.72 & 2.20 & 8.42 & 3.23 & 6.37 & 6.16 & 3.83 & 4.16 \\
 \hdashline[2pt/2pt]
\multirow{2}{*}{$\text{Penalty}$ \citep{eccv}}
 & Acc. & 69.58 & 94.34 & 96.35 & 74.75 & 73.21 & 91.31 & 30.58 & 76.69 & 51.19 & 65.43 & 76.52 & 72.99 \\
 & ECE & 1.82 & 1.71 & 4.21 & 3.05 & 5.12 & 2.99 & 8.12 & 4.13 & 5.87 & 6.56 & 3.93 & 4.75   \\
  \hdashline[2pt/2pt]
\multirow{2}{*}{$\text{Ours}$}
 & Acc. & \cellcolor{green!20}69.50 & \cellcolor{green!20}94.21 & \cellcolor{green!20}97.72 & \cellcolor{green!20}74.39 & \cellcolor{green!20}73.80 & \cellcolor{green!20}91.64 & \cellcolor{green!20}31.63 & \cellcolor{green!20}76.30 & \cellcolor{green!20}55.92 & \cellcolor{green!20}65.76 & \cellcolor{green!20}76.62 & \cellcolor{green!20}73.41 \\
 & ECE & \cellcolor{green!20}1.84 & \cellcolor{green!20}1.22 & \cellcolor{green!20}3.50 & \cellcolor{green!20}3.60 & \cellcolor{green!20}4.78 & \cellcolor{green!20}1.61 & \cellcolor{green!20}7.67 & \cellcolor{green!20}1.91 & \cellcolor{green!20}3.37 & \cellcolor{green!20}4.15 & \cellcolor{green!20}3.01 & \cellcolor{green!20}3.33 \\

\bottomrule
\end{tabular}
}
\end{table*}

\begin{SCtable*}
\small
\centering
\caption {\small Ablation study on the effect of margin regularization ($\mathcal{L}_{\text{Margin}}$) and moment-matching loss ($\mathcal{L}_{\text{mom.}}$) on novel classes.}
\label{table:base_results_ablation}
\resizebox{0.8\textwidth}{!}{ 
\setlength{\tabcolsep}{3pt} % Reduce column spacing
\begin{tabular}{l c ||c c c c c c c c c c c c}
\toprule
\thead{\textbf{Method}} & & \multicolumn{1}{c}{\thead{\rotatebox{90}{Calt}}} & \multicolumn{1}{c}{\thead{\rotatebox{90}{Food}}} & \multicolumn{1}{c}{\thead{\rotatebox{90}{DTD}}} & \multicolumn{1}{c}{\thead{\rotatebox{90}{UCF}}} & \multicolumn{1}{c}{\thead{\rotatebox{90}{Flow.}}} & \multicolumn{1}{c}{\thead{\rotatebox{90}{Pets}}} & \multicolumn{1}{c}{\thead{\rotatebox{90}{Air.}}} & \multicolumn{1}{c}{\thead{\rotatebox{90}{Cars}}} & \multicolumn{1}{c}{\thead{\rotatebox{90}{Sun}}} & \multicolumn{1}{c}{\thead{\rotatebox{90}{Euro}}} & \multicolumn{1}{c}{\thead{\rotatebox{90}{\textbf{Avg}}}}\\
\midrule
\rowcolor{orange!20} \multicolumn{14}{c}{\textbf{MaPLe}~\citep{maple}} \\ 
\hline
\multirow{2}{*}{$\mathcal{L}_\text{Margin}$}
 & Acc. & 94.80 & 91.87 & 60.47 & 77.30 & 71.70 & 97.83 & 33.93 & 74.80 & 78.33 & 76.40 & 75.74 \\
 & ECE & 2.00 & 1.07 & 13.92 & 3.82 & 15.76 & 1.73 & 11.53 & 2.45 & 3.25 & 6.01 & 5.09 \\
\hdashline[2pt/2pt]
\multirow{2}{*}{$\mathcal{L}_\text{Margin}+\mathcal{L}_\text{1}$}
 & Acc. & 94.23 & 92.13 & 59.50 & 79.10 & 73.33 & 97.77 & 26.43 & 74.57 & 78.47 & 70.93 & 74.65 \\
 & ECE & 2.32 & 0.55 & 12.13 & 3.90 & 11.71 & 1.72 & 7.39 & 2.58 & 1.58 & 7.80 & 5.17 \\
\hdashline[2pt/2pt]
\multirow{2}{*}{$\mathcal{L}_\text{Margin}+\mathcal{L}_\text{mom.}$}
 & Acc. & 94.87 & 91.77 & 61.27 & 79.87 & 73.87 & 97.57 & 36.03 & 75.27 & 78.13 & 67.60 & 75.63 \\
 & ECE & 1.42 & 0.86 & 7.37 & 3.32 & 10.07 & 2.29 & 8.33 & 2.60 & 1.11 & 7.45 & 4.48 \\
\bottomrule
\end{tabular}
}
\vspace{-1.0em}
\end{SCtable*}
We follow the standard few-shot protocol~\citep{coop}, each dataset is split into disjoint base and novel classes. Prompt tuning is performed only on base classes using a limited number of labeled samples per class. The calibration performance is reported for both base and novel classes.

%\subsection{Experimental Setup}

\noindent \textbf{Datasets.} We evaluate on 11 datasets spanning coarse-grained, fine-grained, and domain-specific tasks. Object classification is assessed on ImageNet~\citep{ImageNet} and Caltech101~\citep{Caltech-101}. Fine-grained recognition tasks include DTD~\citep{DTD} (textures), Flowers (FLW)~\citep{flowers}, Food101~\citep{food101}, SUN397~\citep{sun397}, and UCF101~\citep{ucf101}. Domain-specific benchmarks comprise Stanford Cars~\citep{stanfordcars}, FGVC-Aircraft~\citep{stanfordplanes}, Oxford Pets~\citep{oxfordpets}, and EuroSAT~\citep{eurosat}. We also provide results on out-of-distribution datasets for
the ImageNet-A~\citep{hendrycks2021natural}, ImageNet-V2~\citep{shankar2020evaluating}, ImageNet-R~\citep{hendrycks2021many}, and
ImageNet-S~\citep{wang2019learning} datasets in the suppl. material.

%We evaluate our approach on a diverse collection of datasets spanning multiple domains. For object recognition, we use ImageNet \citep{ImageNet} and Caltech101 \citep{Caltech-101}. Fine-grained classification is assessed on several benchmarks: DTD \citep{DTD} for textures, Flowers \citep{flowers} (FLW) for floral categories, Food101 \citep{food101} for culinary images, SUN397 \citep{sun397} for scene recognition, and UCF101 \citep{ucf101} for action recognition. Domain-specific benchmarks include Stanford Cars \citep{stanfordcars} and FGVC-Aircraft \citep{stanfordplanes} for vehicle types, Oxford Pets \citep{oxfordpets} for pet breeds, and EuroSAT \citep{eurosat} for satellite-based land use classification. We also provide results on out-of-distribution datasets for the ImageNet-A~\citep{hendrycks2021natural}, ImageNet-V2~\citep{shankar2020evaluating}, ImageNet-R~\citep{hendrycks2021many}, and ImageNet-S~\citep{wang2019learning} datasets in the supplementary material.

\noindent \textbf{Baseline Methods.} We benchmark our approach against a diverse and comprehensive set of prompt-tuning and calibration techniques. Classical baselines include MBLS~\citep{mbls} and Temperature Scaling~\citep{guo2017calibration}. We also compare with recent methods such as DAC~\citep{dac}, a post-hoc approach for adapting CLIP to novel classes, and a training-time regularization method introduced by~\citep{eccv}. For prompt tuning, we evaluate three representative approaches in the main paper: CoOp~\citep{coop}, KgCoOp~\citep{kgcoop}, and MaPLe~\citep{maple}.
%We benchmark our approach against a diverse set of prompt-tuning and calibration techniques. Classical baselines include MBLS~\citep{mbls} and Temperature Scaling~\citep{guo2017calibration}. We also compare with recent methods such as DAC~\citep{dac}, a post-hoc approach for adapting CLIP to novel classes, and a training-time method introduced by~\citep{eccv}. For prompt tuning, we use three representative approaches in the main paper: CoOp~\citep{coop}, KgCoOp~\citep{kgcoop}, and MaPLe~\citep{maple}. 
Results for additional methods including CoCoOp~\citep{cocoop}, ProDA~\citep{proda}, ProGrad~\citep{prograd}, and PromptSRC~\citep{promptsrc} are in the supp. material.

%We benchmark our approach against a diverse set of prompt-tuning and calibration techniques. Classical baselines include MBLS \citep{mbls} and Temperature Scaling \citep{guo2017calibration}. We also compare with recent methods such as DAC \citep{dac}, a post-hoc approach for adapting CLIP to novel classes, and a training-time method introduced by \citep{eccv}  Evaluations are conducted across multiple prompt-tuning strategies, including CoOp \citep{coop}, CoCoOp \citep{cocoop}, ProDA \citep{proda}, KGCoOp \citep{kgcoop}, ProGrad \citep{prograd}, MaPLe \citep{maple}, and PromptSRC \citep{promptsrc}. Note that our framework is prompt-agnostic and can be readily integrated with any of these prompting methods.

\noindent \textbf{Evaluation Metrics and Implementation details.} We evaluate classification performance using top-1 accuracy (ACC). To measure model calibration, we report the Expected Calibration Error (ECE), a widely used metric. We also report results with Adaptive Calibration Error~\citep{ACEM} (ACE) and Maximum Calibration Error~\citep{ECE} (MCE) in the supp. material.
% Given $N$ samples partitioned into $M$ bins $\{b_1, b_2, \dots, b_M\}$, ECE is defined as $\text{ECE} = \sum_{m=1}^{M} \frac{|b_m|}{N} \left| \text{acc}(b_m) - \text{conf}(b_m) \right|$, where $\text{acc}(b_m)$ and $\text{conf}(b_m)$ denote the average accuracy and confidence in bin $b_m$, respectively. 
%We provide results for other calibration metrics including Adaptive Calibration Error~\citep{ACEM} (ACE) and Maximum Calibration Error~\citep{ECE} (MCE) in the supp. material.
For all experiments, we use CLIP (ViT-B/16)~\citep{radford2021learning} as the pre-trained vision-language model. Prompt-tuning is conducted in a few-shot setting with 16 samples per class, using a learning rate of 0.005 and a batch size of 8. For each baseline method, we adopt its official implementation. All experiments are performed on an NVIDIA RTX A6000 GPU with 48GB memory. We provide detailed hyperparameters in the supp. material.
%For all experiments, we use CLIP (ViT-B/16)~\citep{radford2021learning} as the pre-trained vision-language model. Prompt-tuning is conducted in a few-shot setting with 16 samples per class, using a learning rate of 0.005 and a batch size of 8. For each baseline, we adopt the official implementation. All experiments are carried out on an NVIDIA RTX A6000 GPU with 48GB memory. For our $R_{\mathrm{margin}}$ loss, we selected $\alpha = 0.1$ and $\beta = 0.01$, and we set $\lambda = 5.0$ for $L_{\mathrm{mom}}$. All hyperparameters were chosen using grid search.

%-- natural distribution experiments

%-- one more metric...

%\subsection{Results}

\noindent \textbf{Calibration on Base Classes.} Table~\ref{table:main_result} reports the results on 11 fine-grained classification datasets using three prompt-tuning methods. Our approach consistently maintains or slightly improves the classification accuracy while significantly reducing the calibration error. When applied to MaPLe, it improves the average accuracy from 82.41\% to 82.75\% and reduces ECE from 3.19\% to 2.78\%. For CoOp, we observe a larger drop in ECE from 6. 35\% to 2. 93\%, surpassing both post hoc techniques such as temperature scaling (2.96\%) and prior regularization methods. Gains are especially pronounced on highly miscalibrated datasets; for example, CoOp on Aircraft shows a reduction from 25.70\% to 4.96\% in ECE. These improvements show that preserving margin distributions and embedding geometry effectively mitigates underconfidence in base-class predictions.

%Table~\ref{table:main_result} presents our comprehensive evaluation on 11 fine-grained classification datasets across three prompt-tuning methods. Our approach maintains or slightly improves classification accuracy while substantially enhancing calibration on base classes. When integrated with MaPLe, our method achieves an average accuracy of 82.75\% (vs. 82.41\% for vanilla MaPLe) while reducing ECE from 3.19\% to 2.78\%. For CoOp, we reduce average ECE from 6.35\% to 2.93\%, outperforming both post-hoc methods like Temperature Scaling (2.96\%) and other regularization approaches. The most notable improvements occur on datasets with high baseline miscalibration, for instance, with CoOp on Aircraft, our method reduces ECE from 25.70\% to 4.96\%. These results confirm that preserving margin statistics and embedding geometry effectively mitigates underconfidence on base classes.

\noindent \textbf{Calibration on Novel Classes.} Table~\ref{table:novel_classes} reports accuracy and ECE on novel classes under the open-vocabulary setting. While prompt-tuned models such as CoOp and MaPLe offer improved accuracy over zero-shot CLIP, they often exhibit severe overconfidence (e.g., CoOp ECE = 12.45, MaPLe ECE = 5.76). Post-hoc methods like Temperature Scaling and DAC yield marginal improvements but struggle to correct the underlying semantic drift. Our method consistently reduces calibration error while preserving or improving accuracy. For example, with MaPLe, our method reduces average ECE from 5.76 to 4.23 while maintaining comparable accuracy (75.14\%). Notably, our method consistently outperforms recent post-hoc techniques like DAC  and avoids the accuracy-calibration tradeoffs seen with methods like ZS-Norm. These results confirm that our approach effectively mitigates overconfidence on novel classes while preserving generalization capabilities.

%\input{tables/ablation-prompts}

%\input{tables/ablation-shots}

%\input{tables/ablation-8shot-base}
%\input{tables/ablation-8shot-novel}
%\subsection{Ablation Studies}

\noindent \textbf{Ablation Study on Loss Components.}
Table~\ref{table:base_results_ablation} analyzes the contribution of each component on novel class calibration with the MaPLe backbone. All variants are built on a cross-entropy baseline, where fine-tuning with only CE yields 5.76\% ECE. Adding margin regularization ($\mathcal{L}_{\text{Margin}}$) reduces ECE to 5.09\% while maintaining accuracy at 75.74\%. When combining margin regularization with direct $\ell_1$ alignment ($\mathcal{L}_1$), we observe improvements on some datasets but slightly lower overall accuracy (74.65\%). Our full method, combining margin regularization with moment matching ($\mathcal{L}_{\text{mom}}$), achieves the best calibration (4.48\% ECE) while preserving clean accuracy (75.63\%). This demonstrates that moment matching provides superior regularization compared to direct embedding alignment, effectively preserving the global structure of CLIP's embedding space while allowing task-specific adaptations.

%Adding only the margin regularization $\mathcal{L}{\text{margin}}$ improves accuracy but leaves high ECE due to miscalibration beyond decision boundaries. Replacing L1 alignment with our moment-matching loss $\mathcal{L}{\text{mom.}}$ further reduces ECE and improves generalization, particularly on distributionally distant classes like Aircraft and EuroSAT. The full combination consistently outperforms other variants, confirming that both margin consistency and global structure preservation are critical for calibrated open-vocabulary performance.

%\noindent \textbf{Only with R_margin}

%\noindent \textbf{R_margin+l1 loss}

%\begin{landscape}
\begin{figure}[t]
    \centering
    \begin{subfigure}{0.95\textwidth}
        \centering
        \includegraphics[width=\linewidth, trim=10 10 10 10, clip]{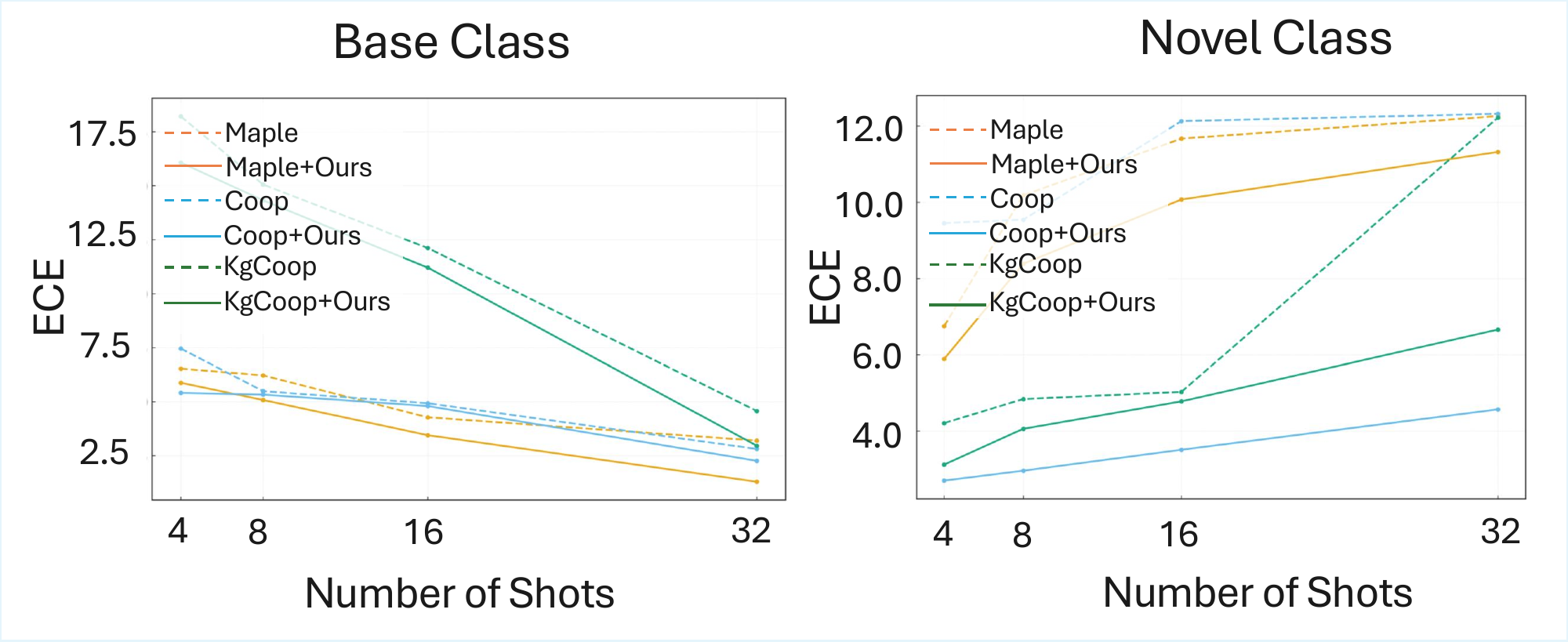}
        \caption{Calibration analyses on number of shots.}
        \label{fig:shots_analyse}
    \end{subfigure}\hfill
    \begin{subfigure}{0.95\textwidth}
        \centering
        \includegraphics[width=\linewidth, trim=10 10 10 10, clip]{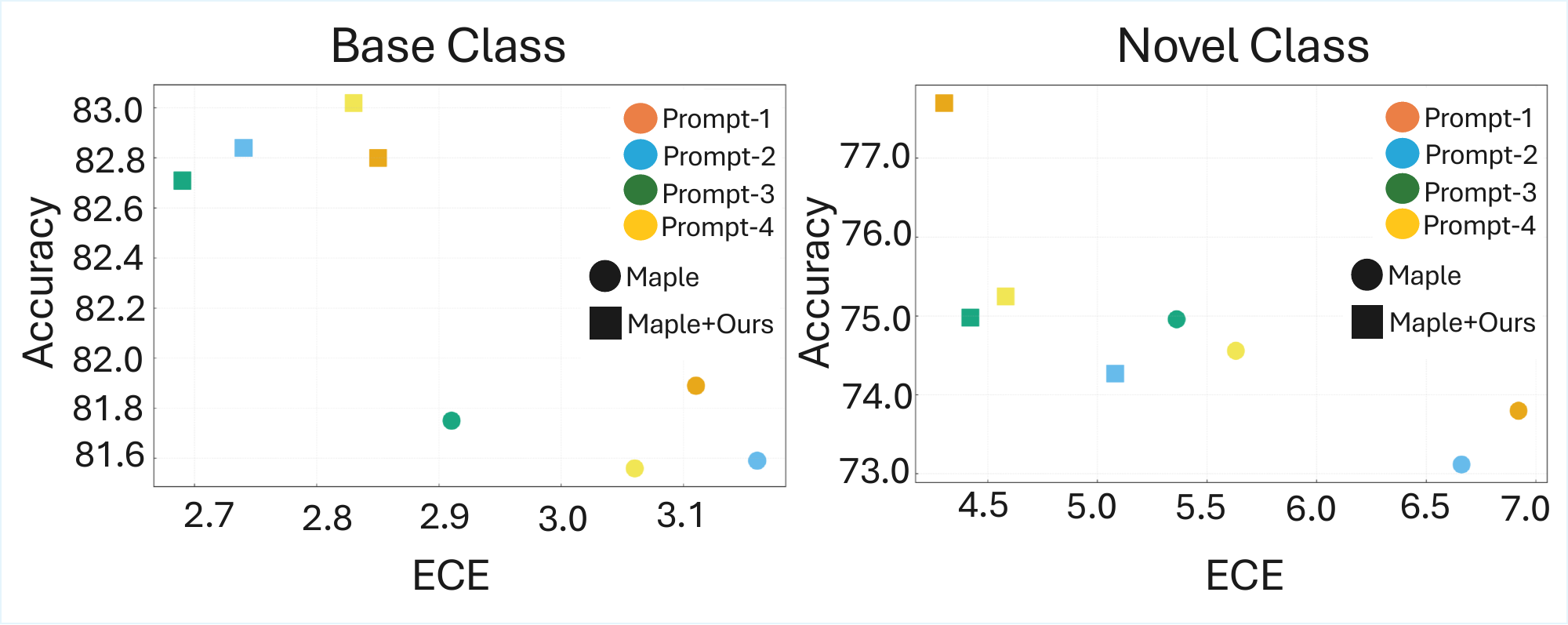}
        \caption{Different prompt initialization performance.}
        \label{fig:prompt_analyse}
    \end{subfigure}
    \caption{\small Performance with different numbers of shots and hard prompt styles.}
    \label{fig:shot_prompt}
\end{figure}
%\end{landscape}

\noindent \textbf{Calibration Across Varying Shots:} Figure~\ref{fig:shots_analyse} shows classification and calibration metrics under different shot counts (4, 8, 16, 32) for both base and novel classes. Ours maintains low ECE across different prompt tuning settings, demonstrating robust calibration even in extreme few-shot scenarios. %As expected, accuracy improves with more labeled examples. However, unlike prior work where calibration often degrades in low-data regimes~\citep{zhou2022learning}, our method maintains low ECE across settings, demonstrating robust calibration even in extreme few-shot scenarios. Notably, on base classes, ECE drops from 3.36\% (4-shot) to 1.32\% (32-shot), while novel class ECE remains competitive.

\noindent\textbf{Robustness to Prompt Initialization:} Figure~\ref{fig:prompt_analyse} evaluates the robustness of our method to different prompt initialization strategies, averaged over 10 datasets, comparing in between vanilla \textbf{Maple}~\cite{maple} and \textbf{Maple+Ours}. \textbf{Maple+Ours} demonstrates consistent performance across different initialization choices (see suppl.) for both base and novel classes. %with base class ECE ranging from 1.44\% to 2.16\% and novel class ECE from 5.02\% to 6.79\%. Similar stability is seen for Maximum Calibration Error (MCE) and Adaptive Calibration Error (ACE) metrics. 
 This robustness across initialization schemes suggests that our regularization framework addresses the fundamental geometric factors underlying miscalibration rather than optimizing for specific conditions. %The consistent performance is particularly valuable in practical deployment scenarios where prompt initialization may not be known in advance.

 \begin{figure}[t]
    \centering
    \includegraphics[width=\linewidth, trim=10 10 10 10, clip]{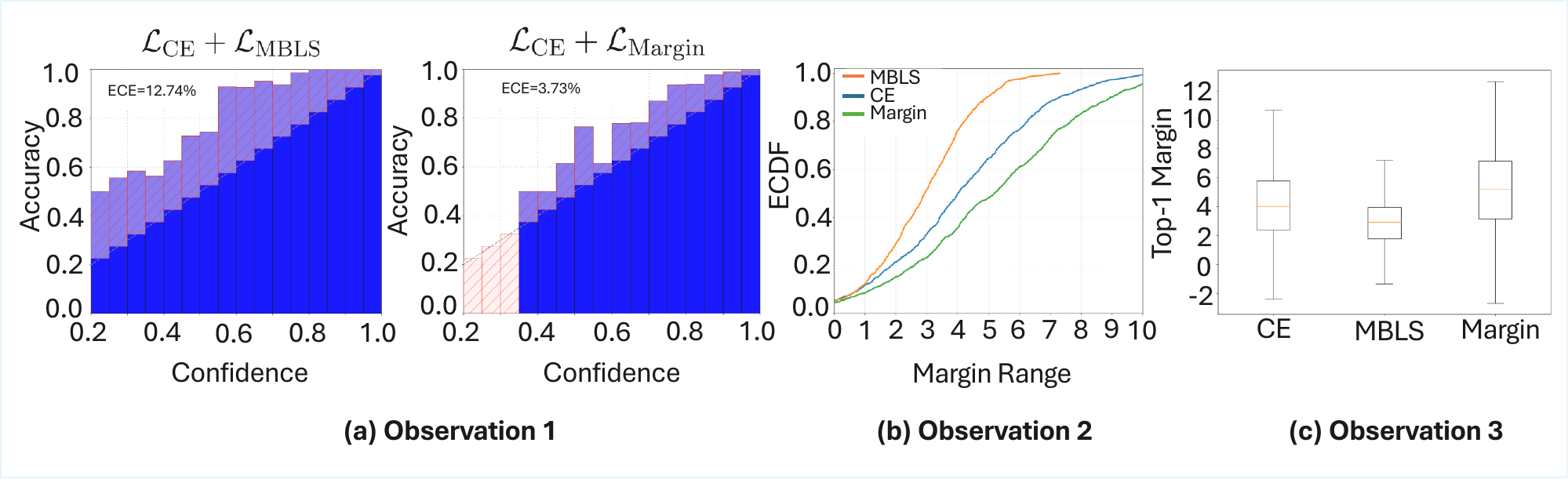}
    \caption{\small \textbf{Margin based Label Smoothing(MBLS) vs Mean-Variance Margin(Margin) Regularization.}
    %Mbls constraint-based objective only allows backpropagation when logits distances are above the margin. In this case, it will only address the overconfidence issue where the margin is high. It doesn’t address the underconfidence if the logit distance is below the margin, where it has very low Inter-class Margin variability. 
    %As we can observe in the base class reliability diagram(Observation 1), Mbls with Cross-Entropy loss have an Underconfidence issue. On the other hand, as you can see in the right figure of the reliability diagram shows the \textbf{CE + Margin} term reduces the underconfidence issue. For Observations 3 and 4, we train \textbf{CE}, \textbf{CE + Mbls} and \textbf{CE + Margin} with MaPLe on Base Classes For each test sample of each loss-based method, we compute the top-1 margin \(m=z_y-\max_{j\neq y}z_j\) and plot the number of samples using the Empirical Cumulative Distribution Function (ECDF) plot. The ECDF plot shows that \textbf{Margin} loss lies lowest in the small margin region, indicating fewer low-margin/under-confident samples. \textbf{MBLS} has very steep curve, showing it trims very large margins but leaves many small ones.
  %\textbf{CE} sits between the two. Concurrently, the Box plots illustrate that the \textbf{Margin} term shows the highest median (\(\approx5\)) and a right-shifted Interquartile range (IQR), confirming it lifts small/medium margins. \textbf{MBLS} has the tightest IQR and the shortest upper whisker, consistent with capping extreme margins without raising the low-margin samples. \textbf{CE} retains a broad spread.
In base class reliability diagram (Observation 1), \textbf{MBLS} with Cross-Entropy (CE) shows underconfidence, 
while adding a \textbf{Margin} term alleviates this. For Observations 3 and 4, we train \textbf{CE}, \textbf{CE + MBLS}, 
and \textbf{CE + Margin} with MaPLe, compute the top-1 margin \(m=z_y-\max_{j\neq y}z_j\), and plot the Empirical Cumulative Distribution Function (ECDF). The 
 ECDF shows \textbf{Margin} yields fewer low-margin samples(Underconfident samples), \textbf{MBLS} trims large but leaves small ones, 
and \textbf{CE} lies in between. Box plots confirm this: \textbf{Margin} has the highest median ($\approx 5$) with a 
right-shifted IQR, \textbf{MBLS} caps extremes with a tight IQR, while \textbf{CE} retains a broad spread.}

    \label{fig:combine_moti}\vspace{-1.0em}
\end{figure}

\noindent\textbf{Margin vs MBLS:} Compared to MBLS, Margin regularization reduces underconfidence by lifting low-margin samples (Figure~\ref{fig:combine_moti}, Observations 1–3), whereas MBLS primarily caps extreme margins. 
%and ECDF plots. Box plots further show that Margin achieves higher median margins and a right-shifted distribution, while MBLS mainly caps extreme margins. 

\vspace{-1em}
\section{Conclusion}
\vspace{-0.5em}
We proposed a method to calibrate prompt-tuned VLMs by addressing a key challenge: preserving predictive reliability without distorting CLIP’s semantic structure. Our approach introduces two complementary regularizers, a mean-variance logit margin loss and a moment-matching constraint on text embeddings, to jointly enforce geometric fidelity and calibration. It requires no architectural changes and integrates seamlessly with existing prompt tuning setups, making it broadly applicable across domains. Evaluated on 11 datasets and multiple prompt-tuning strategies, our method consistently improves calibration, particularly on novel classes, without compromising classification accuracy. By decoupling predictive uncertainty from semantic drift, it enables more trustworthy and robust deployment of VLMs in real-world, open-vocabulary scenarios. We hope this work encourages further research into calibration-aware adaptations of foundation models. %Despite these advances, our work has several limitations. First, our moment-matching approach relies on batch statistics, which may become unstable with very small batch sizes or highly imbalanced class distributions. Second, our method introduces additional hyperparameters that require tuning, potentially increasing computational overhead during development.

 \clearpage
{
    \small
    \bibliographystyle{ieeenat_fullname}
    \bibliography{main}

@String(IJCV = {Int. J. Comput. Vis.})

@String(CVPR= {IEEE Conf. Comput. Vis. Pattern Recog.})

@String(ECCV= {Eur. Conf. Comput. Vis.})

@String(AAAI = {AAAI})

@String(IJCV  = {IJCV})

@String(CVPR  = {CVPR})

@String(ECCV  = {ECCV})

@inproceedings{khattak2023maple,
  title={Maple: Multi-modal prompt learning},
  author={Khattak, Muhammad Uzair and Rasheed, Hanoona and Maaz, Muhammad and Khan, Salman and Khan, Fahad Shahbaz},
  booktitle={Proceedings of the IEEE/CVF conference on computer vision and pattern recognition},
  pages={19113--19122},
  year={2023}
}

@article{zhou2022learning,
  title={Learning to prompt for vision-language models},
  author={Zhou, Kaiyang and Yang, Jingkang and Loy, Chen Change and Liu, Ziwei},
  journal={International Journal of Computer Vision},
  volume={130},
  number={9},
  pages={2337--2348},
  year={2022},
  publisher={Springer}
}

@inproceedings{zhou2022conditional,
  title={Conditional prompt learning for vision-language models},
  author={Zhou, Kaiyang and Yang, Jingkang and Loy, Chen Change and Liu, Ziwei},
  booktitle={Proceedings of the IEEE/CVF conference on computer vision and pattern recognition},
  pages={16816--16825},
  year={2022}
}

@article{shao2024uncertainty,
  title={Uncertainty-aware prediction and application in planning for autonomous driving: Definitions, methods, and comparison},
  author={Shao, Wenbo and Xu, Jiahui and Cao, Zhong and Wang, Hong and Li, Jun},
  journal={arXiv preprint arXiv:2403.02297},
  year={2024}
}

@article{lambert2024trustworthy,
  title={Trustworthy clinical AI solutions: A unified review of uncertainty quantification in Deep Learning models for medical image analysis.},
  author={Lambert, Benjamin and Forbes, Florence and Doyle, Senan and Dehaene, Harmonie and Dojat, Michel},
  journal={Artif. Intell. Medicine},
  volume={150},
  pages={102830},
  year={2024},
  publisher={Elsevier}
}

@article{elhenawy2025vision,
  title={Vision-language models for autonomous driving: Clip-based dynamic scene understanding},
  author={Elhenawy, Mohammed and Ashqar, Huthaifa I and Rakotonirainy, Andry and Alhadidi, Taqwa I and Jaber, Ahmed and Tami, Mohammad Abu},
  journal={Electronics},
  volume={14},
  number={7},
  pages={1282},
  year={2025},
  publisher={MDPI}
}

@article{zhao2025clip,
  title={CLIP in medical imaging: A survey},
  author={Zhao, Zihao and Liu, Yuxiao and Wu, Han and Wang, Mei and Li, Yonghao and Wang, Sheng and Teng, Lin and Liu, Disheng and Cui, Zhiming and Wang, Qian and others},
  journal={Medical Image Analysis},
  pages={103551},
  year={2025},
  publisher={Elsevier}
}

@inproceedings{guo2017calibration,
  title={On calibration of modern neural networks},
  author={Guo, Chuan and Pleiss, Geoff and Sun, Yu and Weinberger, Kilian Q},
  booktitle={International conference on machine learning},
  pages={1321--1330},
  year={2017},
  organization={PMLR}
}

@article{wang2023calibration,
  title={Calibration in deep learning: A survey of the state-of-the-art},
  author={Wang, Cheng},
  journal={arXiv preprint arXiv:2308.01222},
  year={2023}
}

@article{gawlikowski2023survey,
  title={A survey of uncertainty in deep neural networks},
  author={Gawlikowski, Jakob and Tassi, Cedrique Rovile Njieutcheu and Ali, Mohsin and Lee, Jongseok and Humt, Matthias and Feng, Jianxiang and Kruspe, Anna and Triebel, Rudolph and Jung, Peter and Roscher, Ribana and others},
  journal={Artificial Intelligence Review},
  volume={56},
  number={Suppl 1},
  pages={1513--1589},
  year={2023},
  publisher={Springer}
}

@inproceedings{radford2021learning,
  title={Learning transferable visual models from natural language supervision},
  author={Radford, Alec and Kim, Jong Wook and Hallacy, Chris and Ramesh, Aditya and Goh, Gabriel and Agarwal, Sandhini and Sastry, Girish and Askell, Amanda and Mishkin, Pamela and Clark, Jack and others},
  booktitle={International conference on machine learning},
  pages={8748--8763},
  year={2021},
  organization={PmLR}
}

@article{liu2023pre,
  title={Pre-train, prompt, and predict: A systematic survey of prompting methods in natural language processing},
  author={Liu, Pengfei and Yuan, Weizhe and Fu, Jinlan and Jiang, Zhengbao and Hayashi, Hiroaki and Neubig, Graham},
  journal={ACM computing surveys},
  volume={55},
  number={9},
  pages={1--35},
  year={2023},
  publisher={ACM New York, NY}
}

@inproceedings{shankar2020evaluating,
  title={Evaluating machine accuracy on imagenet},
  author={Shankar, Vaishaal and Roelofs, Rebecca and Mania, Horia and Fang, Alex and Recht, Benjamin and Schmidt, Ludwig},
  booktitle={International Conference on Machine Learning},
  pages={8634--8644},
  year={2020},
  organization={PMLR}
}

@article{wang2019learning,
  title={Learning robust global representations by penalizing local predictive power},
  author={Wang, Haohan and Ge, Songwei and Lipton, Zachary and Xing, Eric P},
  journal={Advances in neural information processing systems},
  volume={32},
  year={2019}
}

@inproceedings{hendrycks2021many,
  title={The many faces of robustness: A critical analysis of out-of-distribution generalization},
  author={Hendrycks, Dan and Basart, Steven and Mu, Norman and Kadavath, Saurav and Wang, Frank and Dorundo, Evan and Desai, Rahul and Zhu, Tyler and Parajuli, Samyak and Guo, Mike and others},
  booktitle={Proceedings of the IEEE/CVF international conference on computer vision},
  pages={8340--8349},
  year={2021}
}

@inproceedings{hendrycks2021natural,
  title={Natural adversarial examples},
  author={Hendrycks, Dan and Zhao, Kevin and Basart, Steven and Steinhardt, Jacob and Song, Dawn},
  booktitle={Proceedings of the IEEE/CVF conference on computer vision and pattern recognition},
  pages={15262--15271},
  year={2021}
}

@article{yoon2024c,
  title={C-tpt: Calibrated test-time prompt tuning for vision-language models via text feature dispersion},
  author={Yoon, Hee Suk and Yoon, Eunseop and Tee, Joshua Tian Jin and Hasegawa-Johnson, Mark and Li, Yingzhen and Yoo, Chang D},
  journal={arXiv preprint arXiv:2403.14119},
  year={2024}
}

@article{sharifdeen2025tpt,
  title={O-TPT: Orthogonality Constraints for Calibrating Test-time Prompt Tuning in Vision-Language Models},
  author={Sharifdeen, Ashshak and Munir, Muhammad Akhtar and Baliah, Sanoojan and Khan, Salman and Khan, Muhammad Haris},
  journal={arXiv preprint arXiv:2503.12096},
  year={2025}
}

@article{wang2024open,
  title={Open-vocabulary calibration for fine-tuned CLIP},
  author={Wang, Shuoyuan and Wang, Jindong and Wang, Guoqing and Zhang, Bob and Zhou, Kaiyang and Wei, Hongxin},
  journal={arXiv preprint arXiv:2402.04655},
  year={2024}
}

@inproceedings{niculescu2005predicting,
  title={Predicting good probabilities with supervised learning},
  author={Niculescu-Mizil, Alexandru and Caruana, Rich},
  booktitle={Proceedings of the 22nd international conference on Machine learning},
  pages={625--632},
  year={2005}
}

@article{ovadia2019can,
  title={Can you trust your model's uncertainty? evaluating predictive uncertainty under dataset shift},
  author={Ovadia, Yaniv and Fertig, Emily and Ren, Jie and Nado, Zachary and Sculley, David and Nowozin, Sebastian and Dillon, Joshua and Lakshminarayanan, Balaji and Snoek, Jasper},
  journal={Advances in neural information processing systems},
  volume={32},
  year={2019}
}

@inproceedings{kumar2018trainable,
  title={Trainable calibration measures for neural networks from kernel mean embeddings},
  author={Kumar, Aviral and Sarawagi, Sunita and Jain, Ujjwal},
  booktitle={International Conference on Machine Learning},
  pages={2805--2814},
  year={2018},
  organization={PMLR}
}

@inproceedings{ECE,
  title={Obtaining well calibrated probabilities using bayesian binning},
  author={Naeini, Mahdi Pakdaman and Cooper, Gregory and Hauskrecht, Milos},
  booktitle={Proceedings of the AAAI conference on artificial intelligence},
  volume={29},
  number={1},
  year={2015}
}

@article{PLIP,
  title={A visual--language foundation model for pathology image analysis using medical twitter},
  author={Huang, Zhi and Bianchi, Federico and Yuksekgonul, Mert and Montine, Thomas J and Zou, James},
  journal={Nature medicine},
  volume={29},
  number={9},
  pages={2307--2316},
  year={2023},
  publisher={Nature Publishing Group US New York}
}

@article{QUILTNET,
  title={Quilt-1m: One million image-text pairs for histopathology},
  author={Ikezogwo, Wisdom and Seyfioglu, Saygin and Ghezloo, Fatemeh and Geva, Dylan and Sheikh Mohammed, Fatwir and Anand, Pavan Kumar and Krishna, Ranjay and Shapiro, Linda},
  journal={Advances in neural information processing systems},
  volume={36},
  pages={37995--38017},
  year={2023}
}

@article{KATHER,
  title={Predicting survival from colorectal cancer histology slides using deep learning: A retrospective multicenter study},
  author={Kather, Jakob Nikolas and Krisam, Johannes and Charoentong, Pornpimol and Luedde, Tom and Herpel, Esther and Weis, Cleo-Aron and Gaiser, Timo and Marx, Alexander and Valous, Nektarios A and Ferber, Dyke and others},
  journal={PLoS medicine},
  volume={16},
  number={1},
  pages={e1002730},
  year={2019},
  publisher={Public Library of Science San Francisco, CA USA}
}

@inproceedings{PANNUKE,
  title={Pannuke: an open pan-cancer histology dataset for nuclei instance segmentation and classification},
  author={Gamper, Jevgenij and Alemi Koohbanani, Navid and Benet, Ksenija and Khuram, Ali and Rajpoot, Nasir},
  booktitle={Digital Pathology: 15th European Congress, ECDP 2019, Warwick, UK, April 10--13, 2019, Proceedings 15},
  pages={11--19},
  year={2019},
  organization={Springer}
}

@article{da2022digestpath,
  title={DigestPath: A benchmark dataset with challenge review for the pathological detection and segmentation of digestive-system},
  author={Da, Qian and Huang, Xiaodi and Li, Zhongyu and Zuo, Yanfei and Zhang, Chenbin and Liu, Jingxin and Chen, Wen and Li, Jiahui and Xu, Dou and Hu, Zhiqiang and others},
  journal={Medical Image Analysis},
  volume={80},
  pages={102485},
  year={2022},
  publisher={Elsevier}
}

@INPROCEEDINGS{ImageNet,
  author={Deng, Jia and Dong, Wei and Socher, Richard and Li, Li-Jia and Kai Li and Li Fei-Fei},
  booktitle={2009 IEEE Conference on Computer Vision and Pattern Recognition}, 
  title={ImageNet: A large-scale hierarchical image database}, 
  year={2009},
  volume={},
  number={},
  pages={248-255},
  doi={10.1109/CVPR.2009.5206848}}

@article{Caltech-101,
  title={Learning Generative Visual Models from Few Training Examples: An Incremental Bayesian Approach Tested on 101 Object Categories},
  author={Li Fei-Fei and Rob Fergus and Pietro Perona},
  journal={Computer Vision and Pattern Recognition Workshop},
  year={2004},
}

@inproceedings{DTD,
  title={Describing textures in the wild},
  author={Cimpoi, Mircea and Maji, Subhransu and Kokkinos, Iasonas and Mohamed, Sammy and Vedaldi, Andrea},
  booktitle={Proceedings of the IEEE conference on computer vision and pattern recognition},
  pages={3606--3613},
  year={2014}
}

@inproceedings{flowers,
  title={Automated flower classification over a large number of classes},
  author={Nilsback, Maria-Elena and Zisserman, Andrew},
  booktitle={2008 Sixth Indian conference on computer vision, graphics \& image processing},
  pages={722--729},
  year={2008},
  organization={IEEE}
}

@inproceedings{food101,
  title = {Food-101 -- Mining Discriminative Components with Random Forests},
  author = {Bossard, Lukas and Guillaumin, Matthieu and Van Gool, Luc},
  booktitle = {European Conference on Computer Vision},
  year = {2014}
}

@article{sun397,
  title={Sun database: Exploring a large collection of scene categories},
  author={Xiao, Jianxiong and Ehinger, Krista A and Hays, James and Torralba, Antonio and Oliva, Aude},
  journal={International Journal of Computer Vision},
  volume={119},
  pages={3--22},
  year={2016},
  publisher={Springer}
}

@inproceedings{stanfordcars,
  title={3D Object Representations for Fine-Grained Categorization},
  author={Krause, Jonathan and Stark, Michael and Deng, Jia and Fei-Fei, Li},
  booktitle={Proceedings of the IEEE International Conference on Computer Vision Workshops (ICCVW)},
  pages={554--561},
  year={2013}
}

@article{stanfordplanes,
  title={Fine-grained visual classification of aircraft},
  author={Maji, Subhransu and Rahtu, Esa and Kannala, Juho and Blaschko, Matthew and Vedaldi, Andrea},
  journal={arXiv preprint arXiv:1306.5151},
  year={2013}
}

@article{ucf101,
  title={UCF101: A dataset of 101 human actions classes from videos in the wild},
  author={Soomro, Khurram and Zamir, Amir Roshan and Shah, Mubarak},
  journal={arXiv preprint arXiv:1212.0402},
  year={2012}
}

@inproceedings{oxfordpets,
  title={Cats and dogs},
  author={Parkhi, Omkar M and Vedaldi, Andrea and Zisserman, Andrew and Jawahar, CV},
  booktitle={2012 IEEE conference on computer vision and pattern recognition},
  pages={3498--3505},
  year={2012},
  organization={IEEE}
}

@article{eurosat,
  title={Eurosat: A novel dataset and deep learning benchmark for land use and land cover classification},
  author={Helber, Patrick and Bischke, Benjamin and Dengel, Andreas and Borth, Damian},
  journal={IEEE Journal of Selected Topics in Applied Earth Observations and Remote Sensing},
  volume={12},
  number={7},
  pages={2217--2226},
  year={2019},
  publisher={IEEE}
}

@misc{mbls,
      title={The Devil is in the Margin: Margin-based Label Smoothing for Network Calibration}, 
      author={Bingyuan Liu and Ismail Ben Ayed and Adrian Galdran and Jose Dolz},
      year={2023},
      eprint={2111.15430},
      archivePrefix={arXiv},
      primaryClass={cs.CV},
      url={https://arxiv.org/abs/2111.15430}, 
}

@inproceedings{dac,
  title={ Open-Vocabulary Calibration for Fine-tuned CLIP},
  author={Wang, Shuoyuan and Wang, Jindong and Wang, Guoqing and Zhang, Bob and Zhou, Kaiyang and Wei, Hongxin},
  booktitle = {International Conference on Machine Learning (ICML)},
  year = {2024}
}

@inproceedings{cocoop,
    title={Conditional Prompt Learning for Vision-Language Models},
    author={Zhou, Kaiyang and Yang, Jingkang and Loy, Chen Change and Liu, Ziwei},
    booktitle={IEEE/CVF Conference on Computer Vision and Pattern Recognition (CVPR)},
    year={2022}
}

@article{coop,
    title={Learning to Prompt for Vision-Language Models},
    author={Zhou, Kaiyang and Yang, Jingkang and Loy, Chen Change and Liu, Ziwei},
    journal={International Journal of Computer Vision (IJCV)},
    year={2022}
}

@inproceedings{proda,
  title={Visual-language prompt tuning with knowledge-guided context optimization},
  author={Yao, Hantao and Zhang, Rui and Xu, Changsheng},
  booktitle={Proceedings of the IEEE/CVF conference on computer vision and pattern recognition},
  pages={6757--6767},
  year={2023}
}

@inproceedings{kgcoop,
  title={Visual-language prompt tuning with knowledge-guided context optimization},
  author={Yao, Hantao and Zhang, Rui and Xu, Changsheng},
  booktitle={Proceedings of the IEEE/CVF conference on computer vision and pattern recognition},
  pages={6757--6767},
  year={2023}
}

@inproceedings{prograd,
  title={Prompt-aligned gradient for prompt tuning},
  author={Zhu, Beier and Niu, Yulei and Han, Yucheng and Wu, Yue and Zhang, Hanwang},
  booktitle={Proceedings of the IEEE/CVF international conference on computer vision},
  pages={15659--15669},
  year={2023}
}

@inproceedings{maple,
  title={Maple: Multi-modal prompt learning},
  author={Khattak, Muhammad Uzair and Rasheed, Hanoona and Maaz, Muhammad and Khan, Salman and Khan, Fahad Shahbaz},
  booktitle={Proceedings of the IEEE/CVF conference on computer vision and pattern recognition},
  pages={19113--19122},
  year={2023}
}

@inproceedings{promptsrc,
  title={Self-regulating prompts: Foundational model adaptation without forgetting},
  author={Khattak, Muhammad Uzair and Wasim, Syed Talal and Naseer, Muzammal and Khan, Salman and Yang, Ming-Hsuan and Khan, Fahad Shahbaz},
  booktitle={Proceedings of the IEEE/CVF international conference on computer vision},
  pages={15190--15200},
  year={2023}
}

@inproceedings{eccv,
 title={Robust calibration of large vision-language adapters},
  author={Murugesan, Balamurali and Silva-Rodr{\'\i}guez, Julio and Ayed, Ismail Ben and Dolz, Jose},
  booktitle={European Conference on Computer Vision},
  pages={147--165},
  year={2024},
  organization={Springer}
}

@inproceedings{ACEM,
  title={Measuring Calibration in Deep Learning.},
  author={Nixon, Jeremy and Dusenberry, Michael W and Zhang, Linchuan and Jerfel, Ghassen and Tran, Dustin},
  booktitle={CVPR workshops},
  volume={2},
  number={7},
  year={2019}
}
}

 %\clearpage
%\setcounter{page}{1}

\maketitlesupplementary

\section*{Contents of Supplementary Material}

In this supplementary material, we provide the following:

\begin{enumerate}
    \item \textbf{Calibration analysis for base and novel classes across prompt learning methods} (Sec.~\ref{sec:calib_analysis})
    \item \textbf{Robustness evaluation under natural distribution shifts} (Sec.~\ref{sec:dis_shift})
    \item \textbf{Additional results for ACE and MCE performance metrics} (Sec.~\ref{sec:ace_mce})
    \item \textbf{Hyperparameters details} (Sec.~\ref{sec:hyperparameters})
    \item \textbf{Prompt templates and variations} (Sec.~\ref{sec:prompt_var})
    \item \textbf{Variance analysis }(Sec.~\ref{sec:variance})
    \item \textbf{Results on Different Backbones}(Sec.~\ref{sec:backbones})
    \item \textbf{Decision Boundary Visualization}(Sec.~\ref{sec:decision_b})
     %\item \textbf{Reproducibulity statement}(Sec.~\ref{sec:repr})
    %\item \textbf{The Use of Large Language Models (LLMs)}(Sec.~\ref{sec:llmusage})
    %\item \textbf{Limitations}(Sec.~\ref{sec:limitations})
\end{enumerate}

\section{Calibration Analysis for Base and Novel Classes Across Prompt Learning Methods} \label{sec:calib_analysis}
\begin{table}[t!]
%\small
\centering
\caption {\small  \textbf{ Accuracy and calibration performance on base classes across 10 fine-grained classification benchmarks}.
We report top-1 accuracy (Acc) and Expected Calibration Error (ECE) for multiple prompt-tuning strategies and diverse calibration baselines. Higher Acc. indicates better classification performance, while lower ECE reflects better calibration.}
\label{table:supp_base_classes}
\resizebox{\textwidth}{!}{ 
\setlength{\tabcolsep}{4pt} % Reduce column spacing
\begin{tabular}{l c ||c c c c c c c c c c c}
\toprule
\thead{\textbf{Method}} & &  \multicolumn{1}{c}{\thead{\rotatebox{90}{Calt}}} & \multicolumn{1}{c}{\thead{\rotatebox{90}{Pets}}} & \multicolumn{1}{c}{\thead{\rotatebox{90}{Cars}}} & \multicolumn{1}{c}{\thead{\rotatebox{90}{Flow}}} & \multicolumn{1}{c}{\thead{\rotatebox{90}{Food}}} & \multicolumn{1}{c}{\thead{\rotatebox{90}{Air}}} & \multicolumn{1}{c}{\thead{\rotatebox{90}{SUN}}} & \multicolumn{1}{c}{\thead{\rotatebox{90}{DTD}}} & \multicolumn{1}{c}{\thead{\rotatebox{90}{Euro}}} & \multicolumn{1}{c}{\thead{\rotatebox{90}{UCF}}} & \multicolumn{1}{c}{\thead{\rotatebox{90}{\textbf{Avg}}}}\\
\midrule
\multirow{2}{*}{$\text{Zero Shot}$}
 & Acc. & 97.20 & 91.30 & 63.60 & 71.80 & 90.10 & 27.70 & 69.40 & 53.00 & 57.00 & 71.00 & 69.50 \\
 & ECE & 6.49 & 2.25 & 3.74 & 3.11 & 1.57 & 3.03 & 1.59 & 4.53 & 8.35 & 3.24 & 3.58 \\
\hline
\rowcolor{orange!20} \multicolumn{13}{c}{\textbf{CoCoOp}\cite{cocoop} } \\ 
\hline
\multirow{2}{*}{$\text{CoCoOp}$ \cite{cocoop}} 
 & Acc. & 97.77 & 95.02 & 70.72 & 94.62 & 90.43 & 35.33 & 79.19 & 75.54 & 85.40 & 81.71 & 80.57 \\
 & ECE & 1.43 & 3.21 & 6.89 & 7.85 & 0.86 & 5.42 & 3.78 & 3.88 & 8.09 & 3.78 & 4.52 \\
\hdashline[2pt/2pt]
\multirow{2}{*}{$\text{$\text{ZS-Norm}$}$ \cite{eccv}}
 & Acc. & 97.83 & 95.22 & 70.65 & 95.00 & 90.63 & 36.03 & 79.70 & 76.35 & 82.53 & 81.90 & 80.58 \\
 & ECE & 2.93 & 3.20 & 8.67 & 7.56 & 1.49 & 8.50 & 7.50 & 10.82 & 16.09 & 4.08 & 7.08 \\
 \hdashline[2pt/2pt]
\multirow{2}{*}{$\text{$\text{Penalty}$}$ \cite{eccv}}
 & Acc. & 97.83 & 94.98 & 69.95 & 92.43 & 90.72 & 34.35 & 79.34 & 71.49 & 69.57 & 80.49 & 78.12 \\
 & ECE & 5.00 & 6.06 & 10.36 & 14.90 & 3.95 & 6.83 & 6.69 & 17.22 & 20.94 & 7.17 & 9.91 \\
  \hdashline[2pt/2pt]
 \multirow{2}{*}{$\textbf{Ours}$}
 & Acc. & \cellcolor{green!20}97.93 & \cellcolor{green!20}94.69 & \cellcolor{green!20}69.42 & \cellcolor{green!20}94.11 & \cellcolor{green!20}90.51 & \cellcolor{green!20}34.25 & \cellcolor{green!20}78.92 & \cellcolor{green!20}74.77 & \cellcolor{green!20}84.70 & \cellcolor{green!20}82.34 & \cellcolor{green!20}80.16 \\
 & ECE & \cellcolor{green!20}1.13 & \cellcolor{green!20}2.31 & \cellcolor{green!20}7.01 & \cellcolor{green!20}7.98 & \cellcolor{green!20}0.49 & \cellcolor{green!20}5.81 & \cellcolor{green!20}2.63 & \cellcolor{green!20}3.70 & \cellcolor{green!20}6.47 & \cellcolor{green!20}3.23 & \cellcolor{green!20}4.08 \\
\hline
\rowcolor{orange!20} \multicolumn{13}{c}{\textbf{ProDA} \cite{proda}} \\ 
\hline
\multirow{2}{*}{$\text{ProDA}$ \cite{proda}}
 & Acc. & 97.61 & 94.75 & 69.76 & 89.96 & 89.33 & 33.01 & 76.17 & 70.02 & 81.83 & 79.99 & 78.24 \\
 & ECE & 1.06 & 1.67 & 3.86 & 6.07 & 0.86 & 3.52 & 6.66 & 10.25 & 3.73 & 2.56 & 4.02 \\
\hdashline[2pt/2pt]
\multirow{2}{*}{$\text{$\text{ZS-Norm}$}$ \cite{eccv}}
 & Acc. & 97.55 & 94.37 & 69.77 & 89.62 & 89.50 & 33.03 & 76.46 & 71.33 & 82.00 & 79.33 & 78.30 \\
 & ECE & 1.93 & 2.22 & 4.74 & 7.41 & 1.31 & 3.00 & 6.62 & 3.73 & 11.93 & 2.72 & 4.56 \\
 \hdashline[2pt/2pt]
\multirow{2}{*}{$\text{$\text{Penalty}$}$ \cite{eccv}}
 & Acc. & 97.35 & 94.61 & 69.32 & 89.14 & 90.36 & 32.17 & 76.94 & 59.80 & 63.86 & 78.07 & 75.16 \\
 & ECE & 4.00 & 8.11 & 7.86 & 12.89 & 3.79 & 4.91 & 2.09 & 11.28 & 18.05 & 7.83 & 8.08 \\
  \hdashline[2pt/2pt]
 \multirow{2}{*}{$\textbf{Ours}$}
 & Acc. & \cellcolor{green!20}97.20 & \cellcolor{green!20}94.31 & \cellcolor{green!20}69.50 & \cellcolor{green!20}87.88 & \cellcolor{green!20}90.02 & \cellcolor{green!20}32.99 & \cellcolor{green!20}76.96 & \cellcolor{green!20}72.91 & \cellcolor{green!20}82.70 & \cellcolor{green!20}80.63 & \cellcolor{green!20}78.51 \\
 & ECE & \cellcolor{green!20}1.75 & \cellcolor{green!20}2.29 & \cellcolor{green!20}6.91 & \cellcolor{green!20}8.21 & \cellcolor{green!20}1.18 & \cellcolor{green!20}3.38 & \cellcolor{green!20}1.18 & \cellcolor{green!20}2.27 & \cellcolor{green!20}5.37 & \cellcolor{green!20}2.50 & \cellcolor{green!20}3.50 \\

\hline
\rowcolor{orange!20} \multicolumn{13}{c}{\textbf{ProGrad} \cite{prograd}} \\ 
\hline
\multirow{2}{*}{$\text{ProGrad}$ \cite{prograd}}
 & Acc. & 97.72 & 94.67 & 69.29 & 81.26 & 90.33 & 31.35 & 76.88 & 67.13 & 79.27 & 78.20 & 76.61\\
 & ECE & 3.53 & 3.83 & 6.84 & 6.82 & 1.65 & 2.60 & 3.70 & 6.38 & 12.24 & 3.92 & 5.15\\
\hdashline[2pt/2pt]
\multirow{2}{*}{$\text{$\text{ZS-Norm}$}$ \cite{eccv}}
 & Acc. & 97.87 & 94.36 & 69.27 & 82.91 & 90.45 & 32.35 & 77.92 & 70.37 & 75.57 & 78.70 & 76.98 \\
 & ECE & 5.51 & 4.85 & 10.02 & 10.95 & 2.67 & 8.50 & 10.33 & 23.04 & 17.46 & 6.47 & 9.98 \\
 \hdashline[2pt/2pt]
  \hdashline[2pt/2pt]
\multirow{2}{*}{$\text{$\text{Penalty}$}$ \cite{eccv}}
 & Acc. & 97.81 & 94.21 & 69.02 & 84.14 & 90.47 & 32.65 & 77.33 & 57.29 & 67.21 & 75.65 & 74.58 \\
 & ECE & 5.51 & 6.39 & 8.69 & 13.38 & 3.49 & 6.70 & 5.54 & 7.70 & 16.49 & 5.79 & 7.97 \\
  \hdashline[2pt/2pt]
 \multirow{2}{*}{$\textbf{Ours}$}
 & Acc. & \cellcolor{green!20}97.55 & \cellcolor{green!20}94.45 & \cellcolor{green!20}68.95 & \cellcolor{green!20}82.62 & \cellcolor{green!20}90.29 & \cellcolor{green!20}31.57 & \cellcolor{green!20}77.03 & \cellcolor{green!20}68.52 & \cellcolor{green!20}79.78 & \cellcolor{green!20}79.04 & \cellcolor{green!20}76.98 \\
 & ECE & \cellcolor{green!20}3.18 & \cellcolor{green!20}3.46 & \cellcolor{green!20}7.01 & \cellcolor{green!20}6.90 & \cellcolor{green!20}1.36 & \cellcolor{green!20}3.06 & \cellcolor{green!20}3.10 & \cellcolor{green!20}6.41 & \cellcolor{green!20}11.20 & \cellcolor{green!20}3.12 & \cellcolor{green!20}4.88 \\

\hline
\rowcolor{orange!20} \multicolumn{13}{c}{\textbf{PromptSRC} \cite{promptsrc}} \\ 
\hline
\multirow{2}{*}{$\text{PromptSRC}$ \cite{promptsrc}}
 & Acc. & 98.08 & 95.36 & 78.15 & 97.95 & 90.60 & 40.74 & 82.63 & 83.41 & 93.17 & 87.09 & 84.72 \\
 & ECE & 2.31 & 2.64 & 8.65 & 5.15 & 1.17 & 5.26 & 2.75 & 2.56 & 9.27 & 2.84 & 4.26 \\
\hdashline[2pt/2pt]
\multirow{2}{*}{$\text{$\text{ZS-Norm}$}$ \cite{eccv}}
 & Acc. & 98.21 & 95.43 & 77.55 & 97.50 & 90.78 & 40.78 & 82.62 & 81.56 & 48.78 & 85.76 & 79.90 \\
 & ECE & 4.41 & 4.58 & 12.01 & 8.65 & 3.21 & 10.37 & 5.43 & 20.69 & 20.10 & 6.73 & 9.62 \\
 \hdashline[2pt/2pt]
\multirow{2}{*}{$\text{$\text{Penalty}$}$ \cite{eccv}}
 & Acc. & 98.01 & 95.69 & 77.45 & 97.82 & 90.36 & 40.89 & 82.29 & 81.84 & 48.38 & 85.91 & 79.86 \\
 & ECE & 5.41 & 4.78 & 10.61 & 9.65 & 4.51 & 12.47 & 6.43 & 18.69 & 19.10 & 8.98 & 10.06 \\
  \hdashline[2pt/2pt]
\multirow{2}{*}{$\textbf{Ours}$}
 & Acc. & \cellcolor{green!20}98.30 & \cellcolor{green!20}95.57 & \cellcolor{green!20}79.11 & \cellcolor{green!20}98.39 & \cellcolor{green!20}90.67 & \cellcolor{green!20}42.50 & \cellcolor{green!20}82.77 & \cellcolor{green!20}83.83 & \cellcolor{green!20}95.05 & \cellcolor{green!20}86.87 & \cellcolor{green!20}85.31 \\
 & ECE & \cellcolor{green!20}1.03 & \cellcolor{green!20}0.89 & \cellcolor{green!20}8.96 & \cellcolor{green!20}0.97 & \cellcolor{green!20}0.90 & \cellcolor{green!20}6.04 & \cellcolor{green!20}1.34 & \cellcolor{green!20}4.59 & \cellcolor{green!20}4.33 & \cellcolor{green!20}1.39 & \cellcolor{green!20}3.04 \\

\bottomrule
\end{tabular}
 }
\end{table}
%\FloatBarrier
In the main paper, we provide the calibration results for three representative prompt learning methods: CoOp~\cite{coop}, KgCoOp~\cite{kgcoop}, and MaPLe~\cite{maple}. Here we provide the results for additional prompt learning methods including CoCoOp~\cite{cocoop}, ProDA~\cite{proda}, ProGrad~\cite{prograd}, and PromptSRC~\cite{promptsrc} for both base and novel classes to demonstrate the broader applicability and effectiveness of our calibration approach.

\noindent\textbf{Calibration Performance on Base Classes.} Table~\ref{table:supp_base_classes} presents the accuracy and calibration performance on base classes across 10 fine-grained classification benchmarks. The results consistently demonstrate that our calibration approach achieves superior calibration performance (lower ECE) while maintaining competitive accuracy across all evaluated prompt learning methods. Notably, our method shows the most significant improvements with PromptSRC~\cite{promptsrc}, reducing the average ECE from 4.26 to 3.04 without compromising clean accuracy. The consistent improvements across diverse prompt learning architectures validate the generalizability of our approach.
\begin{table}[t!]
%\small
\centering
\caption {\small  \textbf{ Accuracy and calibration performance on novel classes across 10 fine-grained classification benchmarks}.
We report top-1 accuracy (Acc) and Expected Calibration Error (ECE) for multiple prompt-tuning strategies and diverse calibration baselines. Higher Acc. indicates better classification performance, while lower ECE reflects better calibration.}
\label{table:novel_supp}
\resizebox{1\textwidth}{!}{ 
\setlength{\tabcolsep}{4pt} % Reduce column spacing
\begin{tabular}{l c ||c c c c c c c c c c c}
\toprule
\thead{\textbf{Method}} & &  \multicolumn{1}{c}{\thead{\rotatebox{90}{Calt}}} & \multicolumn{1}{c}{\thead{\rotatebox{90}{Pets}}} & \multicolumn{1}{c}{\thead{\rotatebox{90}{Cars}}} & \multicolumn{1}{c}{\thead{\rotatebox{90}{Flow}}} & \multicolumn{1}{c}{\thead{\rotatebox{90}{Food}}} & \multicolumn{1}{c}{\thead{\rotatebox{90}{Air}}} & \multicolumn{1}{c}{\thead{\rotatebox{90}{SUN}}} & \multicolumn{1}{c}{\thead{\rotatebox{90}{DTD}}} & \multicolumn{1}{c}{\thead{\rotatebox{90}{Euro}}} & \multicolumn{1}{c}{\thead{\rotatebox{90}{UCF}}} & \multicolumn{1}{c}{\thead{\rotatebox{90}{\textbf{Avg}}}}\\
\midrule
\multirow{2}{*}{$\text{Zero Shot}$}
 & Acc. & 94.10 & 97.10 & 75.00 & 77.50 & 91.10 & 35.90 & 75.50 & 60.60 & 63.80  & 78.60 & 74.30 \\
 & ECE & 1.60 & 3.42 & 3.31 & 4.91 & 1.83 & 6.55 & 3.48 & 6.86 & 9.12 & 5.52 & 4.43 \\
\hline
\rowcolor{orange!20} \multicolumn{13}{c}{\textbf{CoCoOp}\cite{cocoop} } \\ 
\hline
\multirow{2}{*}{$\text{CoCoOp}$ \cite{cocoop}} 
 & Acc. & 94.51 & 97.69 & 73.26 & 72.27 & 91.02 & 33.47 & 76.54 & 57.65 & 63.14 & 74.89 & 73.44 \\
 & ECE & 1.84 & 2.64 & 1.88 & 9.17 & 1.64 & 10.93 & 2.21 & 11.26 & 9.06 & 4.90 & 5.55 \\
\hdashline[2pt/2pt]
\multirow{2}{*}{$\text{$\text{ZS-Norm}$}$ \cite{eccv}}
 & Acc. & 94.76 & 97.24 & 73.56 & 70.45 & 91.43 & 32.97 & 76.84 & 54.59 & 57.01 & 71.64 & 72.05 \\
 & ECE & 2.63 & 2.87 & 2.11 & 9.39 & 2.16 & 7.21 & 3.99 & 3.91 & 9.24 & 4.36 & 4.79 \\
 \hdashline[2pt/2pt]
\multirow{2}{*}{$\text{$\text{Penalty}$}$ \cite{eccv}}
 & Acc. & 94.29 & 95.62 & 75.07 & 70.52 & 91.46 & 33.59 & 76.85 & 56.76 & 54.50 & 74.08 & 72.27 \\
 & ECE & 2.22 & 5.11 & 5.69 & 5.80 & 4.22 & 5.30 & 3.93 & 11.13 & 11.44 & 3.66 & 5.85 \\
  \hdashline[2pt/2pt]
\multirow{2}{*}{$\text{DAC}$}
 & Acc. & - & - & - & - & - & - & - & - & - & - & - \\
 & ECE & 3.65 & 2.43 & 2.21 & 7.74 & 1.64 & 9.03 & 1.09 & 7.47 & 13.49 & 2.70 & 5.15 \\
  \hdashline[2pt/2pt]
\multirow{2}{*}{$\textbf{Ours}$}
 & Acc. & \cellcolor{green!20}94.43 & \cellcolor{green!20}97.45 & \cellcolor{green!20}74.71 & \cellcolor{green!20}71.91 & \cellcolor{green!20}91.62 & \cellcolor{green!20}33.97 & \cellcolor{green!20}76.41 & \cellcolor{green!20}56.40 & \cellcolor{green!20}61.67 & \cellcolor{green!20}76.40 & \cellcolor{green!20}73.50 \\
 & ECE & \cellcolor{green!20}1.51 & \cellcolor{green!20}2.58 & \cellcolor{green!20}3.22 & \cellcolor{green!20}6.36 & \cellcolor{green!20}0.96 & \cellcolor{green!20}8.85 & \cellcolor{green!20}0.87 & \cellcolor{green!20}4.77 & \cellcolor{green!20}6.68 & \cellcolor{green!20}2.76 & \cellcolor{green!20}3.86 \\

\hline
\rowcolor{orange!20} \multicolumn{13}{c}{\textbf{ProDA} \cite{proda}} \\ 
\hline
\multirow{2}{*}{$\text{ProDA}$ \cite{proda}}
 & Acc. & 93.99 & 96.90 & 73.16 & 72.51 & 90.64 & 31.35 & 65.02 & 53.99 & 51.86 & 72.76 & 70.22\\
 & ECE & 3.22 & 1.96 & 3.18 & 8.51 & 0.84 & 15.03 & 14.08 & 16.90 & 21.85 & 4.74 & 9.03\\
\hdashline[2pt/2pt]
\multirow{2}{*}{$\text{$\text{ZS-Norm}$}$ \cite{eccv}}
 & Acc. & 93.81 & 97.28 & 72.53 & 72.81 & 90.44 & 30.09 & 66.59 & 52.13 & 57.77 & 72.67 & 71.01 \\
 & ECE & 2.36 & 2.42 & 2.06 & 8.34 & 0.94 & 10.76 & 12.12 & 7.65 & 8.75 & 4.41 & 5.98 \\
 \hdashline[2pt/2pt]
\multirow{2}{*}{$\text{$\text{Penalty}$}$ \cite{eccv}}
 & Acc. & 93.92 & 97.20 & 73.39 & 73.57 & 90.70 & 32.45 & 67.73 & 50.48 & 60.05 & 72.49 & 71.20 \\
 & ECE & 1.53 & 6.14 & 3.76 & 4.36 & 3.47 & 7.82 & 2.51 & 4.96 & 14.47 & 3.86 & 5.29 \\
  \hdashline[2pt/2pt]
\multirow{2}{*}{$\text{DAC}$}
 & Acc. & - & - & - & - & - & - & - & - & - & - & - \\
 & ECE & 4.87 & 4.72 & 3.28 & 6.32 & 0.70 & 7.40 & 1.06 & 5.68 & 3.33 & 4.14 & 4.15 \\
  \hdashline[2pt/2pt]
 \multirow{2}{*}{$\textbf{Ours}$}
 & Acc. & \cellcolor{green!20}93.56 & \cellcolor{green!20}97.56 & \cellcolor{green!20}73.81 & \cellcolor{green!20}72.74 & \cellcolor{green!20}91.14 & \cellcolor{green!20}30.57 & \cellcolor{green!20}66.18 & \cellcolor{green!20}53.82 & \cellcolor{green!20}58.58 & \cellcolor{green!20}75.79 & \cellcolor{green!20}71.38 \\
 & ECE & \cellcolor{green!20}1.48 & \cellcolor{green!20}3.25 & \cellcolor{green!20}2.77 & \cellcolor{green!20}5.12 & \cellcolor{green!20}1.01 & \cellcolor{green!20}6.78 & \cellcolor{green!20}1.90 & \cellcolor{green!20}4.60 & \cellcolor{green!20}4.91 & \cellcolor{green!20}2.35 & \cellcolor{green!20}3.42 \\

\hline
\rowcolor{orange!20} \multicolumn{13}{c}{\textbf{ProGrad} \cite{prograd}} \\ 
\hline
\multirow{2}{*}{$\text{ProGrad}$ \cite{prograd}}
 & Acc. & 94.76 & 97.32 & 74.85 & 75.29 & 91.06 & 34.43 & 75.42 & 56.44 & 61.98 & 78.74 & 74.03\\
 & ECE & 1.67 & 3.52 & 2.68 & 7.46 & 1.76 & 9.21 & 2.05 & 4.48 & 8.83 & 3.57 & 4.52\\
\hdashline[2pt/2pt]
\multirow{2}{*}{$\text{$\text{ZS-Norm}$}$ \cite{eccv}}
 & Acc. & 94.43 & 97.37 & 74.97 & 75.18 & 91.18 & 31.49 & 74.79 & 55.80 & 67.97 & 77.39 & 74.06 \\
 & ECE & 1.80 & 5.11 & 5.32 & 3.73 & 2.68 & 3.79 & 7.10 & 12.79 & 12.83 & 4.83 & 6.00 \\
 \hdashline[2pt/2pt]
\multirow{2}{*}{$\text{$\text{Penalty}$}$ \cite{eccv}}
 & Acc. & 94.87 & 96.98 & 75.81 & 74.54 & 91.05 & 34.55 & 75.03 & 53.74 & 66.97 & 76.80 & 74.03 \\
 & ECE & 1.90 & 5.54 & 5.07 & 4.80 & 3.08 & 5.31 & 3.08 & 4.96 & 14.86 & 5.05 & 5.37 \\
  \hdashline[2pt/2pt]
\multirow{2}{*}{$\text{DAC}$}
 & Acc. & - & - & - & - & - & - & - & - & - & - & - \\
 & ECE & 1.97 & 3.31 & 2.29 & 5.04 & 1.85 & 10.46 & 1.32 & 3.49 & 6.90 & 2.42 & 3.91 \\
  \hdashline[2pt/2pt]
\multirow{2}{*}{$\textbf{Ours}$}
 & Acc. & \cellcolor{green!20}94.29 & \cellcolor{green!20}97.48 & \cellcolor{green!20}75.09 & \cellcolor{green!20}74.66 & \cellcolor{green!20}91.21 & \cellcolor{green!20}32.87 & \cellcolor{green!20}74.81 & \cellcolor{green!20}55.60 & \cellcolor{green!20}67.91 & \cellcolor{green!20}78.53 & \cellcolor{green!20}74.25 \\
 & ECE & \cellcolor{green!20}1.03 & \cellcolor{green!20}3.26 & \cellcolor{green!20}1.98 & \cellcolor{green!20}5.14 & \cellcolor{green!20}1.47 & \cellcolor{green!20}9.34 & \cellcolor{green!20}2.32 & \cellcolor{green!20}3.30 & \cellcolor{green!20}6.06 & \cellcolor{green!20}3.06 & \cellcolor{green!20}3.70 \\

\hline
\rowcolor{orange!20} \multicolumn{13}{c}{\textbf{PromptSRC} \cite{promptsrc}} \\ 
\hline
\multirow{2}{*}{$\text{PromptSRC}$ \cite{promptsrc}}
 & Acc. & 94.21 & 97.31 & 75.58 & 77.28 & 91.51 & 29.73 & 78.79 & 61.03 & 74.72 & 77.86 & 75.80 \\
 & ECE & 1.51 & 3.26 & 2.06 & 5.50 & 1.77 & 12.92 & 1.07 & 6.68 & 8.08 & 2.81 & 4.57 \\
\hdashline[2pt/2pt]
\multirow{2}{*}{$\text{ZS-Norm}$ \cite{eccv}}
 & Acc. & 94.18 & 97.61 & 74.80 & 76.31 & 91.60 & 36.23 & 78.80 & 59.02 & 37.07 & 77.25 & 72.29 \\
 & ECE & 2.06 & 4.30 & 3.63 & 5.86 & 3.66 & 4.64 & 3.57 & 12.29 & 12.56 & 3.95 & 5.65 \\
\hdashline[2pt/2pt]
\multirow{2}{*}{$\text{$\text{Penalty}$}$ \cite{eccv}}
 & Acc. & 94.17 & 97.55 & 74.12 & 76.34 & 91.99 & 36.63 & 78.14 & 59.62 & 37.37 & 77.76 & 72.37 \\
 & ECE & 3.06 & 5.40 & 4.63 & 4.86 & 4.98 & 4.94 & 4.53 & 11.29 & 10.66 & 4.65 & 5.90 \\
  \hdashline[2pt/2pt]
\multirow{2}{*}{$\text{DAC}$}
 & Acc. & - & - & - & - & - & - & - & - & - & - & - \\
 & ECE & 1.58 & 2.98 & 2.39 & 5.03 & 1.55 & 8.55 & 0.79 & 5.50 & 7.24 & 2.46 & 3.81 \\
  \hdashline[2pt/2pt]
 \multirow{2}{*}{$\textbf{Ours}$}
 & Acc. & \cellcolor{green!20}94.29 & \cellcolor{green!20}97.28 & \cellcolor{green!20}74.49 & \cellcolor{green!20}75.44 & \cellcolor{green!20}91.68 & \cellcolor{green!20}36.85 & \cellcolor{green!20}78.39 & \cellcolor{green!20}57.53 & \cellcolor{green!20}72.64 & \cellcolor{green!20}77.72 & \cellcolor{green!20}75.63 \\
 & ECE & \cellcolor{green!20}1.14 & \cellcolor{green!20}1.19 & \cellcolor{green!20}2.39 & \cellcolor{green!20}5.44 & \cellcolor{green!20}0.72 & \cellcolor{green!20}9.26 & \cellcolor{green!20}0.77 & \cellcolor{green!20}6.81 & \cellcolor{green!20}8.37 & \cellcolor{green!20}1.89 & \cellcolor{green!20}3.80 \\

\bottomrule
\end{tabular}
 }
\end{table}
\noindent \textbf{Calibration Performance on Novel Classess.}
We further evaluate the calibration performance on novel classes to assess the generalization capability of our approach when dealing with unseen categories during training. Table~\ref{table:novel_supp} presents the accuracy and calibration performance on novel classes across the same 10 benchmarks. Our method demonstrates consistent calibration improvements across all prompt learning methods when evaluated on novel classes. The results show that our approach effectively generalizes to unseen categories, with particularly notable improvements in ECE reduction. For instance, with ProDA~\cite{proda}, our method reduces the average ECE from 9.03 to 3.42 while maintaining comparable clean accuracy (70.22 vs 71.38). Similarly, with CoCoOp~\cite{cocoop}, we achieve an ECE reduction from 5.55 to 3.86. These results highlight the robustness of our calibration approach in handling the challenging scenario of novel class prediction, where models are more prone to overconfidence due to limited training exposure.
%\vspace{-10em}

\begin{table}[t!]
%\small
\centering
% \caption {\small \textbf{Fine-Grained Classification (Base Classes)}}
\caption {\small  \textbf{ Accuracy and calibration performance on base classes across 4 natural distribution shift datasets}.
We report top-1 accuracy (Acc) and Expected Calibration Error (ECE) for MaPLe~\cite{maple}. Higher Acc. indicates better classification performance, while lower ECE reflects better calibration.}
\label{table:base_distribution_ablation}
\resizebox{0.7\textwidth}{!}{ 
\setlength{\tabcolsep}{4pt} % Reduce column spacing
\begin{tabular}{l c ||c c c c c c c c c c c c}
\toprule
\thead{\textbf{Method}} & & \multicolumn{1}{c}{\thead{\rotatebox{90}{INet-V2}}} & \multicolumn{1}{c}{\thead{\rotatebox{90}{INet-S}}} & \multicolumn{1}{c}{\thead{\rotatebox{90}{INet-A}}} & \multicolumn{1}{c}{\thead{\rotatebox{90}{INet-R}}} & \multicolumn{1}{c}{\thead{\rotatebox{90}{\textbf{Avg}}}}\\
\midrule
\hline
\rowcolor{orange!20} \multicolumn{14}{c}{\textbf{MaPLe} \cite{maple}} \\ 
\hline
\multirow{2}{*}{$\text{MaPLe}$~\cite{maple}}
 & Acc. & 67.35 & 53.35 & 68.31 & 85.22 & 68.56 \\
 & ECE & 3.14 & 3.94 & 2.52 & 3.14 & 3.19 \\
\hdashline[2pt/2pt]
\multirow{2}{*}{$\text{ZS-Norm}$~\cite{eccv}}
 & Acc. & 66.15 & 53.24 & 68.41 & 85.17 & 68.49 \\
 & ECE & 3.41 & 4.12 & 6.98 & 7.13 & 5.41 \\
\hdashline[2pt/2pt]
\multirow{2}{*}{$\text{Penalty}$~\cite{eccv}}
 & Acc. & 66.72 & 53.04 & 68.61 & 85.17 & 68.39 \\
 & ECE & 3.61 & 4.62 & 7.48 & 7.33 & 5.76 \\
\hdashline[2pt/2pt]
\multirow{2}{*}{$\textbf{Ours}$}
 & Acc. & \cellcolor{green!20}67.19 & \cellcolor{green!20}53.15 & \cellcolor{green!20}67.86 & \cellcolor{green!20}85.28 & \cellcolor{green!20}68.37 \\
 & ECE & \cellcolor{green!20}3.09 & \cellcolor{green!20}3.91 & \cellcolor{green!20}2.21 & \cellcolor{green!20}2.05 & \cellcolor{green!20}2.82 \\
 
\hline
\bottomrule
\end{tabular}
}
\end{table}
\begin{table}[t!]
%\small
\centering
% \caption {\small \textbf{Fine-Grained Classification (Base Classes)}}
\caption {\small  \textbf{ Accuracy and calibration performance on novel classes across 4 natural distribution shift datasets}.
We report top-1 accuracy (Acc) and Expected Calibration Error (ECE) for MaPLe ~\cite{maple}. Acc. indicates better classification performance, while lower ECE reflects better calibration.}
\label{table:novel_distribution_ablation}
\resizebox{0.7\textwidth}{!}{ 
\setlength{\tabcolsep}{4pt} % Reduce column spacing
\begin{tabular}{l c ||c c c c c c c c c c c c}
\toprule
\thead{\textbf{Method}} & & \multicolumn{1}{c}{\thead{\rotatebox{90}{INet-V2}}} & \multicolumn{1}{c}{\thead{\rotatebox{90}{INet-S}}} & \multicolumn{1}{c}{\thead{\rotatebox{90}{INet-A}}} & \multicolumn{1}{c}{\thead{\rotatebox{90}{INet-R}}} & \multicolumn{1}{c}{\thead{\rotatebox{90}{\textbf{Avg}}}}\\
\midrule
\hline
\rowcolor{orange!20} \multicolumn{14}{c}{\textbf{MaPLe}~\cite{maple}} \\ 
\hline
\multirow{2}{*}{$\text{MaPLe}$~\cite{maple}}
 & Acc. & 67.36 & 53.36 & 68.31 & 85.22 & 68.56 \\
 & ECE & 3.16 & 3.73 & 2.52 & 3.14 & 3.14 \\
\hdashline[2pt/2pt]
\multirow{2}{*}{$\text{ZS-Norm}$~\cite{eccv}}
 & Acc. & 66.75 & 53.34 & 68.60 & 85.16 & 68.46 \\
 & ECE & 3.54 & 4.01 & 6.47 & 7.21 & 5.31 \\
\hdashline[2pt/2pt]
\multirow{2}{*}{$\text{Penalty}$~\cite{eccv}}
 & Acc. & 66.75 & 53.11 & 68.60 & 85.16 & 68.41 \\
 & ECE & 3.64 & 4.11 & 7.47 & 7.32 & 5.64 \\
\hdashline[2pt/2pt]
\multirow{2}{*}{$\textbf{Ours}$}
 & Acc. & \cellcolor{green!20}67.20 & \cellcolor{green!20}53.21 & \cellcolor{green!20}67.86 & \cellcolor{green!20}85.28 & \cellcolor{green!20}68.39 \\
 & ECE & \cellcolor{green!20}3.11 & \cellcolor{green!20}3.63 & \cellcolor{green!20}2.20 & \cellcolor{green!20}2.05 & \cellcolor{green!20}2.75 \\
 
\hline
\bottomrule
\end{tabular}
}
\end{table}

%\FloatBarrier 
\section{Results on Natural Distribution Shifts} \label{sec:dis_shift}

Here, we provide results on out-of-distribution datasets for the ImageNet-A~\cite{hendrycks2021natural}, ImageNet-V2~\cite{shankar2020evaluating}, ImageNet-R~\cite{hendrycks2021many}, and ImageNet-S~\cite{wang2019learning} datasets demonstrating the robustness of our calibration approach under natural distribution shifts.

\textbf{Base Classes.} Table~\ref{table:base_distribution_ablation} presents the accuracy and calibration performance on base classes across these 4 natural distribution shift datasets. Our method consistently outperforms baseline calibration approaches across all datasets. Notably, our approach achieves superior calibration with an average ECE of 2.82 compared to the vanilla MaPLe baseline (3.19), ZS-Norm (5.41), and Penalty (5.76). The improvements are particularly pronounced on challenging datasets like ImageNet-A and ImageNet-R, where our method reduces ECE from 2.52 to 2.21 and from 3.14 to 2.05, respectively, while maintaining competitive accuracy.

\textbf{Novel Classes.} Table~\ref{table:novel_distribution_ablation} shows the corresponding results on novel classes under distribution shift. The consistent performance across both base and novel classes demonstrates the generalization capability of our calibration approach. Our method achieves an average ECE of 2.75 on novel classes, significantly outperforming ZS-Norm (5.31) and Penalty (5.64) baselines. The robustness across different types of distribution shifts, including adversarial examples (ImageNet-A), renditions (ImageNet-R), and sketch-like images (ImageNet-S), validates that our approach addresses fundamental calibration issues rather than dataset-specific artifacts.

These results are particularly important for real-world deployment scenarios where models encounter data that differs from the training distribution. The consistent calibration improvements across diverse distribution shifts demonstrate that our method provides reliable confidence estimates even under challenging out-of-distribution conditions.

\section{Additional Results: ACE and MCE Performance Metrics} \label{sec:ace_mce}
\begin{table}[t!]
%\small
\centering
\caption {\small  \textbf{Calibration performance on base classes across 10 fine-grained classification benchmarks}.
We report Maximum Calibration Error (MCE) and Adaptive Calibration Error (ACE) for multiple prompt-tuning strategies and diverse calibration baselines. Lower MCE and ACE reflects better calibration.}
\label{table:othermetrics_base}
\resizebox{\textwidth}{!}{ 
\setlength{\tabcolsep}{4pt} % Reduce column spacing
\begin{tabular}{l c ||c c c c c c c c c c c c}
\toprule
\thead{\textbf{Method}} & &  \multicolumn{1}{c}{\thead{\rotatebox{90}{Calt}}} & \multicolumn{1}{c}{\thead{\rotatebox{90}{Pets}}} & \multicolumn{1}{c}{\thead{\rotatebox{90}{Cars}}} & \multicolumn{1}{c}{\thead{\rotatebox{90}{Flow}}} & \multicolumn{1}{c}{\thead{\rotatebox{90}{Food}}} & \multicolumn{1}{c}{\thead{\rotatebox{90}{Air}}} & \multicolumn{1}{c}{\thead{\rotatebox{90}{SUN}}} & \multicolumn{1}{c}{\thead{\rotatebox{90}{DTD}}} & \multicolumn{1}{c}{\thead{\rotatebox{90}{Euro}}} & \multicolumn{1}{c}{\thead{\rotatebox{90}{UCF}}} & \multicolumn{1}{c}{\thead{\rotatebox{90}{\textbf{Avg}}}}\\
\midrule
% \multirow{2}{*}{$\text{Zero Shot}$}
%  & Acc. & 72.40 & 97.20 & 91.30 & 63.60 & 71.80 & 90.10 & 27.70 & 69.40 & 53.00 & 57.00 & 71.00 & 69.50 \\
%  & ECE & 1.51 & 6.49 & 2.25 & 3.74 & 3.11 & 1.57 & 3.03 & 1.59 & 4.53 & 8.35 & 3.24 & 3.58 \\
\hline
\rowcolor{orange!20} \multicolumn{14}{c}{\textbf{CoOp}\cite{coop} } \\ 
\hline
\multirow{2}{*}{$\text{CoOp}$ \cite{coop}} 
 & MCE & 0.24 & 0.31 & 0.91 & 2.46 & 1.72 & 4.77 & 3.99 & 5.68 & 0.60 & 3.34 & 2.40  \\
 & ACE & 0.44 & 0.62 & 3.65 & 4.67 & 3.65 & 25.70 & 8.11 & 12.01 & 1.95 & 6.41 & 6.72 \\
\hdashline[2pt/2pt]
\multirow{2}{*}{$\text{$\text{ZS-Norm}$}$ \cite{eccv}}
 & MCE & 1.40 & 2.17 & 2.20 & 2.76 & 0.90 & 4.34 & 0.82 & 14.76 & 11.82 & 0.87 & 4.20 \\
 & ACE & 4.32 & 7.70 & 11.26 & 11.12 & 3.14 & 13.05 & 4.26 & 49.53 & 36.04 & 3.36 & 14.38 \\
\hdashline[2pt/2pt]
\multirow{2}{*}{$\text{$\text{Penalty}$}$ \cite{eccv}}
 & MCE & 1.62 & 1.85 & 2.13 & 2.50 & 1.38 & 2.23 & 0.94 & 4.30 & 11.34 & 1.63 & 2.99 \\
 & ACE & 4.75 & 6.41 & 10.01 & 9.36 & 5.98 & 8.61 & 4.61 & 21.48 & 20.86 & 7.09 & 9.92 \\
\hdashline[2pt/2pt]
 \multirow{2}{*}{$\textbf{Ours}$}
 & MCE & \cellcolor{green!20}0.33 & \cellcolor{green!20}1.01 & \cellcolor{green!20}1.87 & \cellcolor{green!20}1.68 & \cellcolor{green!20}0.12 & \cellcolor{green!20}1.21 & \cellcolor{green!20}0.31 & \cellcolor{green!20}0.56 & \cellcolor{green!20}1.67 & \cellcolor{green!20}0.22 & \cellcolor{green!20}0.90 \\
 & ACE & \cellcolor{green!20}0.98 & \cellcolor{green!20}2.10 & \cellcolor{green!20}7.55 & \cellcolor{green!20}4.94 & \cellcolor{green!20}0.21 & \cellcolor{green!20}2.40 & \cellcolor{green!20}1.30 & \cellcolor{green!20}2.01 & \cellcolor{green!20}5.12 & \cellcolor{green!20}1.15 & \cellcolor{green!20}2.78 \\

\hline
\rowcolor{orange!20} \multicolumn{14}{c}{\textbf{MaPLe} \cite{maple}} \\ 
\hline
\multirow{2}{*}{$\text{MaPLe}$ \cite{maple}}
 & MCE & 0.51 & 0.60 & 1.68 & 1.29 & 0.34 & 0.91 & 0.19 & 1.04 & 0.74 & 0.56 & 0.79  \\
 & ACE & 2.14 & 1.19 & 6.91 & 3.21 & 0.73 & 2.95 & 1.17 & 3.71 & 2.94 & 1.49 & 2.64  \\
\hdashline[2pt/2pt]
\multirow{2}{*}{$\text{$\text{ZS-Norm}$}$ \cite{eccv}}
 & MCE & 1.28 & 1.01 & 3.79 & 1.45 & 2.25 & 1.95 & 2.27 & 2.80 & 1.24 & 0.61 & 1.87  \\
 & ACE & 5.28 & 3.21 & 20.73 & 7.21 & 11.17 & 6.98 & 8.59 & 12.37 & 6.91 & 3.40 & 8.59  \\
 \hdashline[2pt/2pt]
\multirow{2}{*}{$\text{$\text{Penalty}$}$ \cite{eccv}}
 & MCE & 1.88 & 1.85 & 2.30 & 3.15 & 1.11 & 2.27 & 1.24 & 3.69 & 0.61 & 1.45 & 1.96  \\
 & ACE & 5.68 & 6.29 & 11.17 & 11.97 & 3.71 & 8.72 & 6.81 & 20.23 & 3.10 & 7.61 & 8.53  \\
  \hdashline[2pt/2pt]
 \multirow{2}{*}{$\textbf{Ours}$}
 & MCE & 0.62 & 0.34 & 1.20 & 0.87 & 1.47 & 0.62 & 1.11 & 1.50 & 0.60 & 0.94 & 0.93  \\
 & ACE & 1.62 & 0.80 & 3.98 & 2.18 & 3.66 & 0.94 & 4.38 & 7.55 & 1.54 & 1.34 & 2.80  \\

\hline
\rowcolor{orange!20} \multicolumn{14}{c}{\textbf{KGCoOp} \cite{kgcoop}} \\ 
\hline
\multirow{2}{*}{$\text{KGCoOp}$ \cite{kgcoop}}
& MCE & 1.14 & 1.17 & 2.23 & 2.75 & 0.62 & 1.17 & 0.97 & 1.61 & 3.31 & 1.25 & 1.62  \\
 & ACE & 2.56 & 2.95 & 10.14 & 11.95 & 1.59 & 2.95 & 4.91 & 8.39 & 11.90 & 4.59 & 6.19  \\
\hdashline[2pt/2pt]
\multirow{2}{*}{$\text{$\text{ZS-Norm}$}$ \cite{eccv}}
 & MCE & 1.31 & 1.13 & 2.28 & 3.02 & 0.64 & 3.12 & 1.34 & 3.85 & 4.06 & 1.12 & 2.19  \\
 & ACE & 2.98 & 3.06 & 10.58 & 13.00 & 1.69 & 9.59 & 6.51 & 20.40 & 15.58 & 5.75 & 8.91  \\
 \hdashline[2pt/2pt]
\multirow{2}{*}{$\text{$\text{Penalty}$}$ \cite{eccv}}
 & MCE & 1.27 & 1.40 & 2.10 & 3.03 & 0.81 & 1.67 & 1.20 & 2.87 & 3.30 & 1.33 & 1.90 \\
 & ACE & 4.20 & 4.61 & 9.96 & 12.53 & 2.44 & 6.41 & 5.92 & 10.66 & 13.14 & 6.04 & 7.59  \\
  \hdashline[2pt/2pt]
\multirow{2}{*}{$\textbf{Ours}$}
 & MCE & \cellcolor{green!20}0.86 & \cellcolor{green!20}1.05 & \cellcolor{green!20}1.12 & \cellcolor{green!20}2.01 & \cellcolor{green!20}0.55 & \cellcolor{green!20}1.94 & \cellcolor{green!20}0.81 & \cellcolor{green!20}1.12 & \cellcolor{green!20}3.61 & \cellcolor{green!20}0.86 & \cellcolor{green!20}1.39 \\
 & ACE & \cellcolor{green!20}1.83 & \cellcolor{green!20}2.83 & \cellcolor{green!20}8.1 & \cellcolor{green!20}11.21 & \cellcolor{green!20}1.42 & \cellcolor{green!20}5.11 & \cellcolor{green!20}4.15 & \cellcolor{green!20}7.21 & \cellcolor{green!20}12.65 & \cellcolor{green!20}3.95 & \cellcolor{green!20}5.85 \\

\bottomrule
\end{tabular}
 }
\end{table}
\begin{table}[t!]
%\small
\centering
\caption {\small  \textbf{ Calibration performance on novel classes across 10 fine-grained classification benchmarks}.
We report Maximum Calibration Error (MCE) and Adaptive Calibration Error (ACE) for multiple prompt-tuning strategies and diverse calibration baselines. Lower MCE and ACE reflects better calibration.}
\label{table:othermetrics_novel}
\resizebox{\textwidth}{!}{ 
\setlength{\tabcolsep}{4pt} % Reduce column spacing
\begin{tabular}{l c ||c c c c c c c c c c c c}
\toprule
\thead{\textbf{Method}} & &  \multicolumn{1}{c}{\thead{\rotatebox{90}{Calt}}} & \multicolumn{1}{c}{\thead{\rotatebox{90}{Pets}}} & \multicolumn{1}{c}{\thead{\rotatebox{90}{Cars}}} & \multicolumn{1}{c}{\thead{\rotatebox{90}{Flow}}} & \multicolumn{1}{c}{\thead{\rotatebox{90}{Food}}} & \multicolumn{1}{c}{\thead{\rotatebox{90}{Air}}} & \multicolumn{1}{c}{\thead{\rotatebox{90}{SUN}}} & \multicolumn{1}{c}{\thead{\rotatebox{90}{DTD}}} & \multicolumn{1}{c}{\thead{\rotatebox{90}{Euro}}} & \multicolumn{1}{c}{\thead{\rotatebox{90}{UCF}}} & \multicolumn{1}{c}{\thead{\rotatebox{90}{\textbf{Avg}}}}\\
\midrule
% \multirow{2}{*}{$\text{Zero Shot}$}
%  & Acc. & 72.40 & 97.20 & 91.30 & 63.60 & 71.80 & 90.10 & 27.70 & 69.40 & 53.00 & 57.00 & 71.00 & 69.50 \\
%  & ECE & 1.51 & 6.49 & 2.25 & 3.74 & 3.11 & 1.57 & 3.03 & 1.59 & 4.53 & 8.35 & 3.24 & 3.58 \\
\hline
\rowcolor{orange!20} \multicolumn{14}{c}{\textbf{CoOp}\cite{coop} } \\ 
\hline
\multirow{2}{*}{$\text{CoOp}$ \cite{coop}} 
 & MCE & 2.61 & 0.64 & 2.38 & 5.48 & 1.44 & 5.01 & 4.35 & 7.06 & 4.41 & 6.11 & 3.95  \\
 & ACE & 3.47 & 1.67 & 12.45 & 18.33 & 3.84 & 28.41 & 13.92 & 26.88 & 12.73 & 19.17 & 14.09  \\
\hdashline[2pt/2pt]
\multirow{2}{*}{$\text{DAC}$ \cite{dac}}
 & MCE & 1.50 & 0.66 & 1.21 & 2.08 & 0.49 & 3.57 & 1.03 & 2.24 & 3.03 & 1.70 & 1.76  \\
 & ACE & 2.60 & 1.70 & 5.17 & 10.18 & 1.75 & 17.27 & 4.00 & 10.51 & 8.58 & 8.63 & 7.04  \\
\hdashline[2pt/2pt]
\multirow{2}{*}{$\text{$\text{ZS-Norm}$}$ \cite{eccv}}
 & MCE & 1.46 & 2.06 & 0.75 & 1.08 & 0.82 & 2.24 & 0.55 & 9.32 & 11.82 & 1.14 & 3.12 \\
 & ACE & 2.59 & 7.9 & 3.01 & 5.93 & 3.3 & 9.86 & 2.27 & 21.97 & 37.04 & 3.91 & 9.78 \\
 \hdashline[2pt/2pt]
\multirow{2}{*}{$\text{$\text{Penalty}$}$ \cite{eccv}}
 & MCE & 0.92 & 2.1 & 0.68 & 1.16 & 1.00 & 2.46 & 0.76 & 0.9 & 7.54 & 1.36 & 1.89 \\
 & ACE & 2.23 & 7.32 & 2.69 & 5.46 & 4.69 & 7.44 & 2.91 & 4.35 & 14.94 & 4.51 & 5.65  \\
\hdashline[2pt/2pt]
 \multirow{2}{*}{$\textbf{Ours}$}
  & MCE & \cellcolor{green!20}1.05 & \cellcolor{green!20}1.26 & \cellcolor{green!20}0.55 & \cellcolor{green!20}0.98 & \cellcolor{green!20}0.25 & \cellcolor{green!20}2.9 & \cellcolor{green!20}0.64 & \cellcolor{green!20}1.97 & \cellcolor{green!20}3.9 & \cellcolor{green!20}1.49 & \cellcolor{green!20}1.42  \\
 & ACE & \cellcolor{green!20}2.3 & \cellcolor{green!20}2.92 & \cellcolor{green!20}2.02 & \cellcolor{green!20}3.67 & \cellcolor{green!20}0.91 & \cellcolor{green!20}10.47 & \cellcolor{green!20}2.99 & \cellcolor{green!20}9.6 & \cellcolor{green!20}11.24 & \cellcolor{green!20}5.04 & \cellcolor{green!20}4.83  \\

\hline
\rowcolor{orange!20} \multicolumn{14}{c}{\textbf{MaPLe} \cite{maple}} \\ 
\hline
\multirow{2}{*}{$\text{MaPLe}$ \cite{maple}}
 & MCE & 0.55 & 1.04 & 0.56 & 5.02 & 0.37 & 1.48 & 0.65 & 3.50 & 1.66 & 0.66 & 1.55  \\
 & ACE & 1.26 & 2.44 & 2.83 & 12.67 & 1.12 & 7.27 & 2.49 & 14.90 & 7.90 & 2.81 & 5.57  \\
\hdashline[2pt/2pt]
\multirow{2}{*}{$\text{DAC}$ \cite{dac}}
& MCE & 0.39 & 1.06 & 0.77 & 4.42 & 0.50 & 2.26 & 0.40 & 1.95 & 2.55 & 0.66 & 1.50  \\
 & ACE & 1.19 & 2.44 & 2.57 & 11.28 & 1.48 & 8.90 & 1.37 & 8.24 & 9.12 & 2.32 & 4.89  \\
\hdashline[2pt/2pt]
\multirow{2}{*}{$\text{$\text{ZS-Norm}$}$ \cite{eccv}}
 & MCE & 0.84 & 1.23 & 2.13 & 1.76 & 1.32 & 1.92 & 1.07 & 1.40 & 0.91 & 7.51 & 2.01  \\
 & ACE & 1.62 & 4.15 & 9.46 & 7.77 & 3.83 & 7.85 & 3.77 & 6.11 & 4.84 & 13.73 & 6.31  \\
 \hdashline[2pt/2pt]
\multirow{2}{*}{$\text{$\text{Penalty}$}$ \cite{eccv}}
 & MCE & 0.84 & 1.72 & 1.40 & 1.32 & 1.11 & 1.07 & 0.91 & 2.73 & 7.11 & 1.76 & 2.00  \\
 & ACE & 1.92 & 7.60 & 6.41 & 4.83 & 4.05 & 3.47 & 4.74 & 10.46 & 16.73 & 8.67 & 6.89  \\
  \hdashline[2pt/2pt]
 \multirow{2}{*}{$\textbf{Ours}$}
 & MCE & \cellcolor{green!20}0.44 & \cellcolor{green!20}0.52 & \cellcolor{green!20}1.61 & \cellcolor{green!20}1.08 & \cellcolor{green!20}1.70 & \cellcolor{green!20}0.88 & \cellcolor{green!20}2.66 & \cellcolor{green!20}0.55 & \cellcolor{green!20}0.14 & \cellcolor{green!20}0.87 & \cellcolor{green!20}1.05  \\
 & ACE & \cellcolor{green!20}0.32 & \cellcolor{green!20}0.95 & \cellcolor{green!20}6.38 & \cellcolor{green!20}3.13 & \cellcolor{green!20}11.27 & \cellcolor{green!20}2.07 & \cellcolor{green!20}8.49 & \cellcolor{green!20}2.48 & \cellcolor{green!20}0.58 & \cellcolor{green!20}4.74 & \cellcolor{green!20}4.04  \\

\hline
\rowcolor{orange!20} \multicolumn{14}{c}{\textbf{KGCoOp} \cite{kgcoop}} \\ 
\hline
\multirow{2}{*}{$\text{KGCoOp}$ \cite{kgcoop}}
& MCE & 0.53 & 1.16 & 0.9 & 1.06 & 0.74 & 1.78 & 0.39 & 1.22 & 3.37 & 0.69 & 1.18  \\
 & ACE & 1.22 & 3.26 & 3.36 & 5.45 & 2.01 & 5.86 & 1.83 & 5.02 & 8.69 & 2.59 & 3.93  \\
 \hdashline[2pt/2pt]
\multirow{2}{*}{$\text{DAC}$ \cite{dac}}
& MCE & 0.62 & 1.21 & 0.85 & 1.37 & 0.64 & 3.02 & 0.42 & 1.33 & 1.75 & 0.77 & 1.20  \\
 & ACE & 1.56 & 3.02 & 3.11 & 6.61 & 1.93 & 11.74 & 1.82 & 7.26 & 6.63 & 2.70 & 4.64  \\
\hdashline[2pt/2pt]
\multirow{2}{*}{$\text{$\text{ZS-Norm}$}$~\cite{eccv}}
 & MCE & 0.58 & 1.17 & 0.94 & 1.06 & 0.75 & 2.97 & 0.59 & 1.90 & 1.52 & 0.97 & 1.25  \\
 & ACE & 1.33 & 3.30 & 3.86 & 5.31 & 2.10 & 8.39 & 3.30 & 5.95 & 6.51 & 3.81 & 4.39  \\
 \hdashline[2pt/2pt]
\multirow{2}{*}{$\text{$\text{Penalty}$}$~\cite{eccv}}
 & MCE & 0.56 & 1.43 & 0.99 & 1.38 & 0.87 & 1.34 & 0.58 & 1.68 & 3.72 & 1.28 & 1.38  \\
 & ACE & 1.19 & 3.51 & 3.89 & 5.14 & 2.64 & 5.00 & 3.35 & 5.63 & 13.86 & 4.23 & 4.84  \\
  \hdashline[2pt/2pt]
\multirow{2}{*}{$\textbf{Ours}$}
 & MCE & \cellcolor{green!20}0.51 & \cellcolor{green!20}1.21 & \cellcolor{green!20}0.85 & \cellcolor{green!20}0.98 & \cellcolor{green!20}0.68 & \cellcolor{green!20}2.78 & \cellcolor{green!20}0.47 & \cellcolor{green!20}1.97 & \cellcolor{green!20}1.35 & \cellcolor{green!20}0.91 & \cellcolor{green!20}1.17 \\
 & ACE & \cellcolor{green!20}1.1 & \cellcolor{green!20}3.43 & \cellcolor{green!20}3.68 & \cellcolor{green!20}4.91 & \cellcolor{green!20}1.92 & \cellcolor{green!20}7.85 & \cellcolor{green!20}2.01 & \cellcolor{green!20}4.11 & \cellcolor{green!20}4.32 & \cellcolor{green!20}3.15 & \cellcolor{green!20}3.65
  \\

\bottomrule
\end{tabular}
 }
\end{table}
In the main paper, we evaluate classification performance using top-1 accuracy and model calibration using Expected Calibration Error (ECE). Here we provide comprehensive results for additional calibration metrics including Adaptive Calibration Error (ACE)~\cite{ACEM} and Maximum Calibration Error (MCE)~\cite{ECE} to further validate the effectiveness of our approach.

Table~\ref{table:othermetrics_base} presents the MCE and ACE results on \textbf{base classes} across 10 fine-grained classification benchmarks. Our method demonstrates consistent improvements across both metrics for all evaluated prompt learning methods. For CoOp, our approach reduces the average MCE from 2.40 to 0.90 and ACE from 6.72 to 2.78, representing substantial calibration improvements. Similarly, with KGCoOp, we achieve reductions in MCE from 1.62 to 1.39 and ACE from 6.19 to 5.85. These results are particularly noteworthy as MCE captures the worst-case calibration error, indicating that our method not only improves average calibration but also reduces extreme miscalibration cases.

Table~\ref{table:othermetrics_novel} shows the results on \textbf{novel classes}. The improvements are consistent with the base class results, demonstrating the generalization capability of our calibration approach. For KGCoOp on novel classes, our method maintains similar MCE performance (1.18 vs 1.17) while slightly improving ACE from 3.93 to 3.65. The robustness across different calibration metrics validates that our approach addresses fundamental calibration issues rather than optimizing for specific metrics.

The consistent improvements across ECE, MCE, and ACE metrics provide strong evidence that our calibration method effectively reduces both average and worst-case calibration errors, making it suitable for deployment in safety-critical applications.% where reliable confidence estimates are essential.
\section{Medical Image Analysis} \label{sec:Medical Image Analysis}
\phantomsection

%Improved calibration in medical image analysis ensures that model predictions are not only accurate but also trustworthy, which is essential for clinical applications such as diagnosis and treatment planning.
%Medical image analysis using vision-language models to interpret and generate insights from medical imaging data. These models enable tasks such as automated diagnosis, report generation, and interactive analysis through multimodal learning. Here, evaluate the effectiveness and robustness of this proposed calibration method on the Medical Image and Med-VLMs. 
Furthermore, we evaluated our proposed solution on four Med-VLMs: PLIP \cite{PLIP}, QuiltNet \cite{QUILTNET}, using three downstream datasets: KatherColon (Kather) \cite{KATHER}, PanNuke \cite{PANNUKE}, and DigestPath \cite{da2022digestpath}. We compared the proposed method with other calibration methods, such as CE, MbLs, ZS-Norm, and penalty.

%The results demonstrate that the proposed method consistently performs well in all datasets and models, both in terms of classification accuracy and calibration error. %Key findings include a higher accuracy for which the method matches or exceeds the best-performing baselines in terms of accuracy for most datasets and models.

%Lower Calibration Error, which achieves the lowest expected calibration error in several cases, indicating more reliable confidence estimates. 

Table~\ref{table:results_plip_quiltnet} shows that the proposed method consistently yields the lowest ECE values when compared with Vanilla Cross Entropy Loss(CE). At the same time, our proposed method gives the lowest reduced Overall Average ECE value compared to other baselines, yielding 7.09 while maintaining the stable accuracy. 
\begin{table*}[!htp]

\centering
\caption[Comparison of PLIP & QuiltNet (CE) on Histopathology Datasets]{Comparison of proposed method (with baseline methods using Cross Entropy (CE). Accuracy (ACC, \%) and Expected Calibration Error (ECE, \%) are shown for PLIP and QuiltNet on histopathology datasets (Kather, PanNuke, DigestPath). Best results are in \textbf{bold}, second-best \underline{underlined}.}
\label{table:results_plip_quiltnet}
\setlength{\tabcolsep}{1.5pt}
\renewcommand{\arraystretch}{1.0} 
\resizebox{\linewidth}{!}{%
\begin{tabular}{l cccccc  cccccc cc}
\toprule
\rowcolor{orange!20} Model $\rightarrow$ & \multicolumn{6}{c}{\textbf{PLIP}} & \multicolumn{6}{c}{\textbf{QuiltNet}} & \multicolumn{2}{c}{\textbf{Average}}\\
\cmidrule(lr{3pt}){2-7} \cmidrule(lr{3pt}){8-13} \cmidrule(lr{3pt}){14-15}
\rowcolor{orange!20} Dataset $\rightarrow$ & \multicolumn{2}{c}{Kather} & \multicolumn{2}{c}{PanNuke} & \multicolumn{2}{c}{DigestPath}   & \multicolumn{2}{c}{Kather}  & \multicolumn{2}{c}{PanNuke} & \multicolumn{2}{c}{DigestPath} & \multicolumn{2}{c}{All} \\
\rowcolor{orange!20}Loss $\downarrow$ & ACC $\uparrow$ & ECE $\downarrow$ & ACC $\uparrow$ & ECE $\downarrow$ & ACC $\uparrow$ & ECE $\downarrow$ & Acc $\uparrow$ & ECE $\downarrow$ & ACC $\uparrow$ & ECE $\downarrow$ & ACC $\uparrow$ & ECE $\downarrow$ & ACC $\uparrow$ & ECE $\downarrow$\\ 
\midrule
\rowcolor{orange!20} \multicolumn{15}{c}{\textbf{Cross Entropy-based Losses}} \\
\cmidrule{1-15}
\rowcolor{white!10} Cross Entropy Loss$_{\textcolor{purple}{PL}}$ & 83.91 & 5.92  & 66.70 & 17.82 & 82.87 & 9.50  & 87.97 & 2.49  & 69.82 & 19.70 & 81.59 & 11.27 & 78.81 & 11.12 \\
\rowcolor{white!10} MbLS$_{\textcolor{purple}{PL}}$ & 84.39 & 3.57  & 66.70 & 17.82  & 82.76 & 9.53  & 84.76 & 3.48  & 65.54 & 23.05 & 82.58 & 10.90 & 77.79 & 11.39  \\
\rowcolor{white!10} ZS-Norm$_{\textcolor{purple}{PL}}$ & 85.63 & \underline{3.07}  & 71.52 & 17.22  & 82.84 & 7.31  & 91.91 & \textbf{0.87}  & 69.55 & 19.18 & 84.78 & 6.37 & 81.04 & 9.00  \\
\rowcolor{white!10} Penalty$_{\textcolor{purple}{PL}}$ & 86.48 & 3.90  & 70.49 & \textbf{3.11}  & 69.61 & \textbf{2.40}  & 89.29 & 12.28  & 59.88 & \textbf{3.41} & 79.23 & 17.53 & 75.83 & 7.11  \\
\midrule
\rowcolor{green!10} \textbf{Ours} & 87.98 & \textbf{1.31} & 65.72 & 12.40 & 84.84 & 4.71 & 88.70 & \underline{1.45} & 68.31 & 16.17 & 83.46 & 6.49 & 79.83 & 7.09 \\
\bottomrule
\end{tabular}
}
\renewcommand{\arraystretch}{1.0}

\end{table*}
\section{Loss Component Analyses}
\label{Loss_Component_Analyses}
\begin{figure*}[t]
    \centering
    \includegraphics[width=1.0\linewidth]{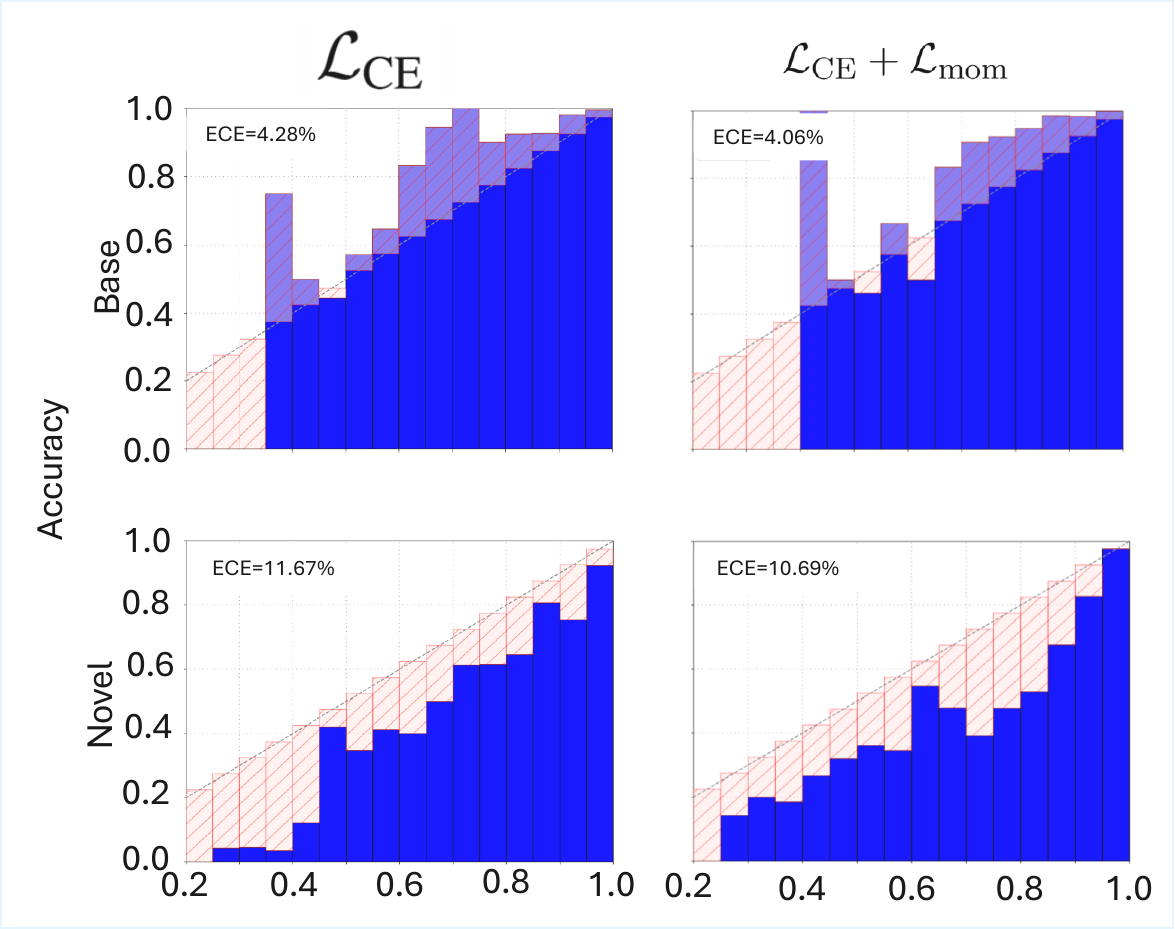}
    \caption{$\mathcal{L}_{\text{mom}}$ loss analyses on both Base and Novel Classes}
\vspace{-1.5em}
    \label{fig:lmom_an}
\end{figure*}
Fig.\ref{fig:lmom_an} shows how $\mathcal{L}_{\text{mom}}$ loss works directly with $\mathcal{L}_{\text{CE}}$ on both Base and Novel classes without $\mathcal{L}_{\text{margin}}$ loss. $\mathcal{L}_{\text{mom}}$ reduces the underconfidence in base classes marginally without any inter-class logit separation. However, it reduces the overconfidence issue alone(without having margin loss) in Novel classes by preserving CLIP's semantic geometry, maintaining relative class structure. As you can see in \textcolor{red}{Fig.2} of the main paper, when incorporating $\mathcal{L}_{\text{margin}}$ with $\mathcal{L}_{\text{mom}}$, it helps to reduce the miscalibration issue further more in both Base and Novel classes by having inter-class separation and maintaining relative class structure geometry.

\section{Hyperparameters Details} \label{sec:hyperparameters}
\begin{table}[ht]
\caption{Hyperparameter search for $\alpha$ (0.1–0.3) and $\beta$ (0.01 and 0.05), with results averaged across Caltech~\cite{Caltech-101}, Food101~\cite{food101}, and DTD~\cite{DTD}, reported on both base and novel classes.}
  \centering
  \small
  \setlength{\tabcolsep}{6pt}
  \begin{tabular}{c ccc ccc}
    \toprule
    \rowcolor{orange!20}
    & \multicolumn{3}{c}{ACC} & \multicolumn{3}{c}{ECE} \\
    
    \cmidrule(lr){2-4} \cmidrule(lr){5-7}
     \rowcolor{orange!20}
    $(\alpha,\beta)$ & Base & Novel & Avg & Base & Novel & Avg \\
    \midrule
    \rowcolor{green!20}
    $(0.1,0.01)$ & 89.65 & 82.38 & 86.02 & 1.99 & 5.66 & 3.83 \\
    $(0.2,0.01)$ & 89.72 & 79.66 & 84.69 & 3.46 & 8.53 & 5.99 \\
    $(0.3,0.01)$ & 89.28 & 80.72 & 85.00 & 4.46 & 9.50 & 6.98 \\
    $(0.1,0.05)$ & 89.12 & 79.99 & 84.56 & 3.95 & 6.26 & 5.10 \\
    $(0.2,0.05)$ & 83.39 & 81.49 & 82.44 & 4.10 & 8.20 & 6.15 \\
    $(0.3,0.05)$ & 89.75 & 80.52 & 85.14 & 2.01 & 8.01 & 5.01 \\
    \bottomrule
  \end{tabular}

  \label{tab:lambda_b_theme}
\end{table}
\begin{table}[ht]
\caption{Hyperparameter search for $\lambda_{\text{mom}}$ over values 1–10, with results averaged across Caltech~\cite{Caltech-101}, Food101~\cite{food101}, and DTD~\cite{DTD}, reported on both base and novel classes.}
  \centering
  \small
  \setlength{\tabcolsep}{6pt}
  \begin{tabular}{c ccc ccc}
    \toprule
     \rowcolor{orange!20}
    & \multicolumn{3}{c}{ACC} & \multicolumn{3}{c}{ECE} \\
    \cmidrule(lr){2-4} \cmidrule(lr){5-7}
     \rowcolor{orange!20}
    $\lambda$ & Base & Novel & Avg & Base & Novel & Avg \\
    \midrule
    1   & 89.44 & 81.51 & 85.48 & 3.47 & 4.54 & 4.01 \\
    3   & 89.17 & 82.14 & 85.65 & 2.11 & 3.94  & 3.03 \\
    \rowcolor{green!20}
    5   & 89.71 & 82.64 & 86.18 & 1.80 & 3.22  & 2.51 \\
    8   & 89.93 & 82.44 & 86.19 & 2.99 & 5.24  & 4.12 \\
    10  & 89.15 & 82.00 & 85.58 & 3.64 & 5.21  & 4.43 \\
    \bottomrule
  \end{tabular}

  \label{tab:lambda_variation}
\end{table}
For all experiments, we use CLIP (ViT-B/16)~\cite{radford2021learning} as the pre-trained vision-language model. Prompt-tuning is conducted in a few-shot setting with 16 samples per class, using a learning rate of 0.005 and a batch size of 8. For each baseline method, we adopt its official implementation and follow the recommended hyperparameter settings from the original papers. All experiments are performed on an NVIDIA RTX A6000 GPU with 48GB memory.

For our proposed calibration method, we use the following hyperparameters across all experiments: $\lambda_{\text{Margin}} = 1.0$ controls the strength of the margin-based regularization, $\alpha = 0.1$ balances the average marin, $\beta = 0.01$ is the weight for the variance loss, and $\lambda_{\text{mom}} = 5.0$ controls the local moment matching regularization. Table\ref{tab:lambda_b_theme} and \ref{tab:lambda_variation} show how we choose these values. These hyperparameters were %selected based on validation performance and remained 
fixed across all datasets and prompt learning methods to ensure fair comparison. We conduct 3 random seeds for each experiment and report the average results.

\section{Prompt Templates and Variations} \label{sec:prompt_var}

In the main paper, Figure {\color{red}4b} presents our method's robustness to different prompt initialization.
The following prompt templates were evaluated to assess initialization robustness: \texttt{``a nice image of a \{\}''}, \texttt{``an example of a \{\}''}, \texttt{``a picture of a \{\}''}, and \texttt{``a photo of the cool \{\}''}.
These templates represent different stylistic and semantic variations commonly used in prompt learning literature. 
This robustness is particularly valuable in practical deployment scenarios where optimal prompt initialization may not be known in advance.

\section{Variance Analysis} \label{sec:variance}
\begin{table}[t!]
%\small
\centering
\caption {\small  \textbf{Variance across 3 random seeds for 9 novel classes of fine-grained classification benchmarks}.}
\label{table:novel_variance}
\resizebox{0.95\textwidth}{!}{ 
\setlength{\tabcolsep}{4pt} % Reduce column spacing
\begin{tabular}{l c ||c c c c c c c c c c c}
\toprule
\thead{\textbf{Method}} & &  \multicolumn{1}{c}{\thead{\rotatebox{90}{Calt}}} & \multicolumn{1}{c}{\thead{\rotatebox{90}{Pets}}} & \multicolumn{1}{c}{\thead{\rotatebox{90}{Cars}}} & \multicolumn{1}{c}{\thead{\rotatebox{90}{Flow}}} & \multicolumn{1}{c}{\thead{\rotatebox{90}{Food}}} & \multicolumn{1}{c}{\thead{\rotatebox{90}{Air}}} & \multicolumn{1}{c}{\thead{\rotatebox{90}{SUN}}} & \multicolumn{1}{c}{\thead{\rotatebox{90}{DTD}}} & \multicolumn{1}{c}{\thead{\rotatebox{90}{Euro}}}  & \multicolumn{1}{c}{\thead{\rotatebox{90}{\textbf{Avg}}}}\\
\midrule
% \multirow{2}{*}{$\text{Zero Shot}$}
%  & Acc. & 72.40 & 97.20 & 91.30 & 63.60 & 71.80 & 90.10 & 27.70 & 69.40 & 53.00 & 57.00 & 71.00 & 69.50 \\
%  & ECE & 1.51 & 6.49 & 2.25 & 3.74 & 3.11 & 1.57 & 3.03 & 1.59 & 4.53 & 8.35 & 3.24 & 3.58 \\
\hline
\rowcolor{orange!20} \multicolumn{13}{c}{\textbf{CoCoOp}\cite{coop} } \\ 
\hline
\multirow{2}{*}{$\text{CoCoOp}$ \cite{coop}} 
 & Var. Acc & 0.81 & 0.08 & 2.65 & 1.53 & 4.41 & 0.01 & 3.61 & 0.19 & 0.69 &  1.55  \\
 & Var. ECE & 0.36 & 0.03 & 3.09 & 8.82 & 3.88 & 0.01 & 6.25 & 0.10 & 0.21 &  2.53 \\
%\hdashline[2pt/2pt]
%\multirow{2}{*}{$\text{DAC}$ \cite{dac}}
% & Acc & - & - & - & - & - & - & - & - & - & - & -  \\
% & ECE & - & - & - & - & - & - & - & - & - & - & -  \\
\hdashline[2pt/2pt]
\multirow{2}{*}{$\text{$\text{ZS-Norm}$}$ \cite{eccv}}
 & Var. Acc & 1.02 & 0.24 & 4.60 & 4.62 & 0.58 & 0.12 & 1.29 & 0.19 & 0.33 &  1.44 \\
 & Var. ECE & 1.69 & 0.12 & 0.88 & 6.66 & 4.45 & 0.10 & 1.04 & 0.23 & 5.11 &  2.25 \\
\hdashline[2pt/2pt]
\multirow{2}{*}{$\text{$\text{Penalty}$}$ \cite{eccv}}
 & Var. Acc & 0.12 & 0.09 & 0.11 & 0.32 & 0.59 & 5.81 & 2.25 & 0.09 & 0.08 &  1.05 \\
 & Var. ECE & 0.08 & 0.02 & 0.20 & 0.04 & 0.75 & 1.00 & 2.50 & 0.19 & 0.02 &  0.53 \\
\hdashline[2pt/2pt]
 \multirow{2}{*}{$\textbf{Ours}$}
 & Var. Acc & \cellcolor{green!20}0.05 & \cellcolor{green!20}0.01 & \cellcolor{green!20}7.72 & \cellcolor{green!20}1.21 & \cellcolor{green!20}1.96 & \cellcolor{green!20}0.07 & \cellcolor{green!20}0.01 & \cellcolor{green!20}0.18 & \cellcolor{green!20}0.01  & \cellcolor{green!20}1.25 \\
 & Var. ECE & \cellcolor{green!20}0.01 & \cellcolor{green!20}0.03 & \cellcolor{green!20}2.43 & \cellcolor{green!20}0.01 & \cellcolor{green!20}1.46 & \cellcolor{green!20}0.00 & \cellcolor{green!20}0.73 & \cellcolor{green!20}0.21 & \cellcolor{green!20}0.06 & \cellcolor{green!20}0.55 \\

\hline
\rowcolor{orange!20} \multicolumn{13}{c}{\textbf{KGCoOp} \cite{kgcoop}} \\ 
\hline
\multirow{2}{*}{$\text{KGCoOp}$ \cite{kgcoop}}
& Var. Acc & 0.03 & 0.00 & 2.19 & 1.21 & 0.58 & 0.01 & 0.85 & 0.15 & 0.08 &  0.57  \\
 & Var. ECE & 0.04 & 0.00 & 4.00 & 0.25 & 0.72 & 0.00 & 0.01 & 0.01 & 0.17 &  0.58  \\
\hdashline[2pt/2pt]
\multirow{2}{*}{$\text{$\text{ZS-Norm}$}$ \cite{eccv}}
 & Var. Acc & 0.05 & 0.00 & 7.29 & 0.92 & 0.50 & 0.01 & 3.35 & 0.35 & 0.00 &  1.39  \\
 & Var. ECE & 0.04 & 0.00 & 2.31 & 0.07 & 0.01 & 0.01 & 1.80 & 0.00 & 0.04 &  0.48  \\
 \hdashline[2pt/2pt]
\multirow{2}{*}{$\text{$\text{Penalty}$}$ \cite{eccv}}
 & Var. Acc & 0.01  &0.00 & 1.66 & 0.16 & 1.23 & 0.01 & 1.04 & 0.11 & 0.15 &  0.49  \\
 & Var. ECE & 0.08 & 0.06 & 0.38 & 0.18 & 1.19 & 0.01 & 2.37 & 0.07 & 0.06 &  0.49  \\
  \hdashline[2pt/2pt]
\multirow{2}{*}{$\textbf{Ours}$}
 & Var. Acc & \cellcolor{green!20}0.05 & \cellcolor{green!20}0.00 & \cellcolor{green!20}2.31 & \cellcolor{green!20}0.17 & \cellcolor{green!20}0.96 & \cellcolor{green!20}0.02 & \cellcolor{green!20}0.00 & \cellcolor{green!20}0.40 & \cellcolor{green!20}0.04  & \cellcolor{green!20}0.44 \\
 & Var. ECE & \cellcolor{green!20}0.02 & \cellcolor{green!20}0.00 & \cellcolor{green!20}0.16 & \cellcolor{green!20}0.74 & \cellcolor{green!20}0.44 & \cellcolor{green!20}0.04 & \cellcolor{green!20}0.02 & \cellcolor{green!20}0.10 & \cellcolor{green!20}0.36  & \cellcolor{green!20}0.21 \\

\bottomrule
\end{tabular}
 }
\end{table}
To assess the statistical robustness of our approach, we evaluate the variance in performance across 3 random seeds for both accuracy and calibration metrics. Table~\ref{table:novel_variance} presents the variance results across 9 novel classes of fine-grained classification benchmarks for CoCoOp and KGCoOp methods. Our approach demonstrates superior stability with consistently lower variance in both accuracy and ECE compared to baseline calibration methods. For CoCoOp, our method achieves significantly lower average variance in accuracy (1.25 vs 1.55) and ECE (0.55 vs 2.53) compared to the vanilla baseline. Similarly, with KGCoOp, we maintain competitive variance performance with average accuracy variance of 0.44 compared to the baseline's 0.57, while substantially reducing ECE variance from 0.58 to 0.21. The reduced variance in calibration error is particularly noteworthy as it indicates that our method provides more consistent and reliable confidence estimates across different experimental runs, which is crucial for deployment in safety-critical applications.% where predictable model behavior is essential.
\section{Results on Different Backbones} \label{sec:backbones}
To evaluate the adaptability of our method, we conduct experiments on CoOp \cite{coop} with different backbones, namely RN-50 and ViT-B/32. The Tables\ref{table:supp_base_backbone} and \ref{table:supp_Novel_backbone}results show that our approach consistently outperforms existing methods %in terms of ECE on base classes
across both backbones, while also maintaining improvements in accuracy. In base classes, for RN-50, our method achieves an average ECE of 3.46 compared to 4.04 for the vanilla baseline. Similarly, for ViT-B/32, our method attains an ECE of 2.87, outperforming the vanilla baseline at 3.15. For novel classes, our method achieves the second-lowest average ECE of 5.46 on RN-50, with ZS-Norm \cite{eccv} performing slightly better at 5.23. In contrast, on ViT-B/32, our method achieves the lowest ECE of 5.82, surpassing all other approaches.
%\FloatBarrier
\begin{table}[t]
%\small
\centering
\caption {\small  \textbf{ Accuracy and calibration performance on base classes across 10 fine-grained classification benchmarks using RN-50 and ViT-B/32}.
We report top-1 accuracy (Acc) and Expected Calibration Error (ECE) for CoOp\cite{coop}, a prompt-tuning strategy, evaluated with different backbones.}
\label{table:supp_base_backbone}
\resizebox{\textwidth}{!}{ 
\setlength{\tabcolsep}{4pt} % Reduce column spacing
\begin{tabular}{l c ||c c c c c c c c c c c}
\toprule
\thead{\textbf{Method}} & &  \multicolumn{1}{c}{\thead{\rotatebox{90}{Calt}}} & \multicolumn{1}{c}{\thead{\rotatebox{90}{Pets}}} & \multicolumn{1}{c}{\thead{\rotatebox{90}{Cars}}} & \multicolumn{1}{c}{\thead{\rotatebox{90}{Flow}}} & \multicolumn{1}{c}{\thead{\rotatebox{90}{Food}}} & \multicolumn{1}{c}{\thead{\rotatebox{90}{Air}}} & \multicolumn{1}{c}{\thead{\rotatebox{90}{SUN}}} & \multicolumn{1}{c}{\thead{\rotatebox{90}{DTD}}} & \multicolumn{1}{c}{\thead{\rotatebox{90}{Euro}}} & \multicolumn{1}{c}{\thead{\rotatebox{90}{UCF}}} & \multicolumn{1}{c}{\thead{\rotatebox{90}{\textbf{Avg}}}}\\
\midrule
\rowcolor{orange!20} \multicolumn{13}{c}{\textbf{CoOp-RN50}\cite{coop} } \\ 
\hline
\multirow{2}{*}{$\text{CoOp-RN50}$ \cite{coop}} 
 & Acc. & 95.22 & 90.5 & 70.22 & 95.25 & 82.54 & 29.95 & 76.37 & 74.85 & 81.32 & 80.4 & 76.99 \\
 & ECE & 1.27 & 2.32 & 6.15 & 4.22 & 1.15 & 2.62 & 2.81 & 8.65 & 9.34 & 1.82 & 4.04 \\
\hdashline[2pt/2pt]
\multirow{2}{*}{$\text{$\text{ZS-Norm}$}$ \cite{eccv}}
 & Acc. & 95.55 & 90.96 & 69.89 & 95.19 & 82.71 & 25.89 & 76.67 & 74.27 & 89.57 & 79.58 & 78.03 \\
 & ECE & 4.04 & 5.63 & 10.52 & 9.92 & 3.4 & 17.28 & 4.35 & 37.15 & 35.76 & 7.18 & 13.52 \\
 \hdashline[2pt/2pt]
\multirow{2}{*}{$\text{$\text{Penalty}$}$ \cite{eccv}}
 & Acc. & 96.31 & 92.1 & 68.58 & 94.65 & 83.66 & 26.63 & 74.57 & 68.48 & 47.61 & 77.9 & 78.12 \\
 & ECE & 6.21 & 8.69 & 11.85 & 11.34 & 5.29 & 6.4 & 5.68 & 18.93 & 22.61 & 9.95 & 10.71 \\
  \hdashline[2pt/2pt]
 \multirow{2}{*}{$\textbf{Ours}$}
 & Acc. & \cellcolor{green!20}95.44 & \cellcolor{green!20}91.08 & \cellcolor{green!20}69.90 & \cellcolor{green!20}95.22 & \cellcolor{green!20}82.77 & \cellcolor{green!20}29.01 & \cellcolor{green!20}77.07 & \cellcolor{green!20}75.42 & \cellcolor{green!20}89.79 & \cellcolor{green!20}79.96 & \cellcolor{green!20}78.57 \\
 & ECE & \cellcolor{green!20}1.37 & \cellcolor{green!20}2.21 & \cellcolor{green!20}8.65 & \cellcolor{green!20}4.74 & \cellcolor{green!20}0.84 & \cellcolor{green!20}3.32 & \cellcolor{green!20}1.23 & \cellcolor{green!20}5.87 & \cellcolor{green!20}4.12 & \cellcolor{green!20}2.21 & \cellcolor{green!20}3.46 \\
\hline
\rowcolor{orange!20} \multicolumn{13}{c}{\textbf{CoOp--ViT-B/32} \cite{coop}} \\ 
\hline
\multirow{2}{*}{$\text{CoOp--ViT-B/32}$ \cite{coop}}
 & Acc. & 97.05 & 92.71 & 73.13 & 95.28 & 84.93 & 31.87 & 79.16 & 77.55 & 91.10 & 82.83 & 79.88 \\
 & ECE & 1.17 & 2.45 & 4.86 & 4.18 & 1.06 & 3.12 & 2.5 & 6.15 & 3.91 & 2.09 & 3.15 \\
\hdashline[2pt/2pt]
\multirow{2}{*}{$\text{$\text{ZS-Norm}$}$ \cite{eccv}}
 & Acc. & 97.18 & 92.08 & 72.86 & 94.81 & 85.12 & 32.19 & 79.64 & 76.23 & 90.09 & 82.76 & 80.25 \\
 & ECE & 1.98 & 7.96 & 9.70 & 7.91 & 6.00 & 12.97 & 3.48 & 38.62 & 39.31 & 5.25 & 13.32 \\
 \hdashline[2pt/2pt]
\multirow{2}{*}{$\text{$\text{Penalty}$}$ \cite{eccv}}
 & Acc. & 96.73 & 93.22 & 73.33 & 94.75 & 86.05 & 29.99 & 78.95 & 69.64 & 53.42 & 5.94 & 75.78 \\
 & ECE & 4.88 & 6.66 & 10.54 & 10.05 & 4.45 & 4.82 & 5.42 & 22.00 & 22.15 & 5.94 & 9.69 \\
  \hdashline[2pt/2pt]
 \multirow{2}{*}{$\textbf{Ours}$}
 & Acc. & \cellcolor{green!20}97.46 & \cellcolor{green!20}93.14 & \cellcolor{green!20}73.78 & \cellcolor{green!20}95.09 & \cellcolor{green!20}85.12 & \cellcolor{green!20}31.83 & \cellcolor{green!20}79.58 & \cellcolor{green!20}79.01 & \cellcolor{green!20}91.09 & \cellcolor{green!20}82.87 & \cellcolor{green!20}80.09 \\
 & ECE & \cellcolor{green!20}0.92 & \cellcolor{green!20}1.68 & \cellcolor{green!20}6.77 & \cellcolor{green!20}4.53 & \cellcolor{green!20}0.79 & \cellcolor{green!20}2.91 & \cellcolor{green!20}0.95 & \cellcolor{green!20}4.77 & \cellcolor{green!20}3.52 & \cellcolor{green!20}1.87 & \cellcolor{green!20}2.87 \\
\bottomrule
\end{tabular}
 }
\end{table}
\begin{table}[t]
%\small
\centering
\caption {\small  \textbf{ Accuracy and calibration performance on novel classes across 10 fine-grained classification benchmarks using RN-50 and ViT-B/32}.
We report top-1 accuracy (Acc) and Expected Calibration Error (ECE) for CoOp\cite{coop}, a prompt-tuning strategy, evaluated with different backbones.}
\label{table:supp_Novel_backbone}
\resizebox{\textwidth}{!}{ 
\setlength{\tabcolsep}{4pt} % Reduce column spacing
\begin{tabular}{l c ||c c c c c c c c c c c}
\toprule
\thead{\textbf{Method}} & &  \multicolumn{1}{c}{\thead{\rotatebox{90}{Calt}}} & \multicolumn{1}{c}{\thead{\rotatebox{90}{Pets}}} & \multicolumn{1}{c}{\thead{\rotatebox{90}{Cars}}} & \multicolumn{1}{c}{\thead{\rotatebox{90}{Flow}}} & \multicolumn{1}{c}{\thead{\rotatebox{90}{Food}}} & \multicolumn{1}{c}{\thead{\rotatebox{90}{Air}}} & \multicolumn{1}{c}{\thead{\rotatebox{90}{SUN}}} & \multicolumn{1}{c}{\thead{\rotatebox{90}{DTD}}} & \multicolumn{1}{c}{\thead{\rotatebox{90}{Euro}}} & \multicolumn{1}{c}{\thead{\rotatebox{90}{UCF}}} & \multicolumn{1}{c}{\thead{\rotatebox{90}{\textbf{Avg}}}}\\
\midrule
\rowcolor{orange!20} \multicolumn{13}{c}{\textbf{CoOp-RN50}\cite{coop} } \\ 
\hline
\multirow{2}{*}{$\text{CoOp-RN50}$ \cite{coop}} 
 & Acc. & 87.27 & 92.11 & 57.84 & 61.87 & 82.55 & 18.72 & 64.46 & 41.67 & 34.15 & 55.07 & 11.38 \\
 & ECE & 3.57 & 2.10 & 8.08 & 10.24 & 0.86 & 18.75 & 9.07 & 25.47 & 22.18 & 13.47 & 11.38 \\
\hdashline[2pt/2pt]
\multirow{2}{*}{$\text{$\text{ZS-Norm}$}$ \cite{eccv}}
 & Acc. & 88.39 & 89.97 & 57.86 & 59.93 & 81.52 & 20.34 & 65.77 & 35.79 & 40.08 & 57.37 & 59.70\\
 & ECE & 4.30 & 3.09 & 2.43 & 5.99 & 6.33 & 4.37 & 2.64 & 10.97 & 9.24 & 2.98 & 5.23 \\
 \hdashline[2pt/2pt]
\multirow{2}{*}{$\text{$\text{Penalty}$}$ \cite{eccv}}
 & Acc. & 88.75 & 93.38 & 61.37 & 83.63 & 20.22 & 67.45 & 45.21 & 31.57 & 58.16 & 3.01 & 61.05 \\
 & ECE & 3.05 & 7.94 & 1.7 & 4.81 & 8.24 & 10.02 & 2.92 & 9.96 & 11.67 & 3.01 & 6.33 \\
  \hdashline[2pt/2pt]
 \multirow{2}{*}{$\textbf{Ours}$}
 & Acc. & \cellcolor{green!20}87.52 & \cellcolor{green!20}93.08 & \cellcolor{green!20}59.88 & \cellcolor{green!20}61.47 & \cellcolor{green!20}82.50 & \cellcolor{green!20}21.66 & \cellcolor{green!20}64.94 & \cellcolor{green!20}39.05 & \cellcolor{green!20}42.75 & \cellcolor{green!20}53.94 & \cellcolor{green!20}60.68 \\
 & ECE & \cellcolor{green!20}2.82 & \cellcolor{green!20}2.68 & \cellcolor{green!20}3.43 & \cellcolor{green!20}4.13 & \cellcolor{green!20}1.72 & \cellcolor{green!20}5.36 & \cellcolor{green!20}4.53 & \cellcolor{green!20}8.83 & \cellcolor{green!20}10.93 & \cellcolor{green!20}10.16 & \cellcolor{green!20}5.46 \\
\hline
\rowcolor{orange!20} \multicolumn{13}{c}{\textbf{CoOp--ViT-B/32} \cite{coop}} \\ 
\hline
\multirow{2}{*}{$\text{CoOp--ViT-B/32}$ \cite{coop}}
 & Acc. & 92.25 & 94.00 & 60.04 & 60.31 & 85.16 & 22.12 & 68.98 & 47.95 & 56.47 & 63.57 & 64.63 \\
 & ECE & 3.29 & 2.30 & 8.53 & 13.28 & 1.33 & 16.75 & 8.57 & 19.74 & 17.18 & 10.32 & 10.13 \\
\hdashline[2pt/2pt]
\multirow{2}{*}{$\text{$\text{ZS-Norm}$}$ \cite{eccv}}
 & Acc. & 91.56 & 93.46 & 59.01 & 54.09 & 84.50 & 21.64 & 70.14 & 43.44 & 52.04 & 65.98 & 63.59 \\
 & ECE & 2.62 & 7.94 & 4.05 & 10.94 & 6.86 & 5.59 & 1.34 & 17.82 & 14.8 & 3.13 & 7.51 \\
 \hdashline[2pt/2pt]
\multirow{2}{*}{$\text{$\text{Penalty}$}$ \cite{eccv}}
 & Acc. & 92.39 & 96.12 & 59.83 & 53.78 & 86.58 & 23.90 & 70.52 & 45.45 & 44.66 & 61.77 & 63.560 \\
 & ECE & 4.41 & 6.93 & 2.66 & 7.31 & 5.23 & 8.63 & 2.19 & 8.77 & 10.71 & 4.22 & 6.11 \\
  \hdashline[2pt/2pt]
 \multirow{2}{*}{$\textbf{Ours}$}
 & Acc. & \cellcolor{green!20}91.41 & \cellcolor{green!20}93.48 & \cellcolor{green!20}61.10 & \cellcolor{green!20}61.65 & \cellcolor{green!20}84.22 & \cellcolor{green!20}23.08 & \cellcolor{green!20}70.59 & \cellcolor{green!20}50.23 & \cellcolor{green!20}56.55 & \cellcolor{green!20}64.02 & \cellcolor{green!20}65.63 \\
 & ECE & \cellcolor{green!20}2.23 & \cellcolor{green!20}2.15 & \cellcolor{green!20}4.33 & \cellcolor{green!20}5.92 & \cellcolor{green!20}0.74 & \cellcolor{green!20}9.90 & \cellcolor{green!20}3.72 & \cellcolor{green!20}14.30 & \cellcolor{green!20}9.14 & \cellcolor{green!20}5.80 & \cellcolor{green!20}5.82 \\
\bottomrule
\end{tabular}
 }
\end{table}

\section{Decision Boundary Visualization} \label{sec:decision_b}

\begin{figure}[htp]
    \centering    \includegraphics[scale=0.40, trim=10 10 10 10, clip]{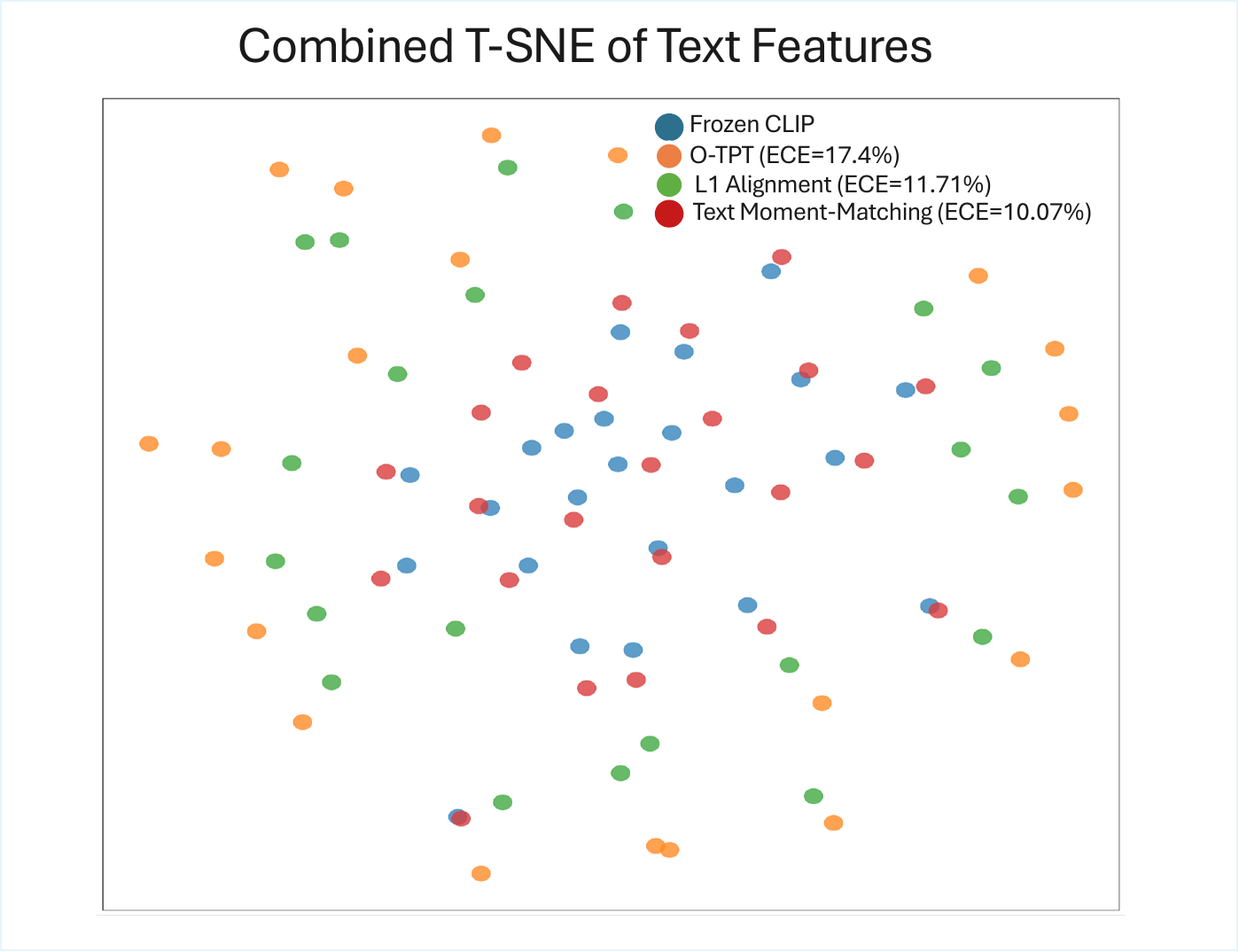}
    \caption{\small T-SNE visualization of Text Features }
    \label{fig:decision_bound}
\end{figure}

Figure \ref{fig:decision_bound} shows that the Text Momentum-Matching loss better preserves the geometric structure of CLIP’s pretrained embedding space by aligning the statistical moments of tuned and frozen text embeddings, compared to $\ell_1$ alignment or Orthogonality-based class-wise dispersion \cite{sharifdeen2025tpt} on novel classes.

\section{Computational Analyses}
\label{sec:Computational Analyses}
\begin{table}[t]
\centering
\caption{Training Time taken per epoch (in seconds) and peak GPU memory usage for 16-shot experiment on Flower~\cite{flowers} datasets}
\label{table:computa_ana}
\resizebox{1.0\textwidth}{!}{ 
\setlength{\tabcolsep}{4pt}
\begin{tabular}{l|| cc}
\toprule
\thead{\textbf{Method}} 
  & \thead{Total-Time-Per Batch} 
  & \thead{Peak GPU Memory} \\
\midrule
\multirow{1}{*}{MaPLe~\cite{maple}}
  & 15.91(seconds) & 1.75 GB\\
\hdashline[2pt/2pt]
\multirow{1}{*}{\textbf{Ours}}
  & 15.83(seconds) & 1.75 GB \\
\bottomrule
\end{tabular}
}
%\vspace{-0.5em}
\end{table}

To analyse the computational and memory overhead, we measured training time and peak GPU memory on Flower101\cite{flowers} 16-shot with ViT-B/16, using the original MaPLe\cite{maple} configuration as baseline. Compared to MaPLe, our method has less training time and the same GPU usage.

\section{Regularizer in scale:}
In Fig.~\ref{fig:rho_curves} and Tab.~\ref{tab:rho_quantiles}, we log \emph{gradient-norm ratios} to check optimisation-scale dominance. %(updates are driven by gradients) 
For the margin regularizer $L_{\text{Margin}}=L_{\text{mean}}+L_{\text{var}}$ with $L_{\text{mean}}=-\alpha\,\mathbb{E}[m]$ and $L_{\text{var}}=\beta\,\mathrm{Var}(m)$, and measure $\rho_{\text{margin}}=\frac{\|\nabla_{\theta}L_{\text{mean}}\|}{\|\nabla_{\theta}L_{\text{var}}\|+\varepsilon}$, which is bounded (median $2.65$; 10--90\% $[1.54,\,5.07]$), indicating comparable contributions of the linear mean and quadratic variance terms. For moment matching, with $L_{\mu}=\|\Delta\mu\|_{2}^{2}$ and $L_{\Sigma}=\|\Delta\Sigma\|_{F}^{2}$, we log $\rho_{\text{mom}}=\frac{\|\nabla_{\theta}(\lambda_{\text{mom}}L_{\mu})\|}{\|\nabla_{\theta}(\lambda_{\text{mom}}L_{\Sigma})\|+\varepsilon}$, which after a brief transient stabilises and shows no covariance-term dominance (median $16.92$; 10--90\% $[10.5,\,80.84]$). %countering the ``4th-order dominates'' concern. 
Across regularizers, $\rho_{\text{mom/margin}}=\frac{\|\nabla_{\theta}L_{\text{mom}}\|}{\|\nabla_{\theta}L_{\text{margin}}\|+\varepsilon}$ stays near a constant scale (median $0.38$; 10--90\% $[0.18,\,1.30]$), and both remain non-trivial yet typically subdominant to CE: $\rho_{\text{margin/CE}}=\frac{\|\nabla_{\theta}L_{\text{margin}}\|}{\|\nabla_{\theta}L_{\text{CE}}\|+\varepsilon}$ and $\rho_{\text{mom/CE}}=\frac{\|\nabla_{\theta}L_{\text{mom}}\|}{\|\nabla_{\theta}L_{\text{CE}}\|+\varepsilon}$ have medians $0.28$ and $0.17$. Thus, our objective is well-conditioned despite mixed-order terms.
%Overall, these diagnostics confirm the objective is well-conditioned and mixed-order terms do not induce optimisation-scale dominance. 
$\varepsilon=10^{-12}$ only for numerical stability.
{\floatsetup[table]{capposition=top}
 \floatsetup[figure]{capposition=top}
\begin{figure}[t]
\centering
\begin{floatrow}
% ---------------- Left: TABLE ----------------
%\begin{minipage}[t]{0.5\columnwidth}
\ttabbox[0.52\columnwidth]{%
  \caption{\small Grad-norm ratio stats.}%
  \label{tab:rho_quantiles}%
}{%
\centering
%\vspace{-1em}
\resizebox{0.98\linewidth}{!}{%
  \setlength{\tabcolsep}{4pt}%
  \renewcommand{\arraystretch}{1.10}%
  \begin{tabular}{lccc}
    \toprule
    \rowcolor{orange!20}
    \textbf{metric} & \textbf{median} & \textbf{q10} & \textbf{q90} \\
    \midrule
    $\rho_{\text{margin}}$  & 2.65  & 1.54  & 5.07 \\
    $\rho_{\text{mom}}$    & 16.92  & 10.50  & 80.84  \\
    $\rho_{\text{mom/margin}}$   & 0.38  & 0.18  & 1.30 \\
    $\rho_{\text{margin/CE}}$   & 0.28  & 0.18  & 1.42 \\
    $\rho_{\text{mom/CE}}$  & 0.17  & 0.05  & 0.51  \\
    \bottomrule
  \end{tabular}%
}
}
%\captionsetup{font=footnotesize}
%\caption{\small Grad-norm ratio statistics}
%\label{tab:rho_quantiles}
%\end{minipage}%
%\vspace{-1em}
%\hfill
\hspace{-1.8em}
% ---------------- Right: FIGURE ----------------
%\begin{minipage}[t]{0.5\columnwidth}
% \vspace{-1em}
\ffigbox[0.46\columnwidth]{%
  \caption{\small Grad-norm ratios.}% \vspace{-1.2em}
  \label{fig:rho_curves}%
}{%
\centering
% \vspace{-2em}
%\vspace{3em}
\includegraphics[width=\linewidth,trim=10 10 10 10,clip]{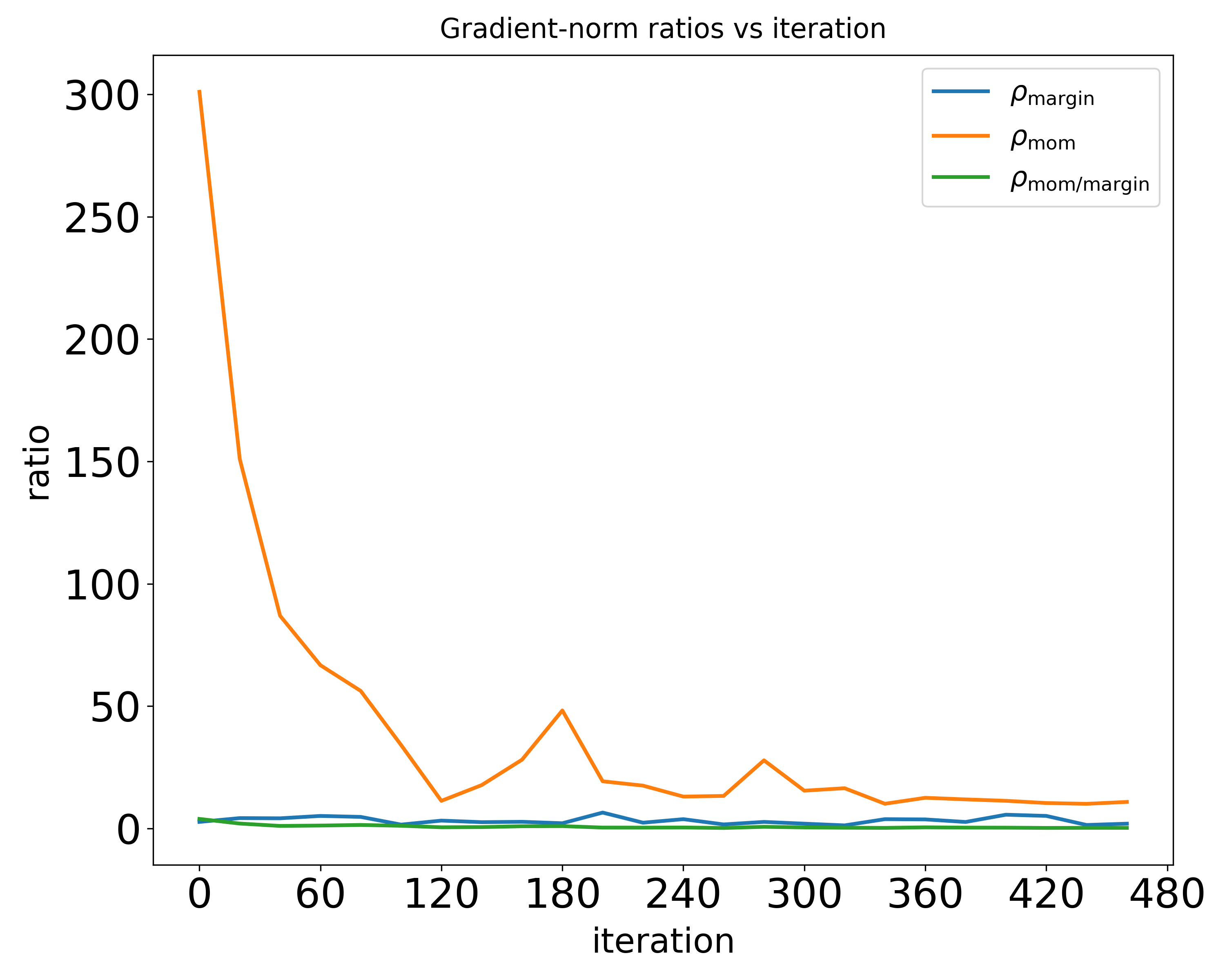}
%\vspace{-2em}
%\captionsetup{font=footnotesize}
% \caption{\small Gradient-norm ratios}
%\label{fig:rho_curves}
%\end{minipage}
}
\end{floatrow}
%\vspace{-1em}
\end{figure}
}

%\input{tables/varience_base}

%\input{tables/varience_novel}
%%%%%%%%%%%%%%%%%%%%%%%%%%%%%%%%%%%%%%%%%%%%%%%%%%%%%%%%%%%%

%\bibliography{main}
%\bibliographystyle{abbrv}
%\iffalse
%\textbf{References}

\end{document}